\documentclass{ieeeaccess}
\usepackage{cite}
\usepackage{amsmath,amssymb,amsfonts}
\usepackage{algorithmic}
\usepackage{graphicx}
\usepackage{textcomp} 
\usepackage{amssymb}
\usepackage{amsmath}
\usepackage{siunitx}
\usepackage{tabularx}
\usepackage{algorithmic}

\usepackage{algorithm2e} 
\usepackage{amsthm}
\usepackage{graphicx}
\usepackage{lscape}
\usepackage{verbatim}
\usepackage{color,soul}
\usepackage{subfigure}
\usepackage{tabularx,ragged2e,booktabs,caption,array,multirow,multicol} 
\usepackage{mathtools}  
\usepackage{siunitx} 
\usepackage{csvsimple}
\usepackage[utf8]{inputenc}
\usepackage[english]{babel}
\usepackage{blindtext}
\usepackage{multicol}
\usepackage{url}

\def\BibTeX{{\rm B\kern-.05em{\sc i\kern-.025em b}\kern-.08em
    T\kern-.1667em\lower.7ex\hbox{E}\kern-.125emX}}

\begin{document}

\history{Date of publication 31 May 2021}
\doi{https://doi.org/10.1109/ACCESS.2021.3085085}

\title{Evaluation of deep  learning models for multi-step ahead time series prediction}
\author{\uppercase{Rohitash Chandra}\authorrefmark{1} \IEEEmembership{Senior Member, IEEE},
\uppercase{Shaurya Goyal \authorrefmark{2}, and Rishabh Gupta}.\authorrefmark{3} }

\address[1]{School of Mathematics and Statistics, University of New South Wales,  Sydney, NSW 2052, Australia}
\address[2]{ Department of Mathematics, Indian Institute of Technology, Delhi, 110016 , India  }
\address[3]{Department of Geology and Geophysics, Indian Institute of Technology,Kharagpur, 721302 , India}
 

\corresp{Corresponding author: R. Chandra (e-mail: rohitash.chandra@unsw.edu.au)}

\begin{abstract} 
Time series prediction with neural networks has been the focus of much research in the past few decades. Given the recent deep learning revolution, there has been much attention in using deep learning models for time series prediction, and hence it is important to evaluate their strengths and weaknesses.  In this paper, we present an evaluation study that compares the performance of deep learning models for multi-step ahead time series prediction.  The deep learning methods \textcolor{black}{comprise} simple recurrent neural networks,   long short-term memory (LSTM) networks, bidirectional  LSTM networks,  encoder-decoder LSTM networks, and convolutional neural networks. We provide a further comparison with simple neural networks that use stochastic gradient descent and adaptive moment estimation (Adam) for training. We focus on univariate time series for multi-step-ahead prediction from benchmark time-series datasets and provide a further comparison of the results with related methods from the literature. The results show that the bidirectional and encoder-decoder LSTM network provides the best performance in accuracy for the given time series problems.

\end{abstract}


\titlepgskip=-15pt

\maketitle

\section{Introduction}
\label{sec:introduction}

Apart from econometric models, machine learning methods became extremely popular for time series  prediction and forecasting in the last few decades \cite{tealab2018time,cheng2015time,taieb2012review,ahmed2010empirical,Gooijer2006,li2005recent,hendry1983econometric}. Some of the popular categories include one-step, multi-step, and multivariate prediction. Recently, some attention has been given to dynamic time series prediction where the size of the input to the model can   dynamically change \cite{chandra2018co}. Just as the term indicates,  one-step prediction refers to the use of a model to make a prediction one-step ahead in time whereas a multi-step prediction refers to a series of steps ahead in time  from an observed trend in a time series 
\cite{sandya2013feature, chandra2012cooperative}. In the latter case, the \textit{prediction horizon}
defines the extent  of future prediction. The challenge is to develop models that 
produce low  
prediction errors as the prediction horizon increases  given the chaotic nature 
and noise in the dataset
\cite{taieb2015bias,chang2012reinforced,bone2002multi}. There are two major approaches  for  multi-step-ahead 
prediction which include  \textit{recursive} and  
\textit{direct}  strategies. The recursive strategy features the  prediction from 
a one-step-ahead prediction model  as the input for future
prediction horizon
\cite{zhang2013iterated,ben2012recursive}, where 
error in the prediction for the next horizon is accumulated in future horizons. 
The direct strategy encodes  the multi-step-ahead problem   as a multi-output problem   \cite{Sorjamaa2007,BenTaieb2010},  which in the case of neural networks can be represented by multiple neurons in the output layer for the prediction horizons. 
The major challenges in multi-step-ahead prediction include   highly chaotic time series 
and those that have missing data which has been approached  with non-linear filters and neural 
networks
\cite{Wu2014missingdate}.

Neural networks have been popular for time series prediction for various applications \cite{frank2001time}. Different neural network architectures have different strengths and weaknesses. Time series prediction requires careful integration of knowledge in temporal sequences; hence, it is important to choose the right neural network architecture and training algorithm. Recurrent neural networks (RNNs) are well known  for  modelling temporal sequences  \cite{elman_Zipser1988,Werbos_1990, connor1994recurrent,hochreiter1997long,schmidhuber2015deep} and  dynamical systems when compared to feedforward networks  \cite{Omlin_thonberetal1996, Omlin_Giles1992,Giles_etal1999}.    The Elman RNN   \cite{elman_Zipser1988,Elman_1990} is one of the earliest architectures to be  trained by backpropagation through-time, which is an extension of  the backpropagation algorithm   \cite{Werbos_1990}. The limitation  in   learning  long-term dependencies in temporal sequences  using canonical RNNs  \cite{hochreiter1998vanishing,bengio1994learning} have been addressed by  \textit{long short-term memory} (LSTM)  networks   \cite{hochreiter1997long}.

Recent  deep learning revolution \cite{schmidhuber2015deep} contributed to further improvements  in LSTM  networks with \textit{gated recurrent unit} (GRU)  networks \cite{chung2014empirical,cho2014learning}, which provides similar performance and are simpler to implement. Some of the other extensions include  predictive state  RNNs \cite{downey2017predictive} that combines RNNs with the  power  of predictive state representation  \cite{singh2004predictive}.  Bidirectional RNNs connect two hidden layers of opposite directions to the same output, where the output layer can get information from past   and future  states simultaneously \cite{schuster1997bidirectional}. The idea was further extended into bidirectional-LSTM networks for phoneme classification \cite{graves2005framewise} which performed better than standard RNNs and LSTM networks. Further work has been \textcolor{black}{done} by combining bidirectional LSTM networks with convolutional neural networks (CNNs) for natural language processing with problem of named entity recognition \cite{chiu2016named}. Further extensions have been \textcolor{black}{done} by encoder-decoder LSTM networks that used a LSTM to map the input sequence to a vector of a fixed dimensionality, and uses another LSTM to decode the target sequence for language task such as  English to French  translation  \cite{sutskever2014sequence}.  CNNs with  regularisation methods such as dropouts during training can  improve generalisation \cite{srivastava2014dropout}. Adaptive gradient methods such as the adaptive moment estimation (Adam optimiser) has become very prominent for training neural networks \cite{Adams2017}. Apart from these,  neuroevolution that uses evolutionary algorithms and multi-task learning have been used for time series prediction \cite{Chandra2018NC-CMTL,chandra2018co}.
 RNNs have also been trained by neuroevolution with applications for time series prediction \cite{cai2007time,chandra2012cooperative}.

 We note that limited work has \textcolor{black}{been done to compare} FNN and RNNs for multi-step time series prediction \cite{koskela1996time,chandra2016evaluation}. It is important to evaluate the advancements of deep learning methods  for a challenging problem which in our case is multi-step time series prediction.    LSTM network  applications have dominated applications in natural language processing and signal processing  such as phoneme recognition; however, there is no work that evaluates their performance for time series prediction, particularly multi-step ahead prediction. Since the underlying feature of LSTM networks is in \textcolor{black}{handling} temporal sequences, it is worthwhile to investigate   their predictive power, i.e. accuracy as the prediction horizon increases.  

 In this paper, we present an evaluation study that compares the performance  of selected   deep learning models  for multi-step ahead time series prediction.  We   examine univariate  time series prediction  with selected models and learning algorithms for benchmark time series datasets. The  deep learning methods \textcolor{black}{comprise} of   standard LSTM, bidirectional  LSTM,    encoder-decoder LSTM, and CNNs. We also compare the results with canonical  neural networks that use stochastic gradient descent learning and Adam optimiser.  We further compare our results with other related machine learning methods for multi-step time series prediction  from the literature.

The rest of the paper is organised as follows. Section 2 presents a background and literature review of related work. Section 3 presents the details of the different deep learning models,  and Section 4 presents experiments and results. Section 5 provides a discussion and Section 6 concludes the paper with discussion of future work. 

\section{Related Work} 

\subsection{Multi-step time series prediction}


One of the first attempts for  recursive strategy multi-step-ahead prediction used state-space Kalman filter and smoothing      \cite{ng1990recursive} followed by  recurrent neural 
networks  \cite{Su1992}.  Later,   a dynamic recurrent network used current and delayed  observations  as inputs
to the network which reported   
excellent generalization performance  
\cite{Parlos2000}. The
non-parametric Gaussian process model was used to   incorporate the uncertainty about intermediate regressor 
values \cite{BeckerNIPS2002}.  The \textit{Dempster–Shafer} regression technique  for  prognosis of data-driven machinery used iterative strategy      with promising 
performance \cite{Niu2009}.   Lately, reinforced real-time recurrent learning  was used with iterative strategy for  flood forecasts  
\cite{chang2012reinforced}.  One of the earliest work \textcolor{black}{done} using direct strategy for multi-step-ahead prediction used  RNNs trained by backpropagation through-time 
algorithm \cite{bone2002multi}. A  review of single-output versus multiple-output 
approaches  showed  direct strategy 
more  promising choice over recursive strategy  \cite{BenTaieb2010}.    Multiple-output support vector regression (M-SVR)   achieved  
better  forecasts  when compared to   standard SVR using direct 
and iterated strategies \cite{Bao2014}. 
  
The  combination of recursive and direct strategies has also been prominent such as multiple SVR   models  that were trained 
independently 
based on the same training data and with different
targets \cite{zhang2013iterated}. Optimally pruned 
extreme learning machine (OP-ELM) used 
recursive, direct and a combination of the two strategies in an ensemble 
approach where the combination   gave better 
performance than  standalone methods \cite{Grigorievskiy2014}. Chandra et al. \cite{chandra2017CMTLMulti} presented recursive and cascaded neural networks inspired by multi-task learning  trained via cooperative neuroevolution where the tasks represented different prediction horizons. We note that neuroevolution provides an alternate training method that does not require gradients \cite{rawal2016evolving}. 
Ye and Dai \cite{YE2019227} presented a multitask learning method  which considers different prediction  horizons as  tasks and explores the relatedness amongst prediction horizons. The method consistently achieved lower error values over all horizons when compared to other related  iterative and direct prediction methods. 
 A  comprehensive study on  the  different strategies was given
using a large experimental
benchmark (NN5 forecasting competition) \cite{taieb2012review}, and further comparison for  macroeconomic time series. It 
was reported that the  iterated forecasts
typically outperformed the direct forecasts \cite{marcellino2006}. The relative performance of the iterated forecasts improved with the
forecast horizon, with further  comparison that  presented  an encompassing 
representation for  derivation  auto-regressive coefficients  
 \cite{Proietti2011}. A study on the properties  
shows that direct strategy provides prediction values that are relatively robust  and the benefits increases with the prediction horizon \cite{Chevillon2016}.

 The applications  for  real-world 
problems  include 1.)  auto-regressive models   for    predicting critical levels of abnormality in 
physiological
signals \cite{TranCyber2016}, 2.) flood forecasting 
using  recurrent neural networks \cite{Chen2013Flood,Chang2014}, 3.) emissions of 
nitrogen oxides using a neural network and related approaches 
\cite{Smrekar2013Nox}, 4.) photo-voltaic power forecasting using hybrid support 
vector machine  \cite{DeGiorgi2016}, 5.) Earthquake ground motions
and seismic response prediction \cite{Yang2016}, and 6.   central-processing unit (CPU) load prediction \cite{YANG20131257}. Recently, Wu \cite{WU2019} employed an adaptive-network-based fuzzy inference system  with uncertainty quantification  the prediction of short-term wind and wave conditions for marine operations. Wang and Li \cite{WANG2018429} used  multi-step ahead prediction for wind speed prediction which was based on optimal feature extraction,   LSTM networks, and an error correction strategy. The method  showed lower error values for one, three and five-step ahead predictions in comparison to  related methods. Wang and Li \cite{WANG2019296} also used hybrid strategy for  wind speed prediction with empirical wavelet transformation for feature extraction. Moreover, they used autoregressive fractionally integrated moving average   and  swarm-based   backpropagation neural network.


\subsection{Deep learning for time series prediction}

Deep learning has  been very successful for computer vision \cite{he2016deep},  computer games \cite{mnih2013playing}, multimedia, and big data related problems. Deep learning  methods have also been prominent for modelling temporal sequences \cite{lecun2015deep,schmidhuber2015deep}. RNNs   have been popular in forecasting time series with their ability to capture temporal information \cite{connor1994recurrent,husken2003recurrent, chandra2012cooperative,chandra2015competition,SALINAS2020}.    Mirikitani and Nikolaev   used \cite{mirikitani2010recursive}  variational  inference  for implementing Bayesian RNNs in order to provide  uncertainty quantification in predictions. CNNs have gained attention recently in  forecasting time series. Wang et al.  \cite{wang2017deep}  used CNNs with  wavelet transform for  probabilistic wind power forecasting. Xingjian et al.   \cite{xingjian2015convolutional} used   CNNs  in conjunction with LSTM networks to capture spatiotemporal sequences   for forecasting precipitation. Amarasinghe et al. \cite{Amarasinghe2017Deepelf} employed  CNNs for   energy load forecasting, and  Huang and Kuo \cite{Huang2018CNN-LSTM} combined CNNs and LSTM networks for air pollution quality forecasting. 
Sudriani et al. \cite{sudriani2019long} employed  LSTM networks for forecasting discharge level of a river for  managing water resources. Ding et al. \cite{Ding2015} employed  CNNs to evaluate   different events on stock price behavior,  and  Nelson et al. \cite{nelson2017stock} used LSTM networks to forecast stock market trends. Chimmula and Zhand  employed LSTM networks for forecasting COVID-19 transmission in Canada  \cite{CHIMMULA2020}.

\section{Methodology} 

\subsection{Data reconstruction }
   
The original time series data needs to be embedded (reconstructed) for    multi-step-ahead  
prediction. Taken's embedding theorem expresses that the
reconstruction can reproduce  important
features  of the original time series  \cite{Takens1981}. Therefore, given an 
observed time series $x(t)$, an embedded  phase
space $ Y(t) = [(x(t),
x(t-T),..., x(t-(D-1)T)]$ can be generated; where $T$ is the time delay, $D$ is
the embedding dimension (window size) given $t=  0,1,2,..., N-DT-1$, and $N$ is the length 
of the
original time series. A study needs to be done to determine optimal values for $D$ and $T$
 in order to efficiently apply Taken's theorem 
\cite{frazier2004}.   Taken's proved that if the original attractor is of
dimension $d$, then $D = 2d+1$ would be sufficient \cite{Takens1981}.    

 \subsection{Shallow learning via simple neural networks }
    
We refer to \textcolor{black}{the} backpropagation neural network and \textcolor{black}{multilayer perceptron}  as simple neural networks which has been typically trained by the stochastic gradient descent (SGD) algorithm. SGD maintains a single learning rate for all \textcolor{black}{the} weight updates which does not change during training. 
 The Adam optimiser \cite{kingma2014adam} extends  SGD by adapting the learning rate  for each parameter (weight)  as learning unfolds. Using first and second moments of the gradients, Adam computes adaptive learning rate, inspired by the \textit{adaptive gradient algorithm} (AdaGrad) \cite{duchi2011adaptive}.  In the literature, Adam has shown better results when compared to SGD and AdaGrad for a wide range of problems. In our experiments, we evaluate them further for multi-step ahead time series prediction.  Adam \textcolor{black}{optimiser updates} for the set of neural network parameters represented  by weights $w$ and bias $b$ for iteration   $t$ can be formulated as

\begin{eqnarray}
\Theta_{t-1}&=&[{w_{t-1},b_{t-1}}]\nonumber\\
g_t &=& \nabla_{\Theta}J_t(\Theta_{t-1})\nonumber\\
m_t &=& \beta_1.m_{t-1} + (1-\beta_1).g_t\nonumber\\
v_t &=& \beta_2.v_{t-1} + (1-\beta_2).g_t^2\nonumber\\
\hat{m}_t &=& m_t/(1-\beta^t_1)\nonumber\\
\hat{v}_t &=& v_t/(1-\beta^t_2)\nonumber\\
\Theta_t &=& \Theta_{t-1} -  \alpha.\hat{m}_t/(\sqrt[]{\hat{v}_t} + \epsilon)
\end{eqnarray}

 where, $m_t, v_t$ are the respective first  and second  moment vectors for iteration $t$; $\beta_1, \beta_2$ are constants $\in [0,1]$, $\alpha$ is the learning rate, and $\epsilon$ is a close to zero constant.

 \subsection{Simple RNN}
 
  The Elman RNN \cite{Elman_1990} is a prominent example of simple RNNs that  feature  a  context layer to act as memory  and incorporate current state for propagating information into future states to handle given future inputs. The   use \textcolor{black}{of context layer is to} store the output of the state neurons from computation of the previous time steps \textcolor{black}{making} them applicable for time-varying patterns in  data.  The context layer   maintains memory of the prior hidden layer result as shown in Figure 1. A vectorised formulation for simple RNNs is given as follows
 
  \begin{equation} 
  \begin{aligned} 
  h_t &= \sigma_h(W_{h} x_t + U_{h} h_{t-1} + b_h) \\
y_t &= \sigma_y(W_{y} h_t + b_y)
  \end{aligned}
 \end{equation}

 \noindent  where; $x_{t}$ input vector, $h_{t}$ hidden layer vector, $y_{t}$ output vector, $W$ represent the weights for hidden  and output layer,  $U$ is the context state weights,   $b$  is the bias,  and $\sigma _{h}$   and   $\sigma_y$ are the respective activation functions. Backpropagation through time (BPTT) \cite{Werbos_1990} has been \textcolor{black}{a} prominent method for training simple RNNs. In comparison to simple neural networks, BPTT in RNNs propagate the error for a deeper network architecture that features states defined by time.

 \begin{figure*}[tb]
  \begin{center}  
   \includegraphics[scale=0.25]{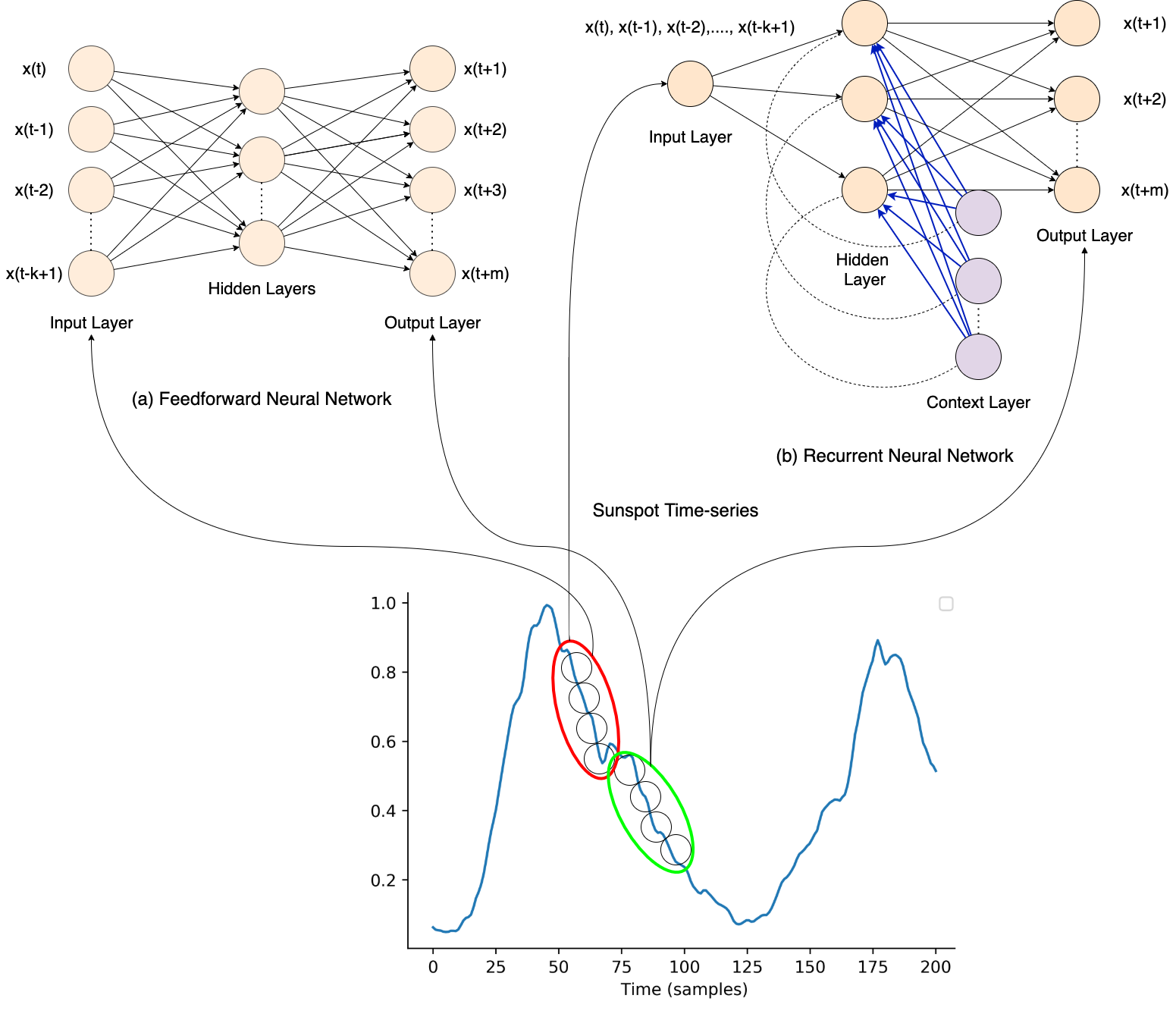} \\
   
   \label{fig:FNNRNN}
    \caption{ Feedforward neural network and Elman recurrent neural network for time series prediction. }
  \end{center}
\end{figure*}
 
 \subsection{LSTM  networks}
 
\begin{figure}[htbp!]
  \begin{center}  
   \includegraphics[scale=0.3]{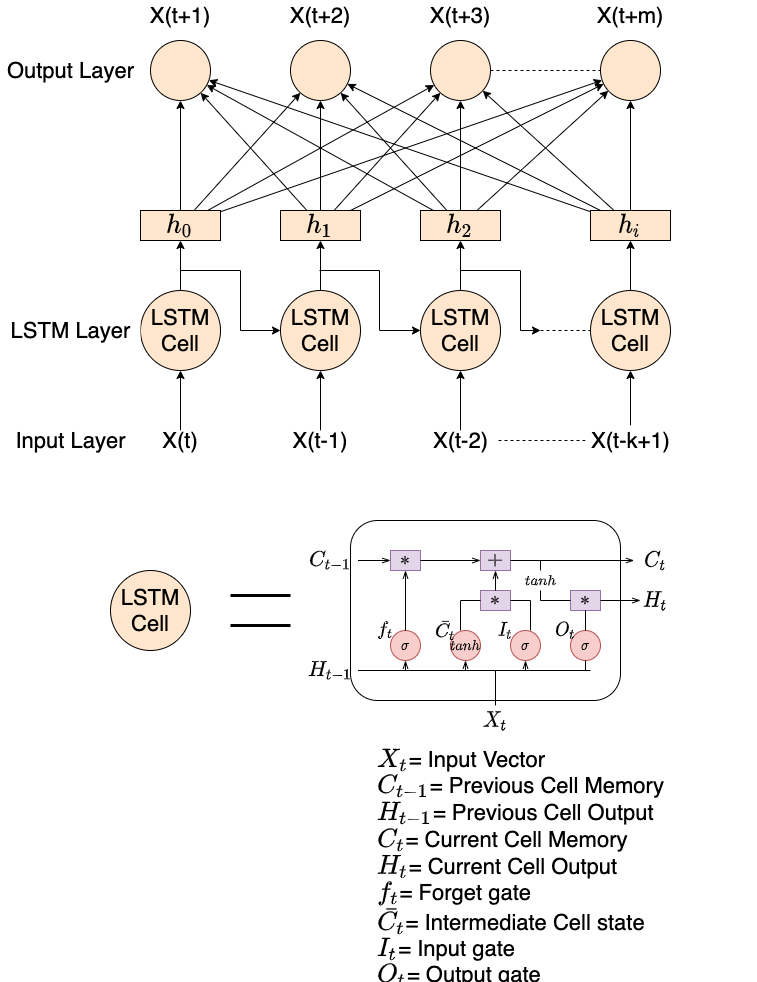} 
    \caption{ Long Short-Term Memory (LSTM)   networks. The LSTM cell denotes  memory cells that uses gates and cell memory for remembering long-term dependencies. 
  }
 \label{fig:LSTM}
  \end{center}
\end{figure}

 Simple RNNs have the \cite{hochreiter1997long} limitation of learning long-term dependencies with problems in vanishing and exploding gradients  \cite{hochreiter1998vanishing}. LSTM networks employ  memory cells and gates for much better capabilities in remembering the   long-term dependencies   in temporal sequences   as shown in Figure \ref{fig:LSTM}. LSTM units are trained in a supervised fashion on a set of training sequences  using an adaptation of the BPTT algorithm that considers the respective gates \cite{hochreiter1997long}.   LSTM networks calculate a hidden state  $h_{t}$ as
 \begin{equation}
 \begin{aligned}
    i_{t} & =\sigma\big(x_{t}U^{i}+h_{t-1}W^{i}\big)\\
    f_{t} & =\sigma\big(x_{t}U^{f}+h_{t-1}W^{f}\big)\\
    o_{t} & =\sigma\big(x_{t}U^{o}+h_{t-1}W^{o}\big)\\
    \tilde{C}_{t} & =\tanh\big(x_{t}U^{g}+h_{t-1}W^{g}\big)\\
    C_{t} & =\sigma\big(f_{t}\ast C_{t-1}+i_{t}\ast\tilde{C}_{t}\big)\\
    h_{t} & =\tanh(C_{t})\ast o_{t}
\end{aligned}
\end{equation}

 where,  $i_t$, $f_t$ and $o_t$ refer to the input, forget and output gates, at time $t$, respectively.  $x_t$ and  $h_t$ refer to the number of input features and number of hidden units, respectively. $W$   and $U$ is the weight matrices adjusted during learning along with  $b$ which is the bias.   The initial values are $c_{0}=0$ and   $h_{0}=0$. All the gates have the same dimensions $d_h$, the size of your hidden state. $\tilde{C}_t$ is a “candidate” hidden state, and  $C_t$ is the internal memory of the unit as shown in Figure \ref{fig:LSTM}.  Note that *  denotes element-wise multiplication.

  \subsection{Bi-directional LSTM networks }

A major shortcoming of conventional RNNs is that they 
only  make use of previous context state for determining future states. Bidirectional
RNNs (BD-RNNs) \cite{schuster1997bidirectional}  process information in both
directions with two separate hidden layers, which are then
propagated forward to the same output layer. BD-RNNs  consist of  placing two independent RNNs together to allow  both backward and forward information about the sequence at every time step. BD-RNN computes the forward hidden sequence $h_{f}$, the backward hidden sequence $h_{b}$, and the output sequence $y$ by iterating information from the backward layer, i.e. $t = T$ to $t = 1$. Then information in the other network is propagated  from $t =1$ to $t = T$  in order to update the output layer; when  both networks are combined, information is propagated in bidirectional manner.

\begin{figure}[htbp!]
  \begin{center}  
   \includegraphics[scale=0.3]{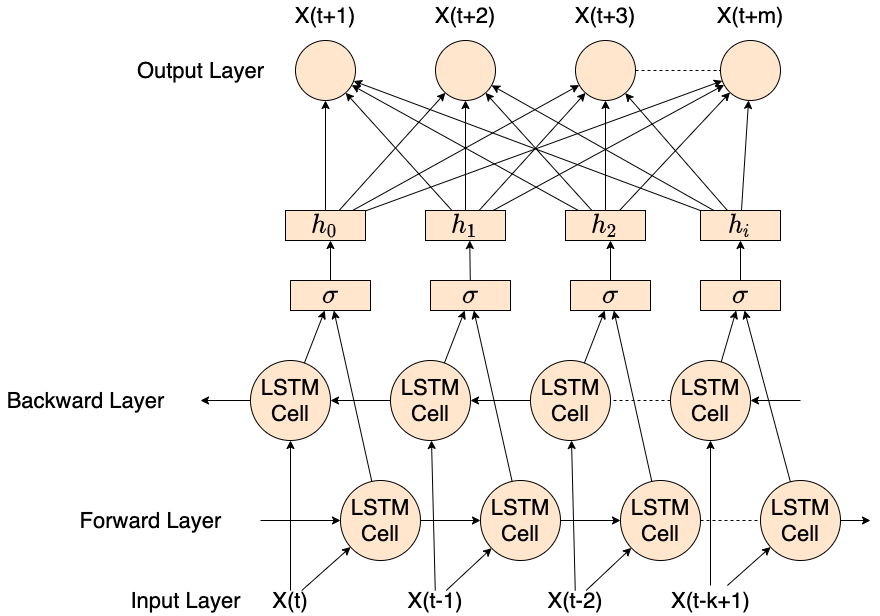} \\
    \caption{ Bi-directional LSTM network.}
 \label{fig:BDLSTM}
  \end{center}
\end{figure}

Bi-directional LSTM networks (BD-LSTM)  \cite{graves2005framewise} have been originally proposed for \textcolor{black}{word}-embedding in natural language processing in order to access long-range context or state in both  directions, similar to BD-RNNs.  
BD-LSTM would intake inputs in two ways, one from past to future and one from future to past which differ from conventional LSTM networks.  By running information backwards, state information from the future is preserved. Hence, with two hidden states combined, \textcolor{black}{at} any point in time the network can preserve information from both past and future as shown in Figure \ref{fig:BDLSTM}.
BD-LSTM networks have been used in several real-world sequence processing problems such as phoneme classification
\cite{graves2005framewise}, continuous speech recognition \cite{Fan2014TTSSW}, and speech synthesis \cite{graves2013hybrid}.

 \subsection{Encoder-Decoder LSTM networks}

\begin{figure}[htbp!]
  \begin{center}  
   \includegraphics[scale=0.3]{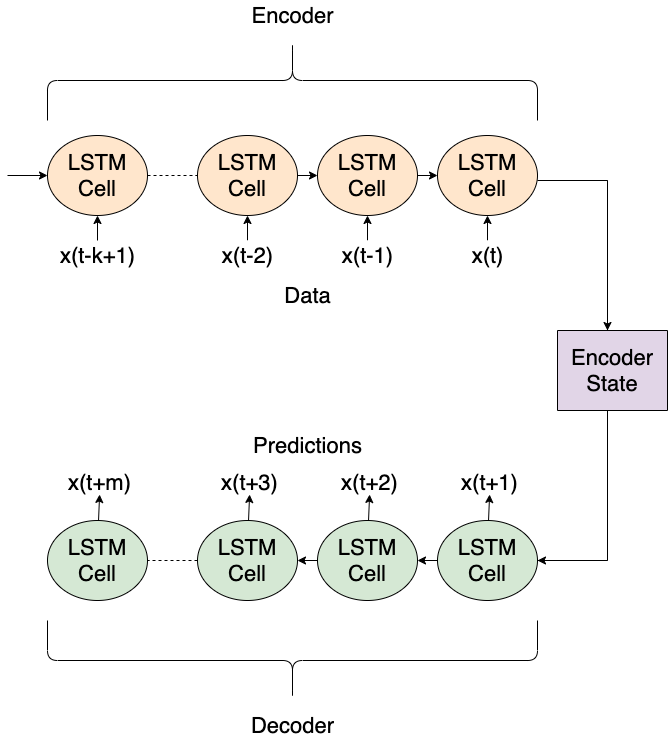} 
    \caption{ Encoder-Decoder LSTM network. }
 \label{fig:EN-DC LSTM}
  \end{center}
\end{figure}

 Sutskever et al. \cite{NIPS2014_5346} introduced the encoder-decoder LSTM network (ED-LSTM) which is a sequence to sequence model for mapping a fixed-length input to a fixed-length output \cite{cho2014}. The length of the input and output may differ which makes them applicable in automatic language translation tasks (such as English to French). Hence,  the input can be  the sequence of video frames
$(x_1, . . . , x_n)$, and the output is the sequence of words
$(y_1, . . . , y_m)$. Therefore, we estimate the conditional probability of
an output sequence $(y_1, . . . , y_m)$, given an input sequence
$(x_1, . . . , x_n)$, i.e.
$p(y_1, . . . , y_m|x_1, . . . , x_n)$. In the case of multi-step series prediction,  both the input and outputs are of
variable lengths. ED-LSTM networks  handle variable-length input and outputs by  first encoding the input sequences one at a time,
 using a latent vector representation,
and then decoding \textcolor{black}{them} from that representation. 
In the encoding phase, given an input sequence, the ED-LSTM computes a sequence of hidden
states\textcolor{black}{.} In \textcolor{black}{the} decoding phase, it defines a distribution over the output sequence   given the input sequence  as shown in Figure \ref{fig:EN-DC LSTM}.

\subsection{CNNs} 

\begin{figure*}[htbp!]
\centering 
   \includegraphics[scale = 0.25] {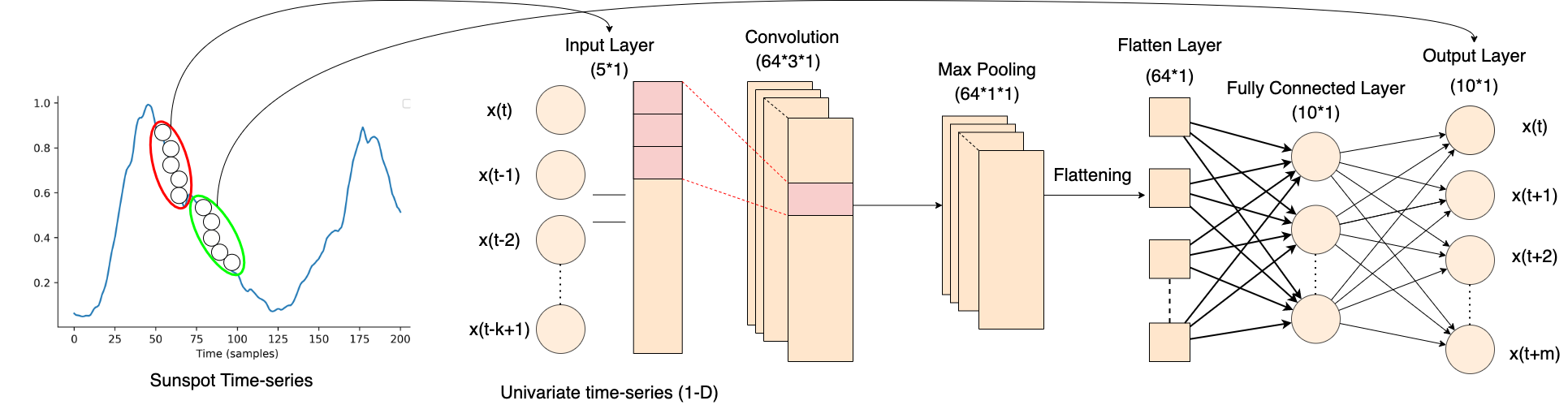}

\caption{ One-dimensional convolutional neural network for multi-step ahead time series prediction. 
}
 \label{fig:cnn}
\end{figure*}

 CNNs  introduced by LeCun \cite{lecun1990cnn,lecun1998cnn} are  prominent deep learning architecture inspired by the natural visual system of mammals. CNNs can be trained using backpropagation algorithm for tasks such as handwritten digit classification \cite{Hecht1989backprop}. CNNs have been prominent in many computer vision and image processing tasks. Recently, CNNs have been applied for time series prediction and produced very  promising results  \cite{Amarasinghe2017Deepelf,xingjian2015convolutional,wang2017deep}. CNNs   learn spatial hierarchies of features by using multiple building blocks, such as convolution, pooling layers, and fully connected layers.   Figure  \ref{fig:cnn} shows an example of a CNN used for time series prediction using a  univariate time series as input where multiple output neurons represent different prediction horizons. We note that CNNs are more appropriate for multivariate time series  with use of features extracted via the convolutional and the pooling layers.

 \section{Experiments and Results}

\subsection{Experimental Design}

 We use a combination of benchmark problems that include  simulated and real-world time series. The 
simulated time series are Mackey-Glass\cite{Mackey1977}, Lorenz 
\cite{lorenz1963}, Henon \cite{Henon1976}, and 
Rossler \cite{rossler}. The real-world time series are Sunspot  
\cite{Sunspot2001}, Lazer \cite{weigendtime} and ACI-financial time 
series \cite{timeDataSet}. They  have been used in our previous works and have been prominent for time series problems \cite{ChandraLangevinNC2019,ChandraTNNLS2015,chandra2017_CMTL}. 
The Sunspot 
time series indicates solar activities from November 1834 to June 2001 and consists of 2000 data 
points \cite{Sunspot2001}. The ACI-finance time series  contains
closing stock prices from December 2006 to February 2010,
featuring 800 data points \cite{timeDataSet}. The  
Lazer time 
series is from the  \textit{Santa Fe 
competition} that consists of 500 points \cite{weigendtime}.

The respective time series are processed into a state-space 
vector \cite{Takens1981} with embedding dimension
$D=5$ and time-lag  $T=1$ for 10-step-ahead prediction.
We determine respective  model hyper-parameters from trial experiments that include    number of hidden 
neurons, and learning rate. Table \ref{tab:config} gives details for the topology of the respective models in terms of input, hidden and output layers. We use maximum time of 1000 epochs with rectifier linear units (Relu)   in all the respective models.  The simple neural networks  feature SGD and Adam optimiser (FNN-SGD and FNN-Adam). Adam optimiser is used in the deep learning models that include simple RNNs,  LSTM networks, ED-LSTM, BD-LSTM, and CNNs. 

The  time series  are  scaled in the range [0,1].  We used first 1000 data points  from which the first 60\% are used for training and remaining for testing. We use the root-mean-squared error (RMSE)   as the main performance measure ( Equation \ref{rmse})
for the different prediction horizons  
\begin{equation} 
RMSE =  \sqrt{\frac{ 1}{N }   \sum_{i=1}^{N} (y_i - \hat{y}_i)^2}
\label{rmse}  
\end{equation}

\noindent where, $y_i, \hat{y}_i$   are the observed data,
predicted data, respectively. $N$ is the length of the observed data.

\begin{table*}[htbp!]
 \small 
 \centering
 \caption{Configuration of models }
\label{tab:config}
\begin{tabular}{llllp{11cm}}
\hline
 &  Input & Hidden Layers & Output & Comments  \\
\hline
\hline
FNN-Adam &      5&1&10& Hidden layer =  10 neurons\\
FNN-SGD & 5 & 1&10& Hidden layer = 10 neurons\\

LSTM &      5&1&10& Hidden layer = 10 cells \\
BD-LSTM &      5&1&10&Forward and backward layer = 10 cells each \\
ED-LSTM&      5&4&10& Two LSTM networks with a time distributed layer \\
RNN &      5&2&10& Hidden layer = 10 neurons \\
CNN &      5&4&10&  Filters=64, convolutional window size=3, and max-pooling window size=2\\ 

\hline &
\end{tabular}

\end{table*}

\subsection{Results}

 We report the mean and 95 \% confidence interval of RMSE for each prediction horizon for the respective problem  for train and test datasets from 30 experimental runs with different initial neural network weights. Figure \ref{fig:henon} to Figure  \ref{fig:rossler} presents the results for the simulated time series  (Tables \ref{tab:henon} to \ref{tab:rossler} in Appendix). Figure \ref{fig:finance} to \ref{fig:lazer} presents the results for the real-world time series   (Tables \ref{tab:finance} to \ref{tab:lazer} in Appendix). We define robustness as the confidence interval which must be as low as possible to indicate high confidence in prediction. We consider scalability as the ability to provide consistent performance as the the prediction horizon increases. The results are given in terms of the RMSE where the lower values indicate better performance. Each problem reports 10-step-ahead prediction results  with RMSE mean and 95\% confidence interval  as  error bars, shown in Figures \ref{fig:finance} to \ref{fig:rossler}.

  We first review results for real-world time series that  feature noise (ACI-Finance, Sunspot, Lazer). Figure \ref{fig:finance} shows the results for the ACI-fiancee problem. We observe that the test performance is better than the train performance in Figure \ref{fig:finance} (a), where deep learning models provide more reliable performance. The prediction error (RMSE) increases with the prediction horizon, and the deep learning methods do much better than simple neural networks (FNN-SGD and FNN-Adam). We find that LSTM provides the best overall   performance as shown in Figure \ref{fig:finance} (b).  The overall test performance shown in Figure \ref{fig:finance} (a) indicates that FNN-Adam  and  LSTM  provide similar performance, which are better than rest of the problems.  Figure \ref{fig:financeBest} shows ACI-finance prediction performance of the best experiment run with selected prediction horizons that indicate how the prediction deteriorates  as prediction horizon increases. 

Next, we consider the results for the  Sunspot time series  shown in Figure \ref{fig:sunspot} which follows a similar trend as the ACI-finance problem in terms of the increase in prediction error along with the prediction horizon. The test performance is better than the train performance as evident from Figure \ref{fig:sunspot} (a). The LSTM methods (LSTM, ED-LSTM, BD-LSTM) gives better performance than the other methods as can be observed from Figure \ref{fig:sunspot} (a) and \ref{fig:sunspot} (b). Note that the FNN-SGD gives the worst performance and the performance of RNN is better than that of CNN, FNN-SGD, and FNN-Adam, but poorer than LSTM methods. Figure \ref{fig:sunspotsingle}  shows Sunspot prediction performance of the best experiment run with selected prediction horizons.

The results for  Lazer time series  is shown in Figure \ref{fig:lazer}, which exhibits a similar trend in terms of the train and test performance as the other real-world time series problems. Note that the Lazer problem is highly chaotic (as visually evident in Figure \ref{fig:lazersingle}), which seems to be the primary reason behind the difference in performance for the prediction horizon in contrast to other problems as displayed in Figure \ref{fig:lazer} (b). It is striking that none of the methods appear to be showing any trend for the prediction accuracy along the prediction horizon, as seen in previous problems. In terms of scalability, all the methods appear to be performing better in comparison with the other problems. The performance of CNN is better than that of RNN, which is different from other real-world time series. Figure \ref{fig:lazersingle}  shows Lazer prediction performance of the best experiment run using ED-LSTM with selected prediction horizons. We note that due to the chaotic nature of the time series, the prediction performance is visually not clear.

We now consider simulated time series that do not feature noise (Henon, Mackey-Glass, Rosssler, Lorenz). The Henon time series in  Figure \ref{fig:henon}  shows that ED-LSTM provides the best performance. Note that there is a more significant difference between the three LSTM methods   when compared to other problems. The trends are similar to the ACI-finance  and the Sunspot problem  given the prediction horizon performance in Figure \ref{fig:henon} (a) and \ref{fig:henon} (b), where the simple neural networks (FNN-SGD and FNN-Adam) appear to be more scalable than the other methods along the prediction horizon, although they perform poorly.  Figure \ref{fig:mackeysingle}  and Figure \ref{fig:henonsingle} show Mackey-Glass and Henon prediction  performance of the best experiment run using ED-LSTM for selected prediction horizons. The Henon prediction in Figure \ref{fig:henonsingle}  indicates that it is far more  chaotic than Mackey-Glass; hence, it  faces more challenges. We show them since these are cases with no noise when compared to real-world time series previously shown. They have a larger   deterioration in prediction performance as the prediction horizon increases (Figures \ref{fig:financeBest} and Figure \ref{fig:sunspotsingle}). 

In the Lorenz, Mackey-Glass and Rossler simulated time series, the deep learning methods are performing far better than the simple neural networks as shown in Figures \ref{fig:lorenz}, \ref{fig:mackey} and \ref{fig:rossler}. The trend along the prediction horizon is similar to previous problems, i.e., the prediction error increases along with the prediction horizon. If we consider scalability, the deep learning  methods are more scalable in the Lorenz, Mackey-Glass and Rossler problems than the previous problems.   This is the first instance where the CNN has outperformed LSTM for Mackey-Glass and Rossler time series.


 We note that there have been distinct trends in prediction for the different types of problems. In the simulated time series, given that we exclude Henon,  we observe a similar trend for Mackey-Glass, Lorenz and Rossler time series. The trend indicates  that simple neural networks face major difficulties.   ED-LSTM and BD-LSTM networks provides the best performance, which also applies to Henon time series, except that it has close performance for simple neural networks when compared to deep learning models for 7-10  prediction horizons (Figure 9 b). This difference  reflects in the nature of the time series which is  highly chaotic in nature (Figure 15). We further note that in the case of the simple neural networks,   Henon   (Figure 9) does not   deteriorate in performance as the prediction horizon increases when compared to Mackley-Glass, Lorenz and Rossler problems. Simple neural networks in this case performs poorly from the beginning prediction horizon.  
 
 The performance of simple neural networks in Lazer problem shows a similar trend in Lazer time series, where the predictions are poor from the beginning and its striking that LSTM networks actually improve the performance as the prediction horizon increases (Figure 8 b). This trend is a clear \textcolor{black}{outlier} when compared to \textcolor{black}{the} rest of real-world and simulated problems, since they all show deep learning models deteriorate as the prediction horizon \textcolor{black}{increases}.

\begin{figure*}[htbp!]
\centering
\subfigure[RMSE  across 10 prediction horizons]{
\includegraphics[scale =0.55]{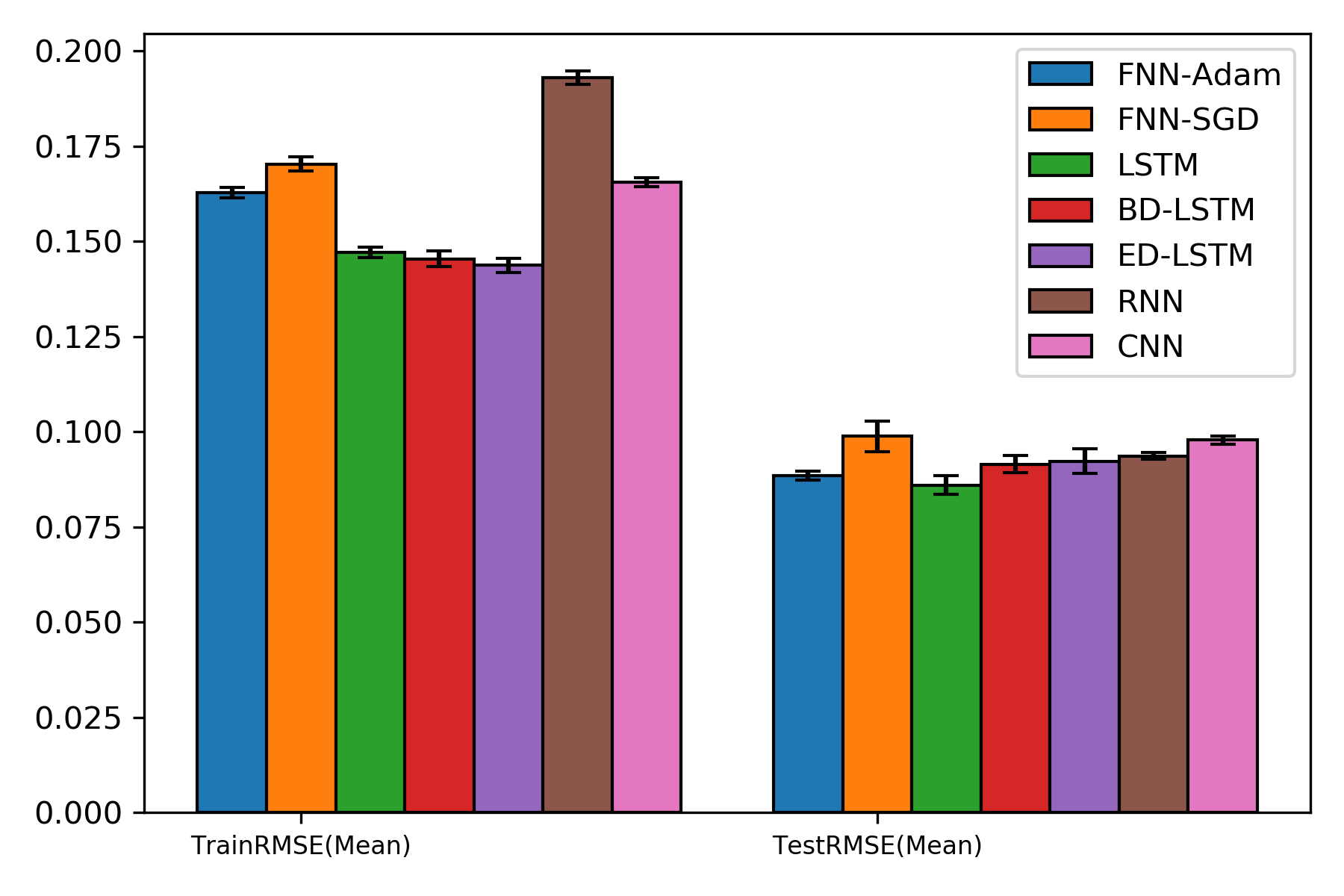}
 }
 \subfigure[10 step-ahead prediction]{
   \includegraphics[scale =0.55] {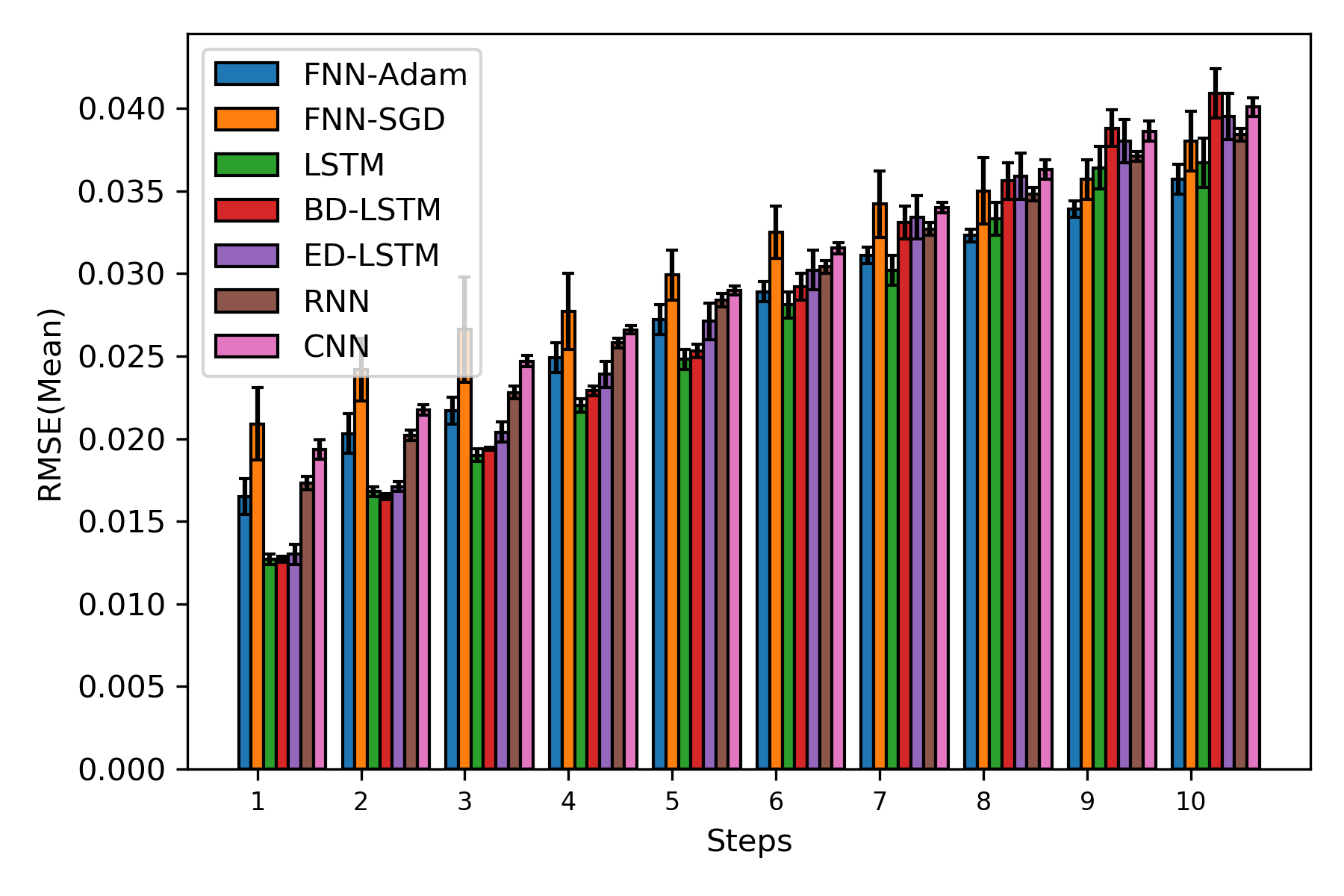}
 }
\caption{ACI-finance time series: performance evaluation of respective methods (RMSE mean and 95\% confidence interval as error bar).}
\label{fig:finance}
\end{figure*}

\begin{figure*}[htbp!]
\centering
\subfigure[RMSE  across 10 prediction horizons]{
\includegraphics[scale =0.55]{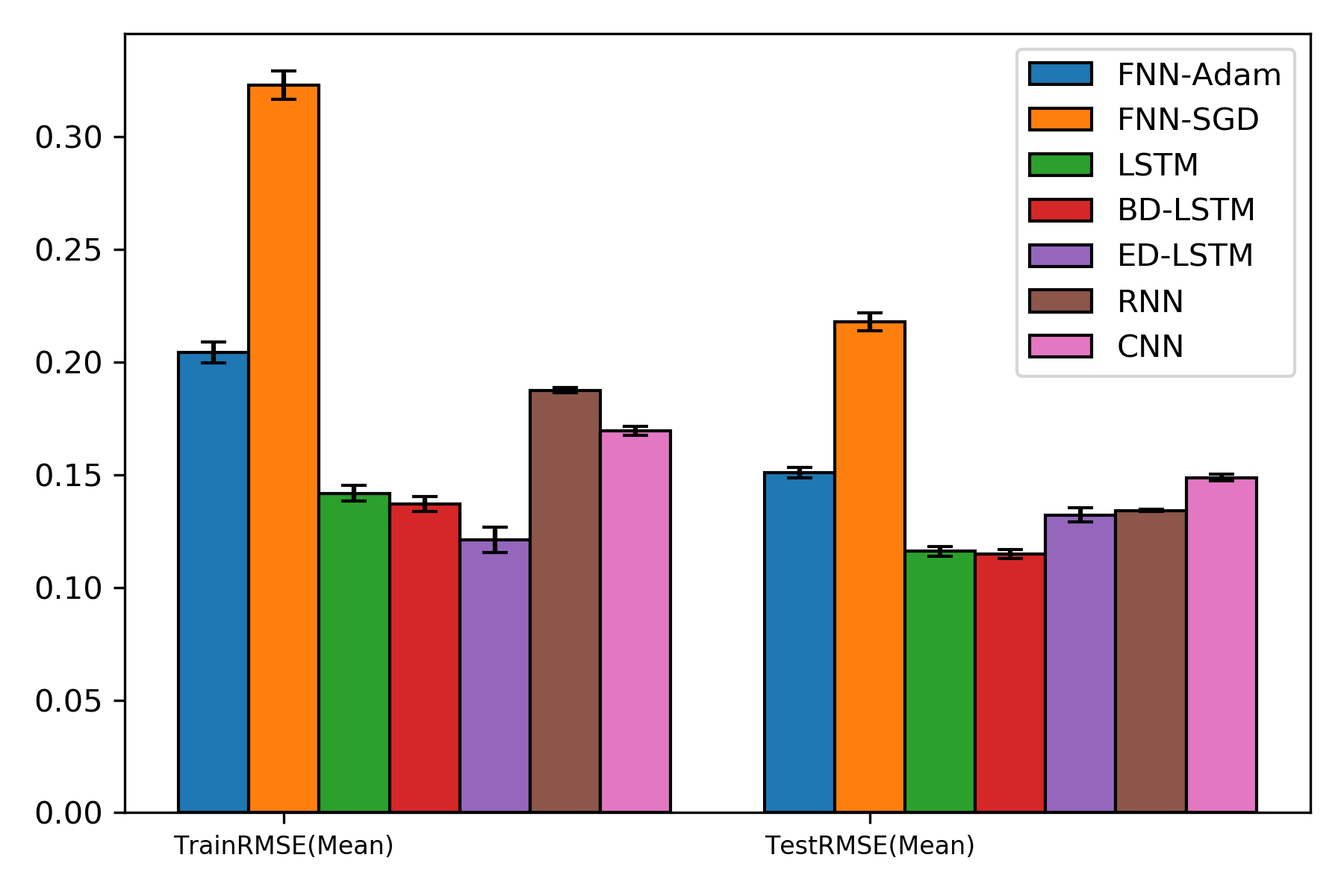}
 }
 \subfigure[10 step-ahead prediction]{
   \includegraphics[scale =0.55] {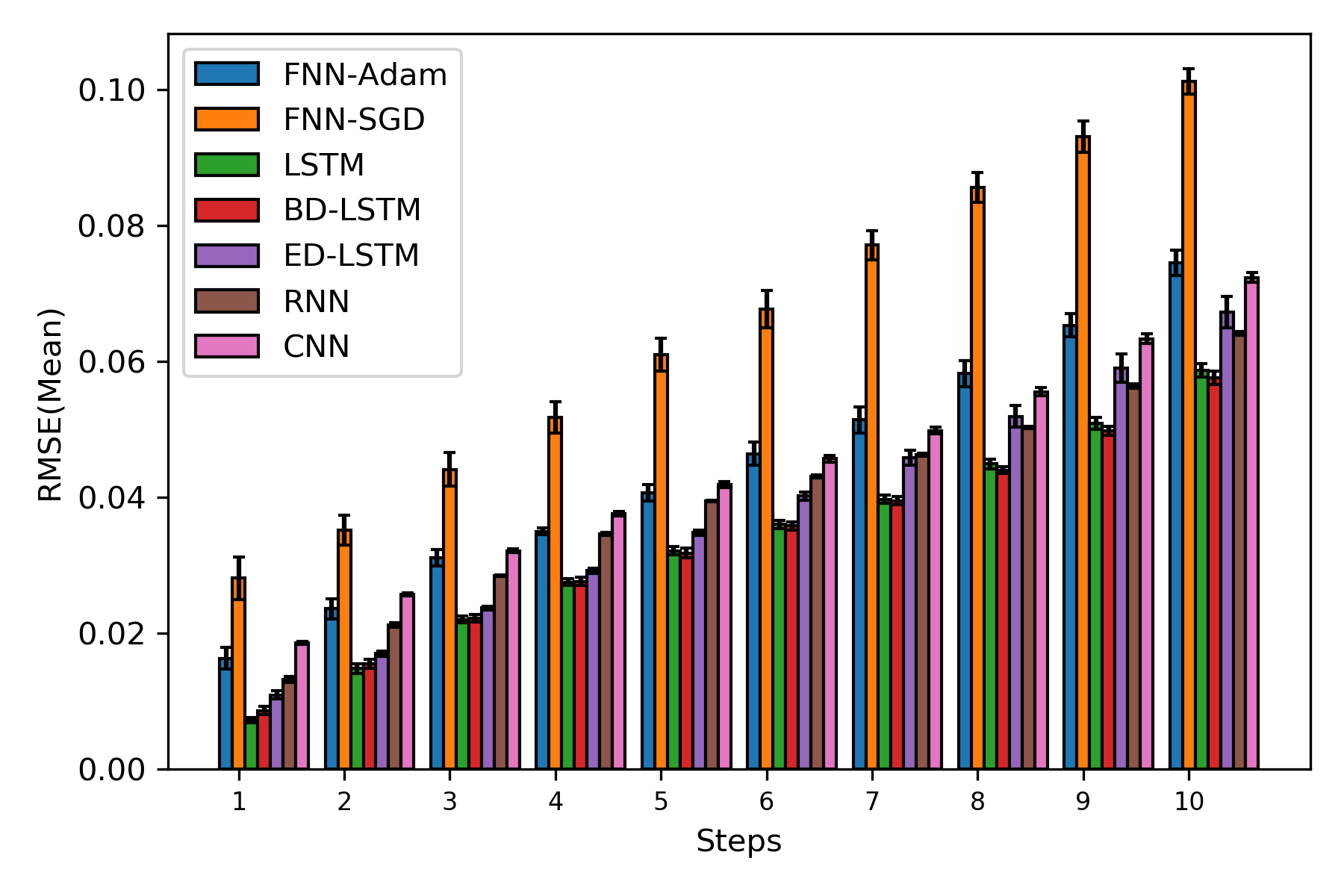}
 }
\caption{Sunspot time series: performance evaluation of respective methods (RMSE mean and 95\% confidence interval as error bar).}
\label{fig:sunspot}
\end{figure*}

\begin{figure*}[htbp!]
\centering
\subfigure[RMSE  across 10 prediction horizons]{
\includegraphics[scale =0.55]{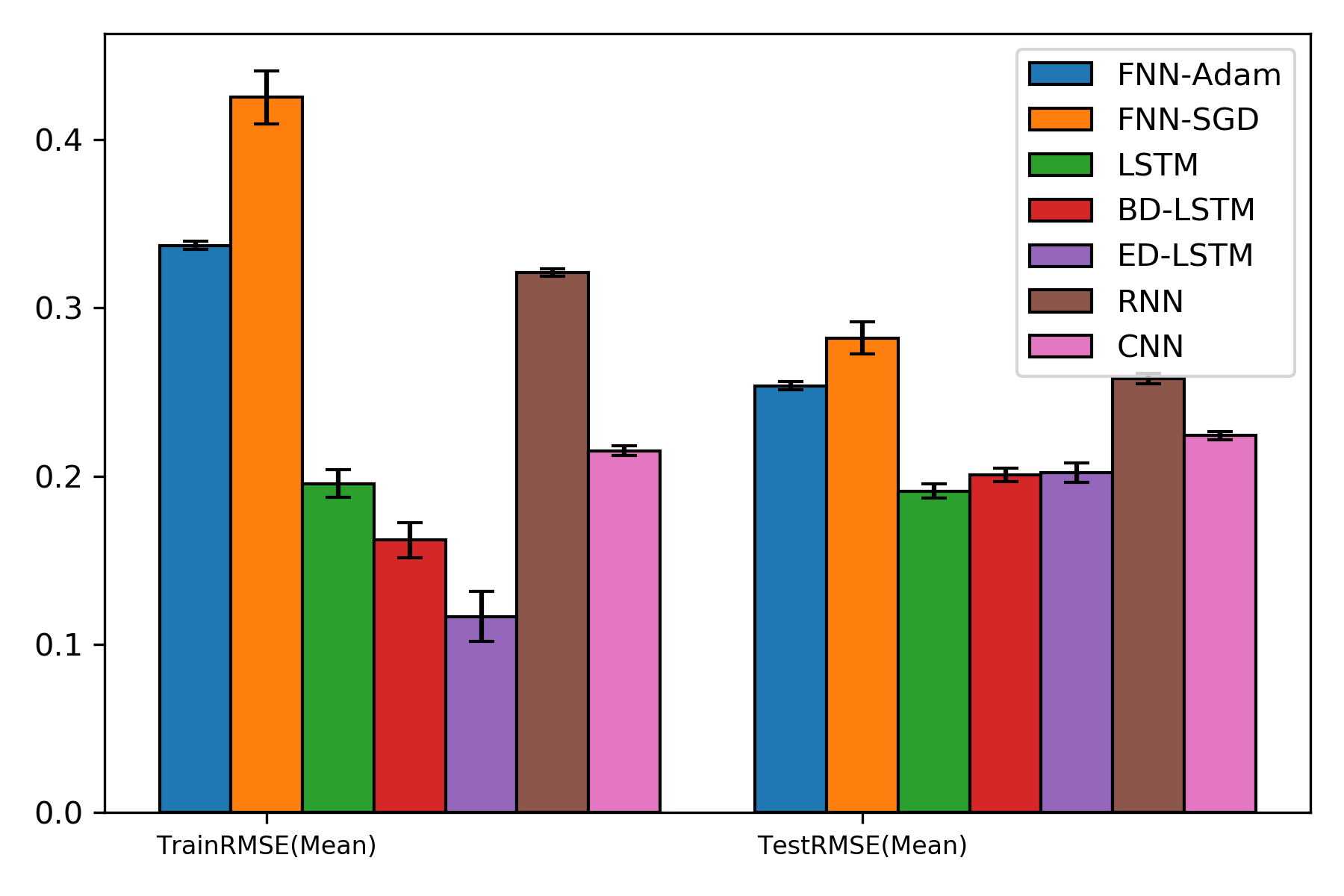}
 }
 \subfigure[10 step-ahead prediction]{
   \includegraphics[scale =0.55] {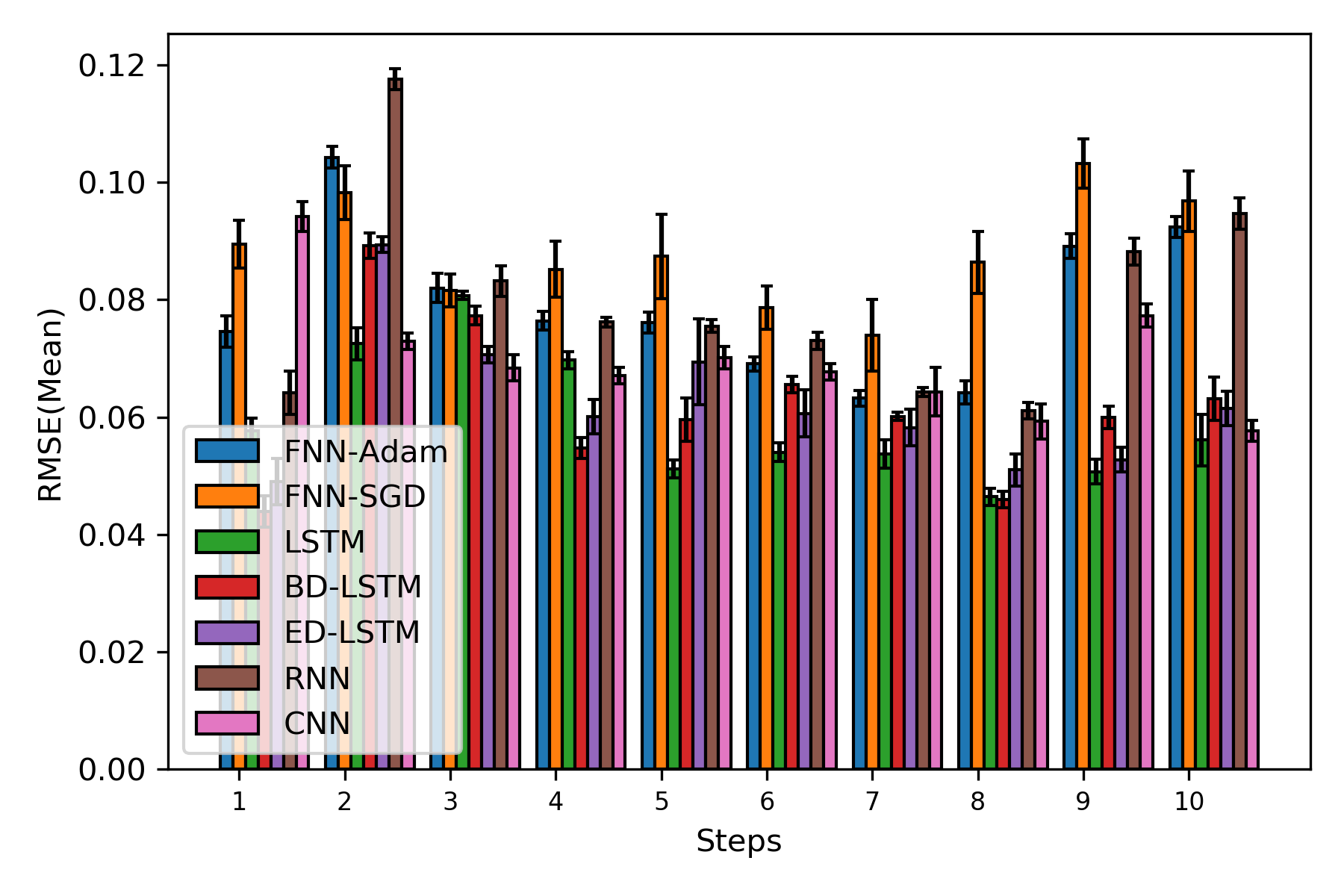}
 }
\caption{Lazer time series: performance evaluation of respective methods (RMSE mean and 95\% confidence interval as error bar).}
\label{fig:lazer}
\end{figure*}

\begin{figure*}[htb]
\centering
\subfigure[RMSE-Mean]{
\includegraphics[scale =0.55]{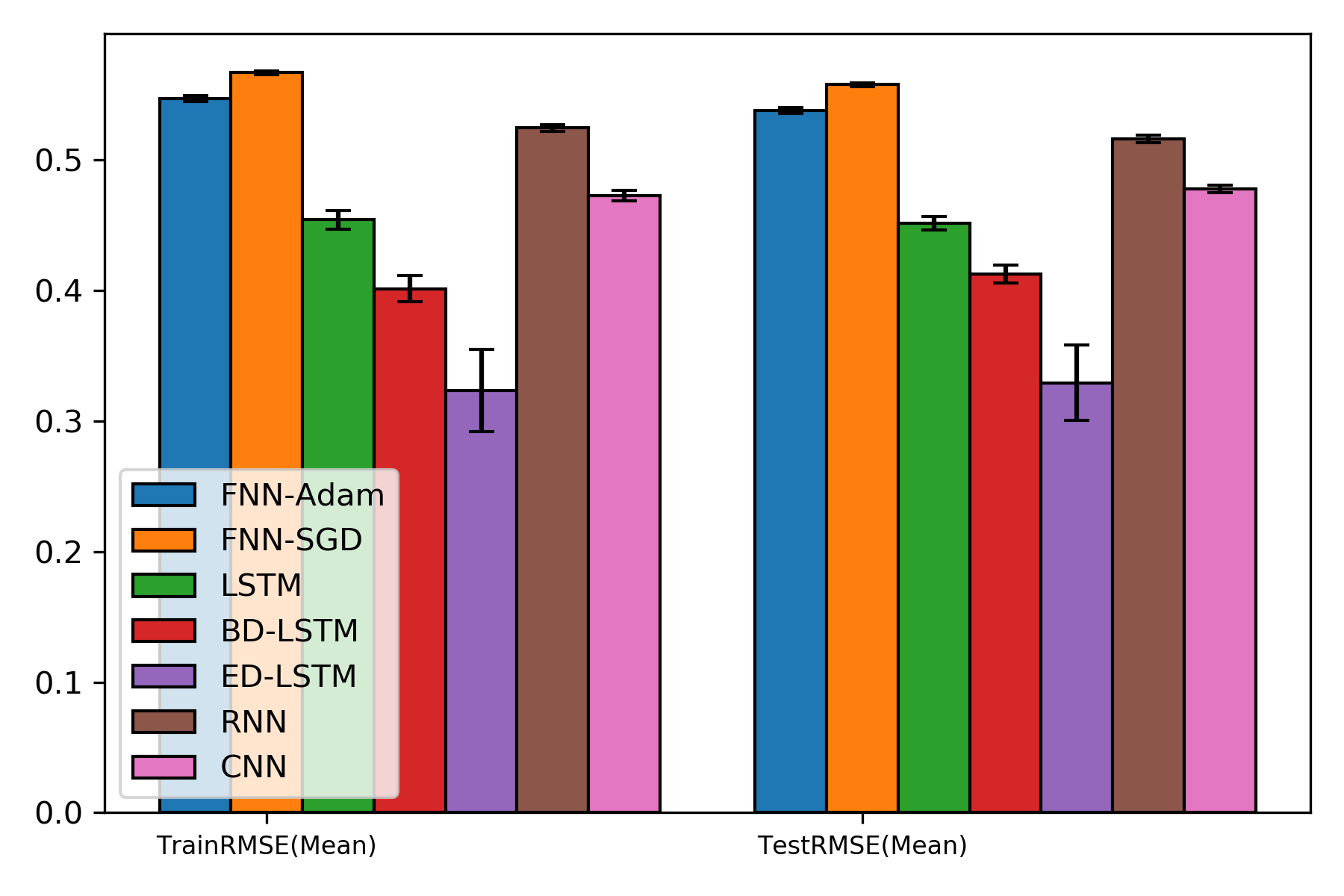}
 }
 \subfigure[10 step-ahead prediction]{
   \includegraphics[scale =0.55] {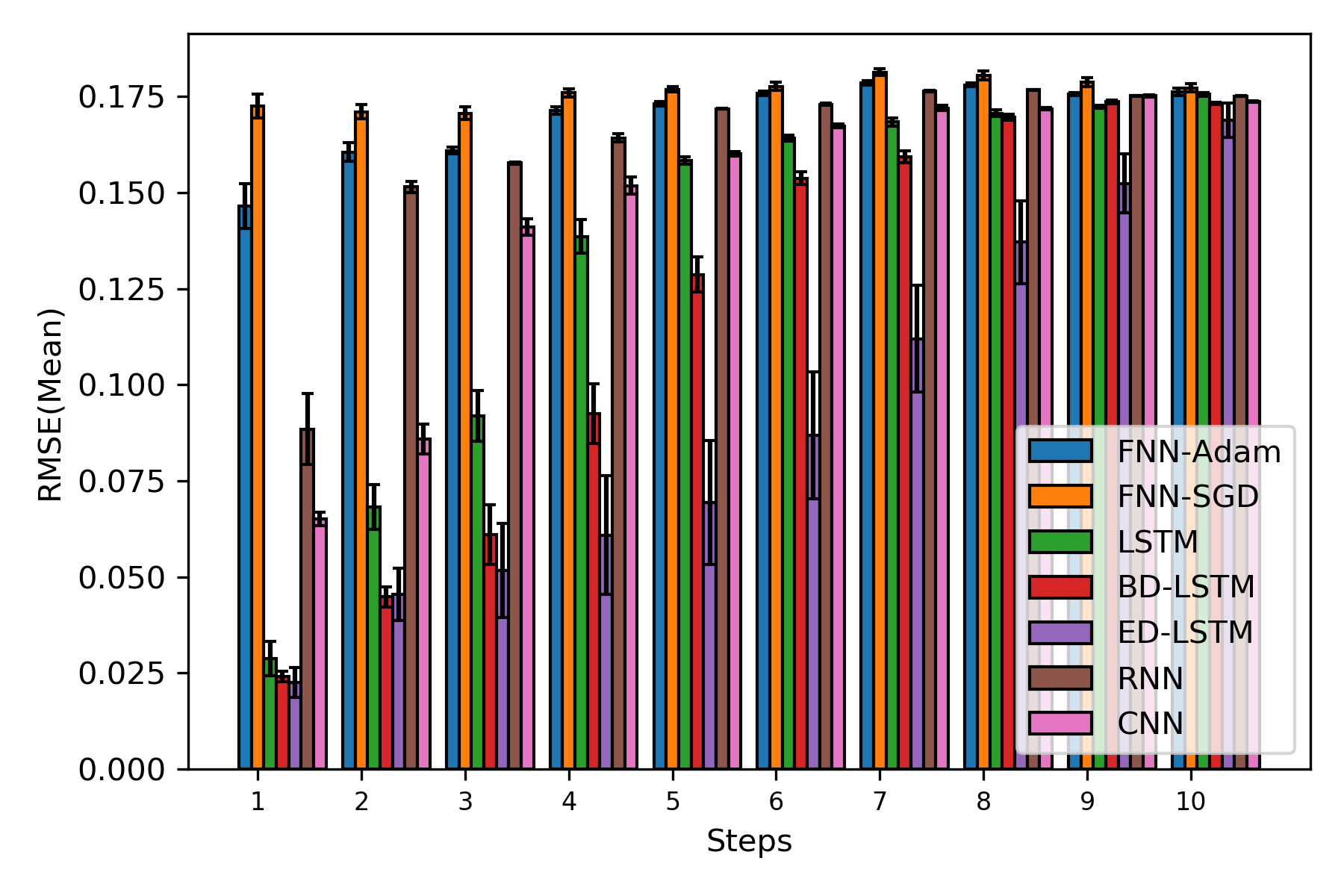}
 }
\caption{Henon time series: performance evaluation of respective methods (RMSE mean and 95\% confidence interval as error bar).}
\label{fig:henon}
\end{figure*}

\begin{figure*}[htbp!]
\centering
\subfigure[RMSE  across 10 prediction horizons]{
\includegraphics[scale =0.55]{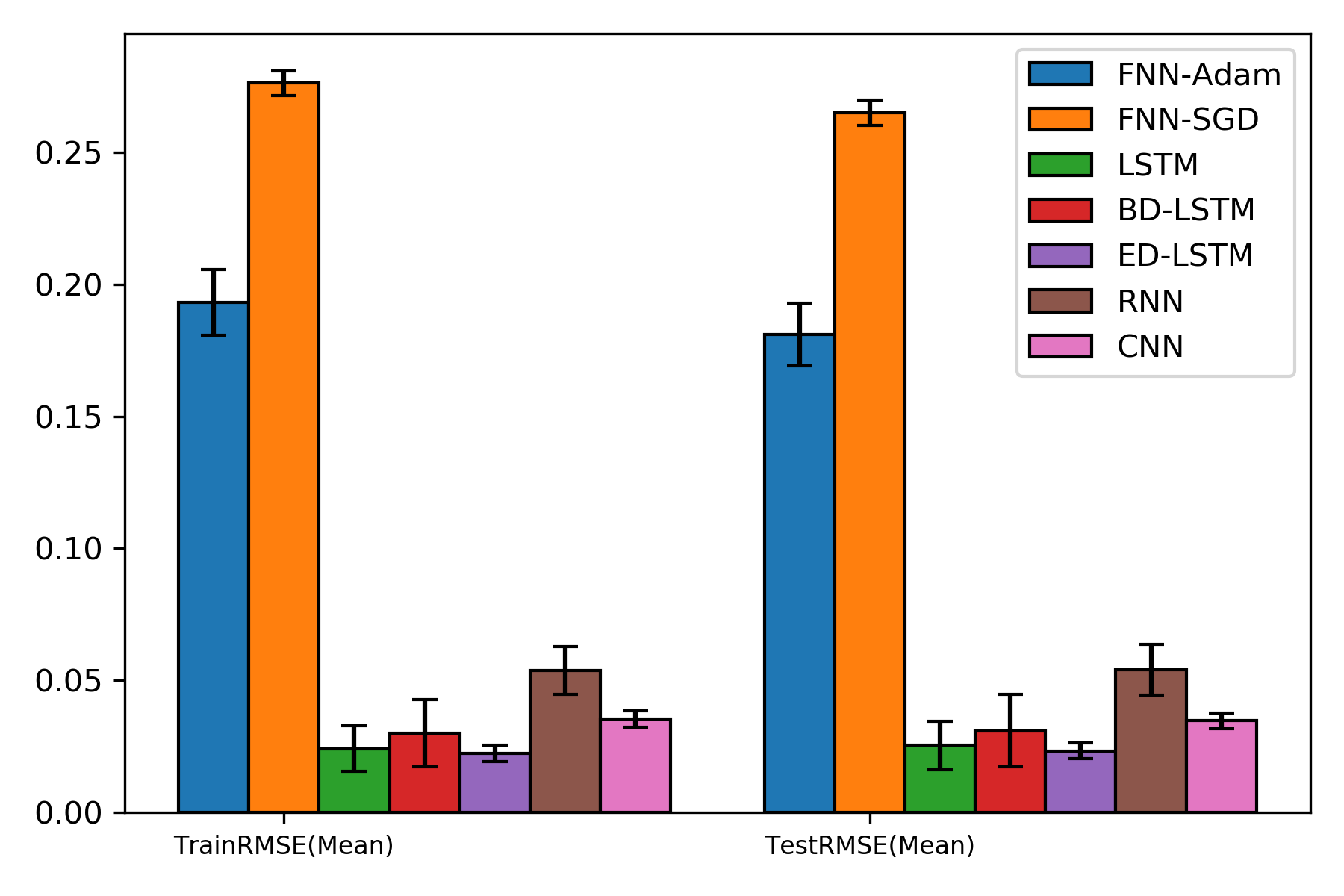}
 }
 \subfigure[10 step-ahead prediction]{
   \includegraphics[scale =0.55] {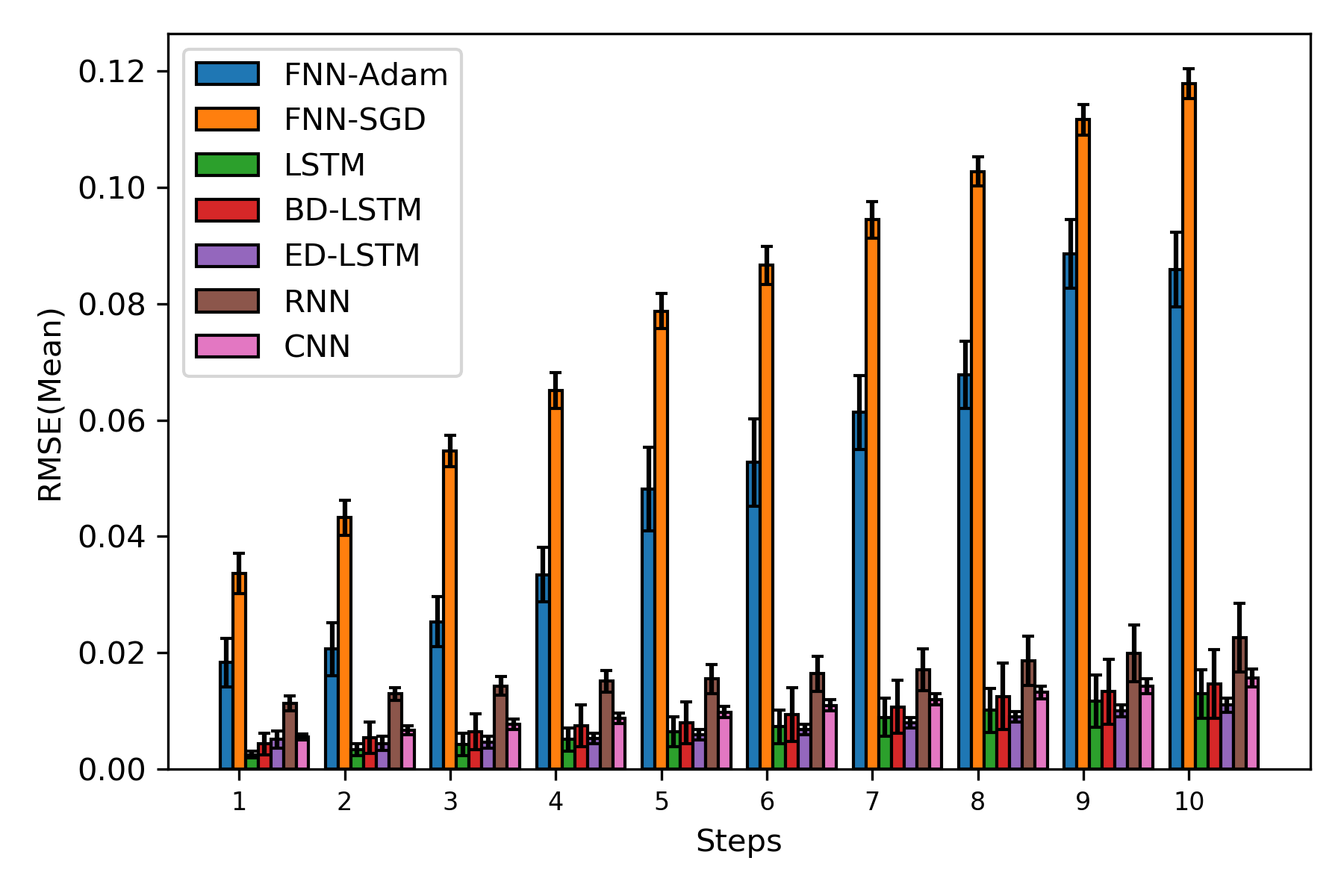}
 }
\caption{Lorenz time series: performance evaluation of respective methods (RMSE mean and 95\% confidence interval as error bar).}
\label{fig:lorenz}
\end{figure*}

\begin{figure*}[htbp!]
\centering
\subfigure[RMSE  across 10 prediction horizons]{
\includegraphics[scale =0.55]{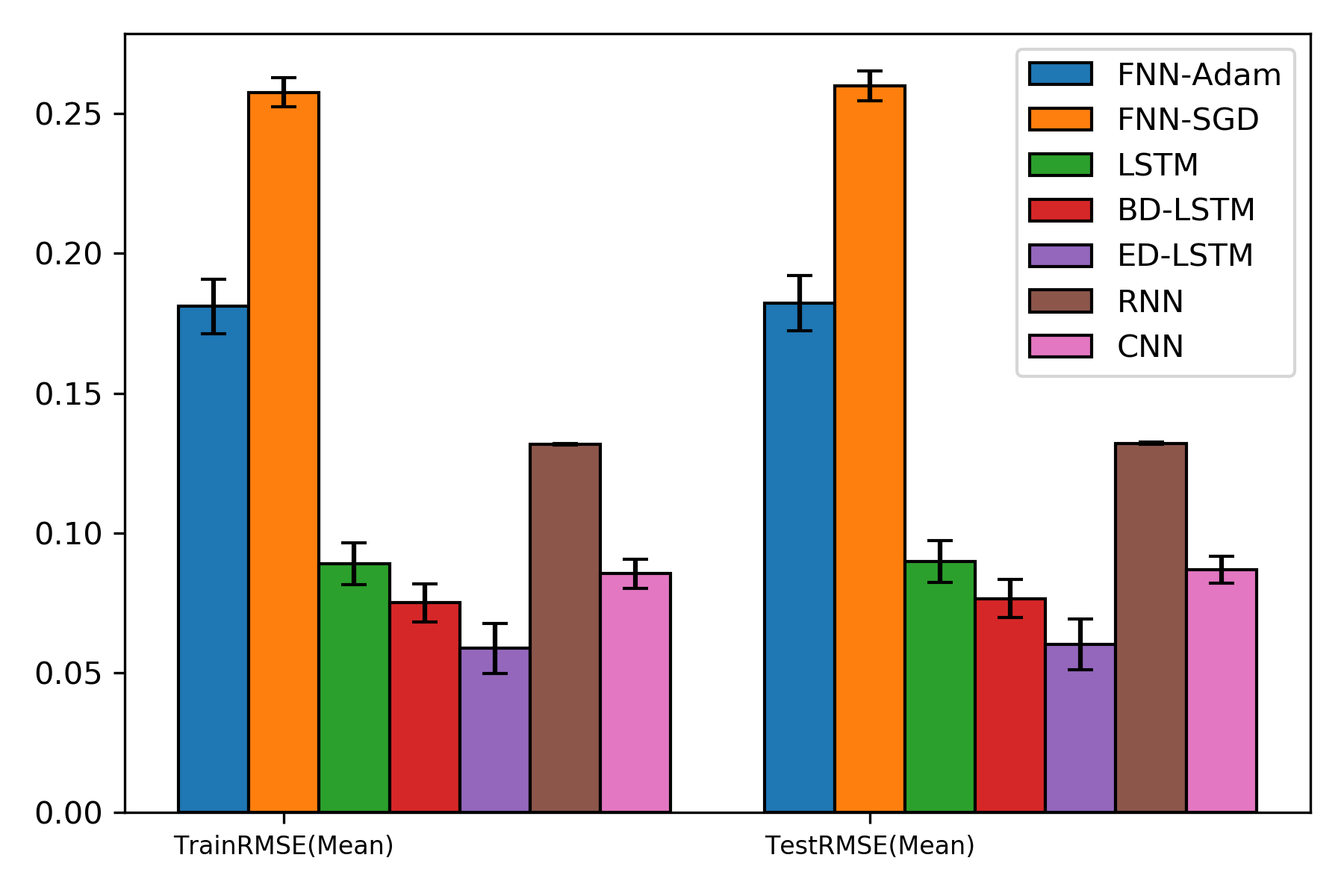}
 }
 \subfigure[10 step-ahead prediction]{
   \includegraphics[scale =0.55] {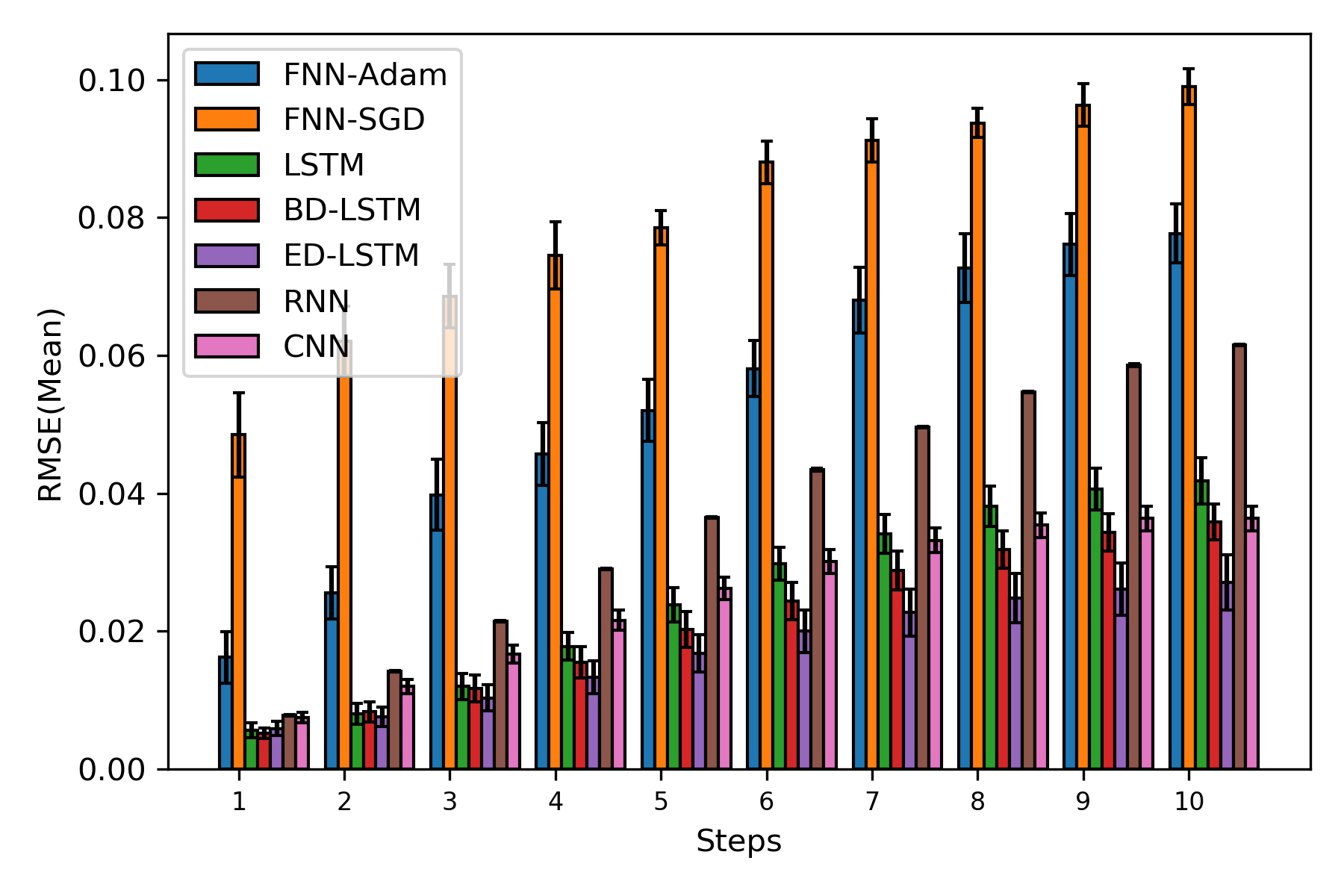}
 }
\caption{Mackey-Glass time series: performance evaluation of respective methods (RMSE mean and 95\% confidence interval as error bar).}
\label{fig:mackey}
\end{figure*}

\begin{figure*}[htbp!]
\centering
\subfigure[RMSE  across 10 prediction horizons]{
\includegraphics[scale =0.55]{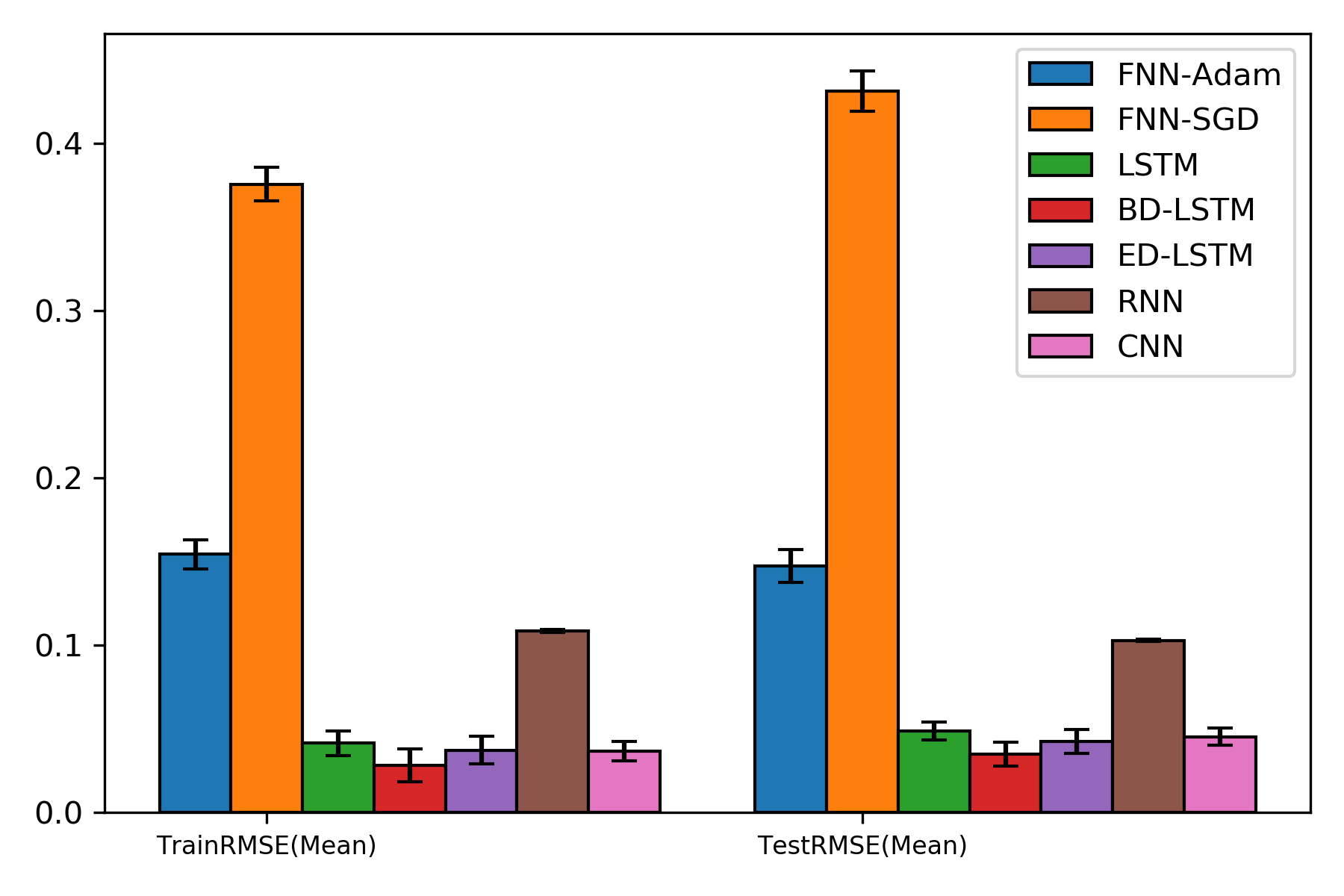}
 }
 \subfigure[10 step-ahead prediction]{
   \includegraphics[scale =0.55] {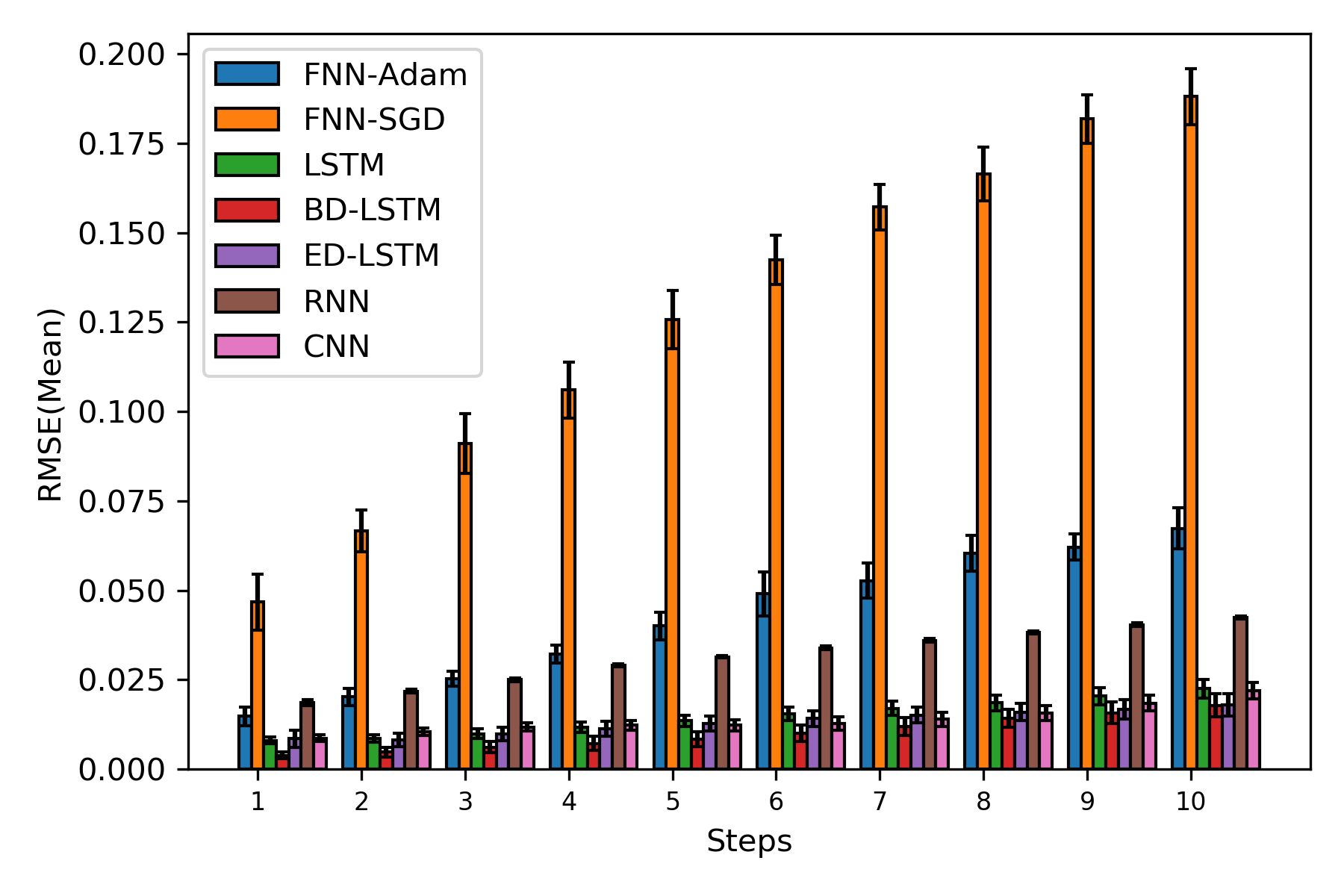}
 }
\caption{Rossler time series: performance evaluation of respective methods (RMSE mean and 95\% confidence interval as error bar).}
\label{fig:rossler}
\end{figure*}

\begin{figure*}[htb]
\centering
\subfigure[Step 1]{
\includegraphics[scale =0.13]{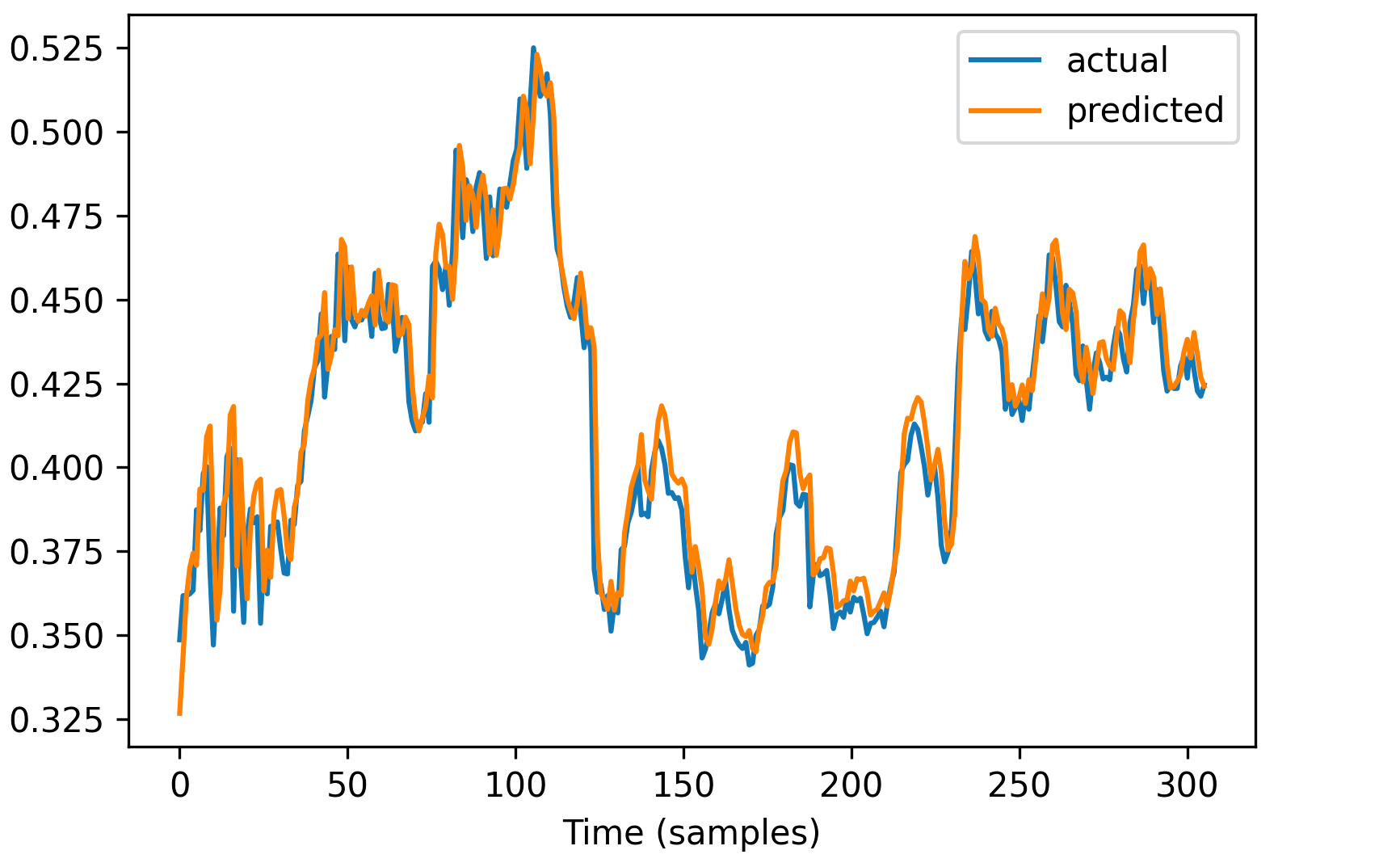}
 }
 \subfigure[Step 3]{
\includegraphics[scale =0.13]{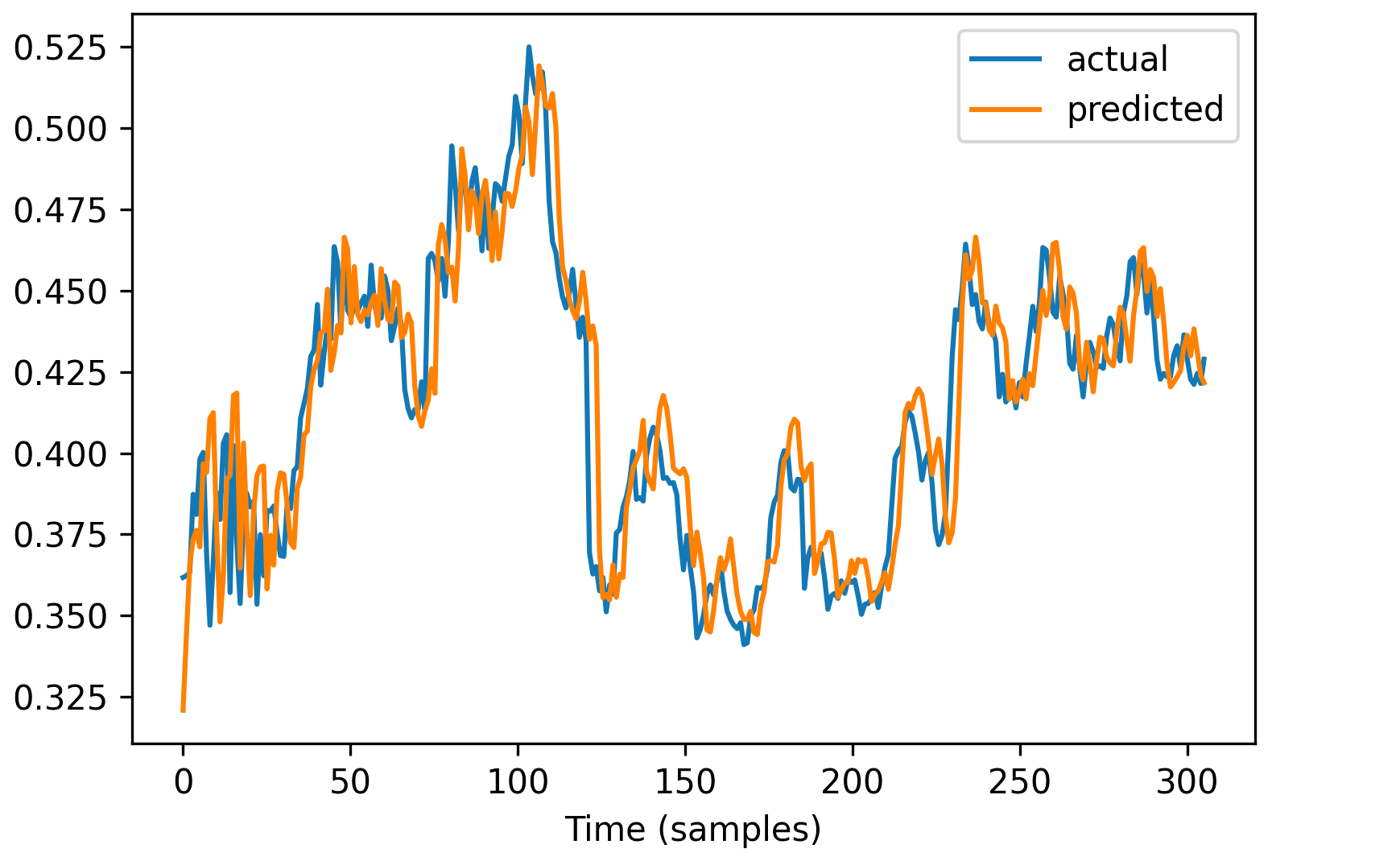}
 }
 \subfigure[Step 5]{
\includegraphics[scale =0.13]{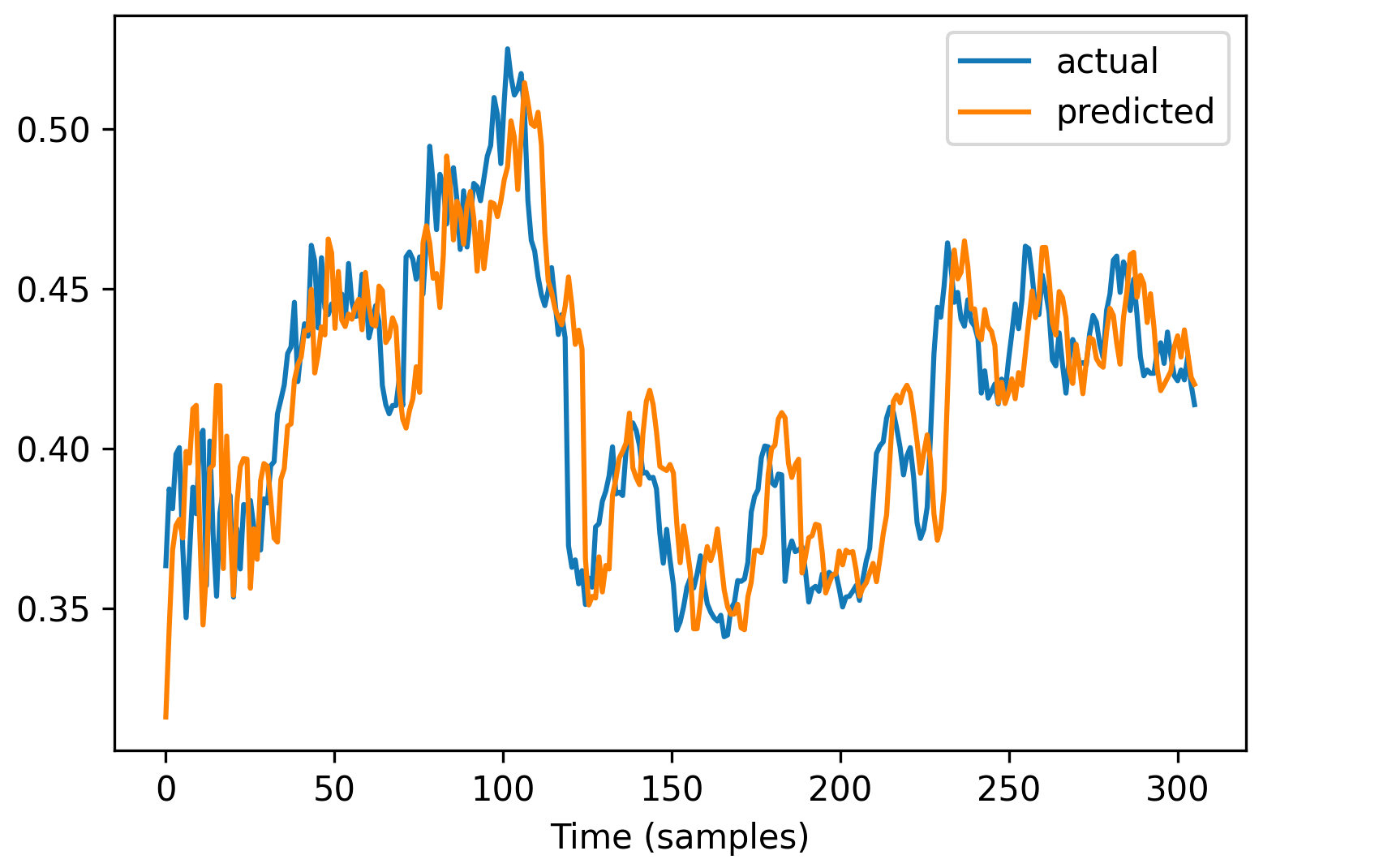}
 }
 \subfigure[Step 10]{
\includegraphics[scale =0.13]{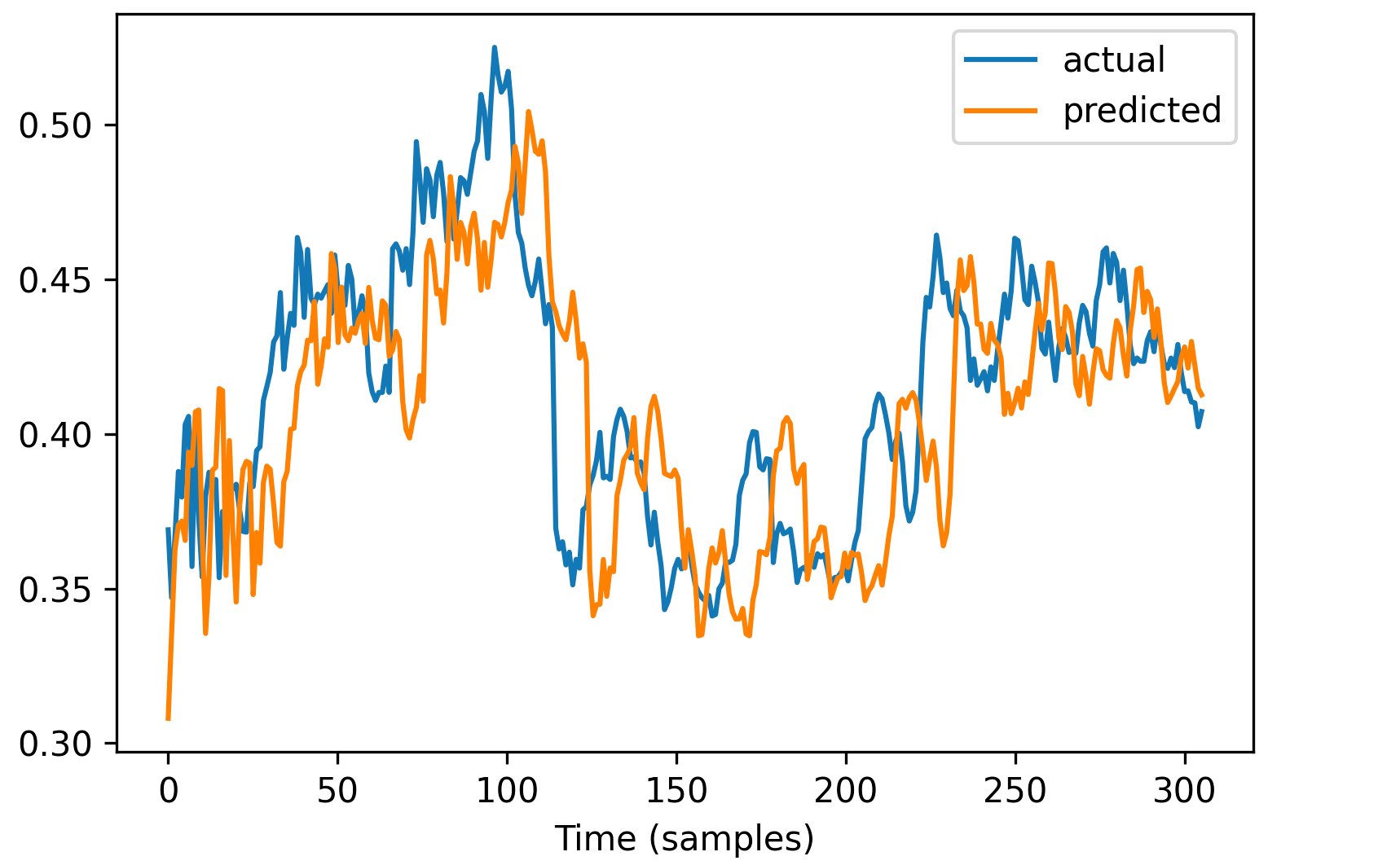}
 }
\caption{ACI-finance actual vs predicted values for Encoder-Decoder LSTM Model}
\label{fig:financeBest}
\end{figure*}

\begin{figure*}[htb]
\centering
\subfigure[Step 1]{
\includegraphics[scale =0.13]{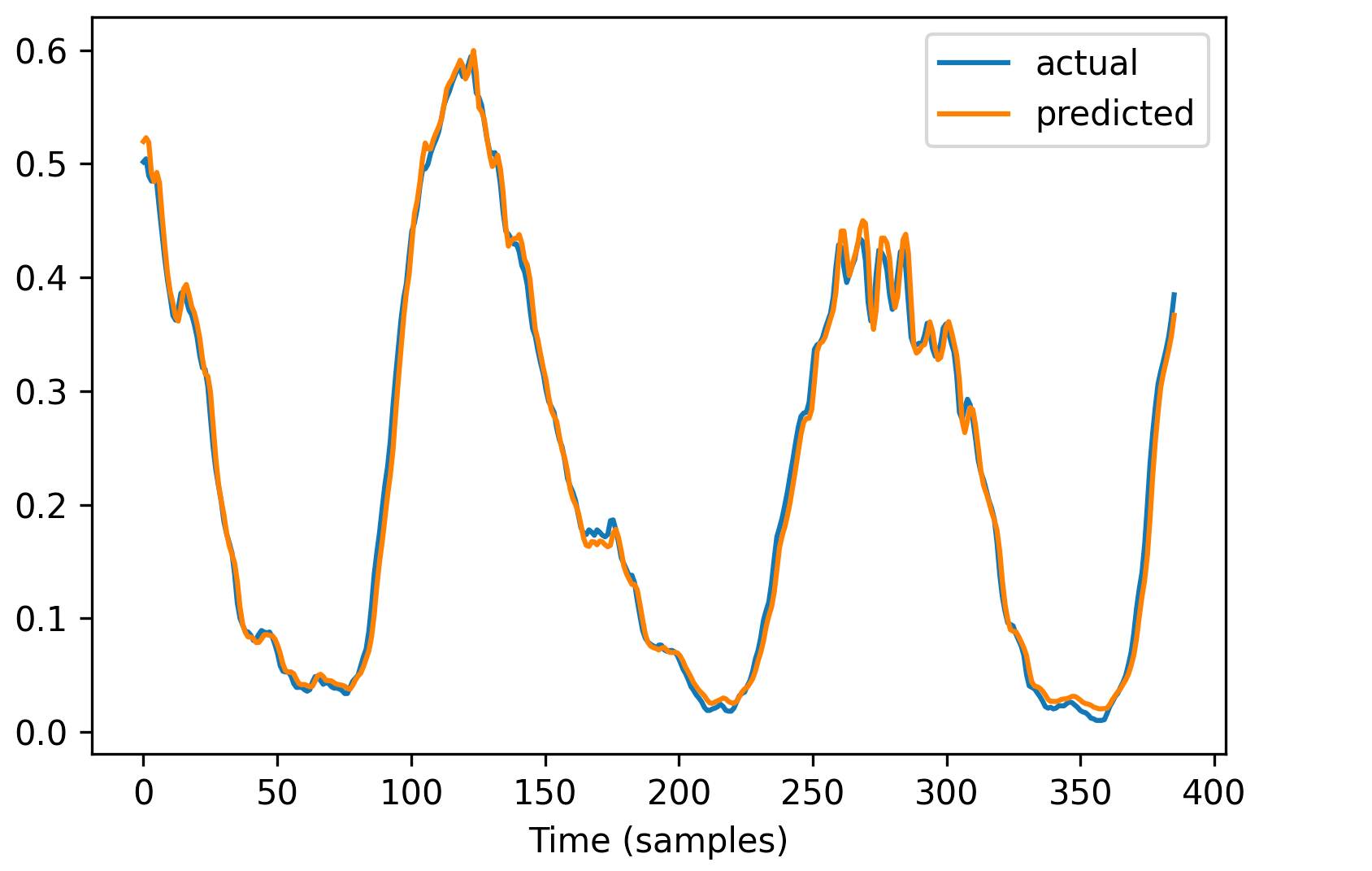}
 }
 \subfigure[Step 3]{
\includegraphics[scale =0.13]{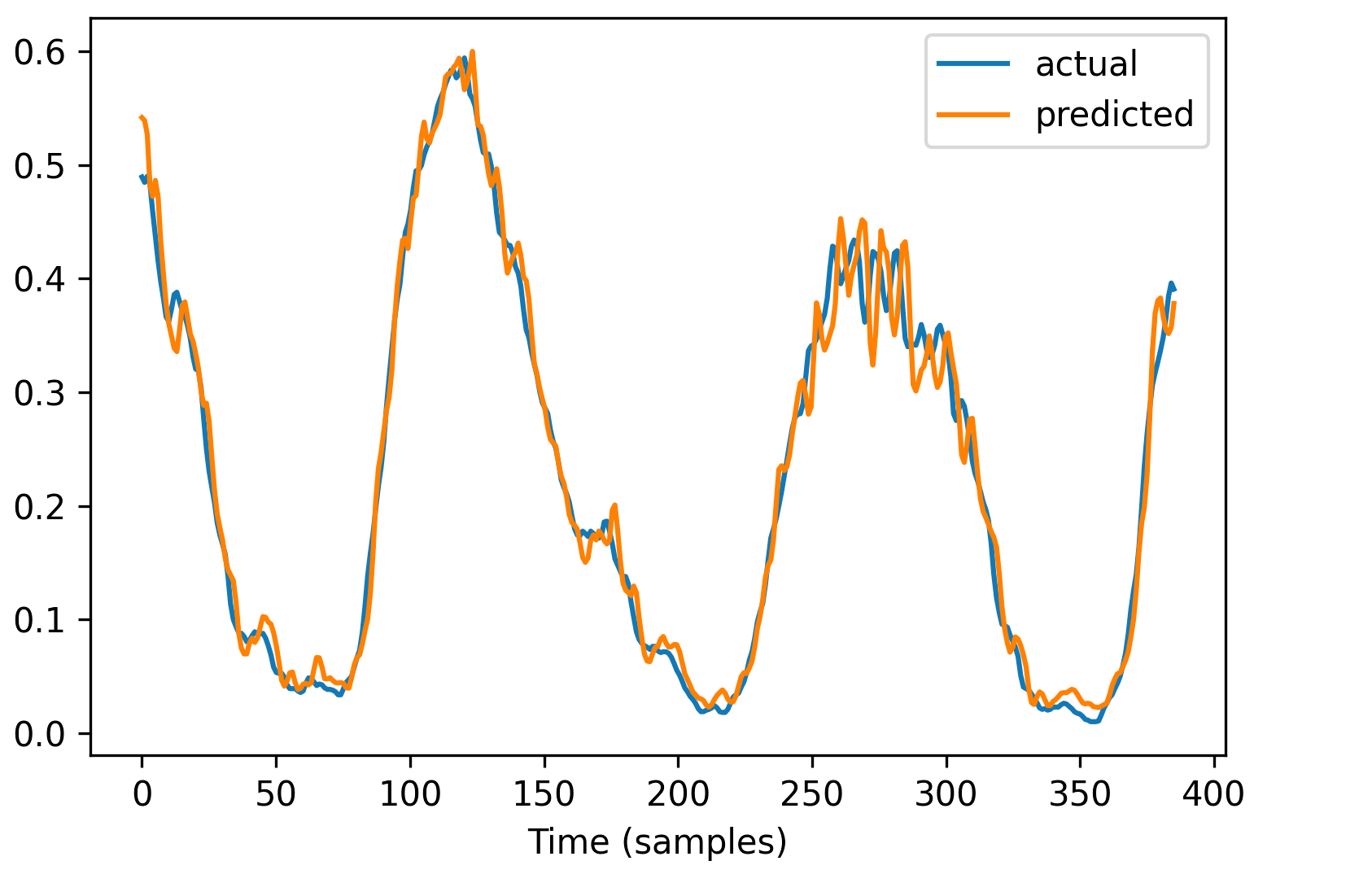}
 }
 \subfigure[Step 5]{
\includegraphics[scale =0.13]{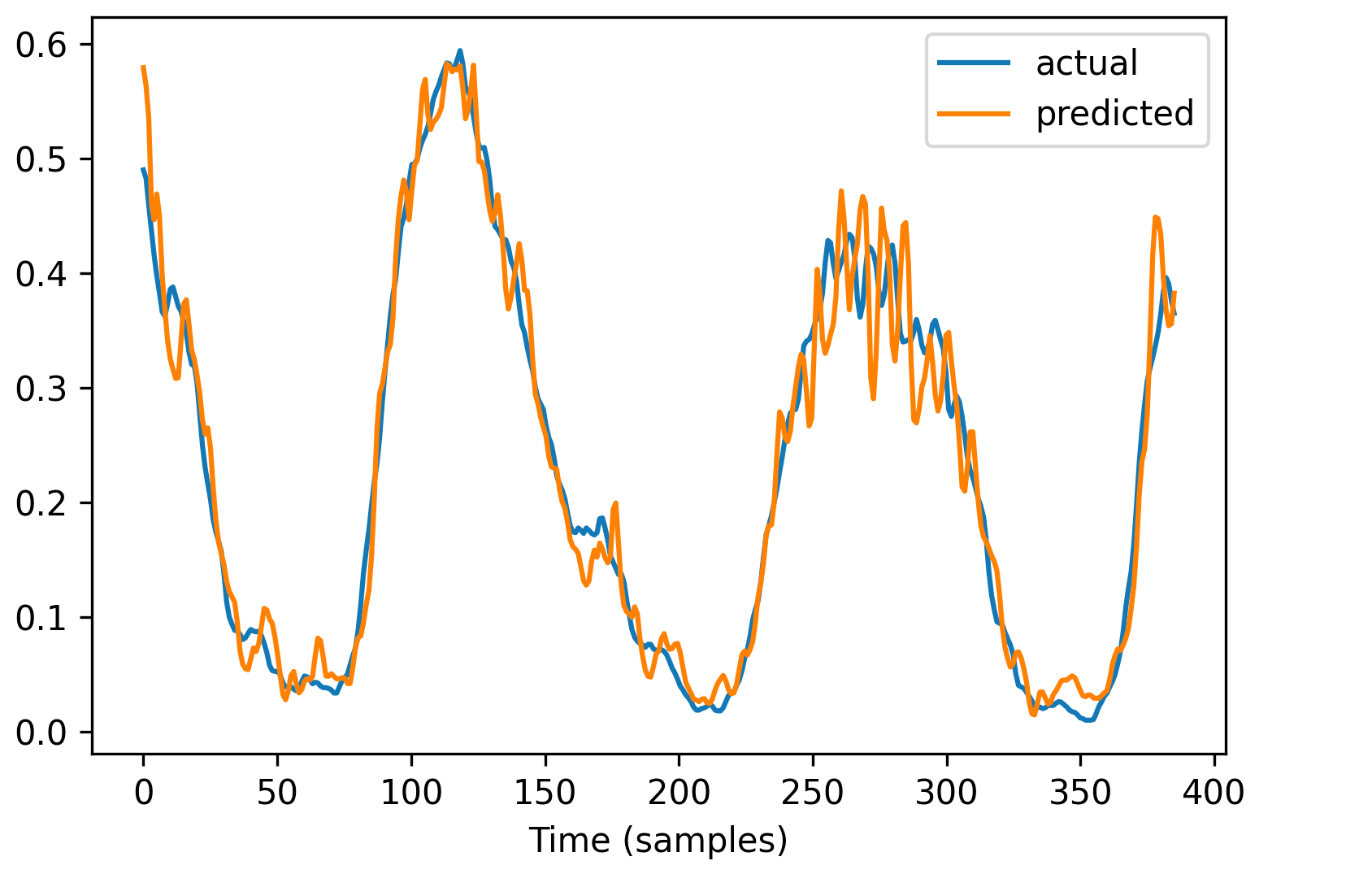}
 }
 \subfigure[Step 10]{
\includegraphics[scale =0.13]{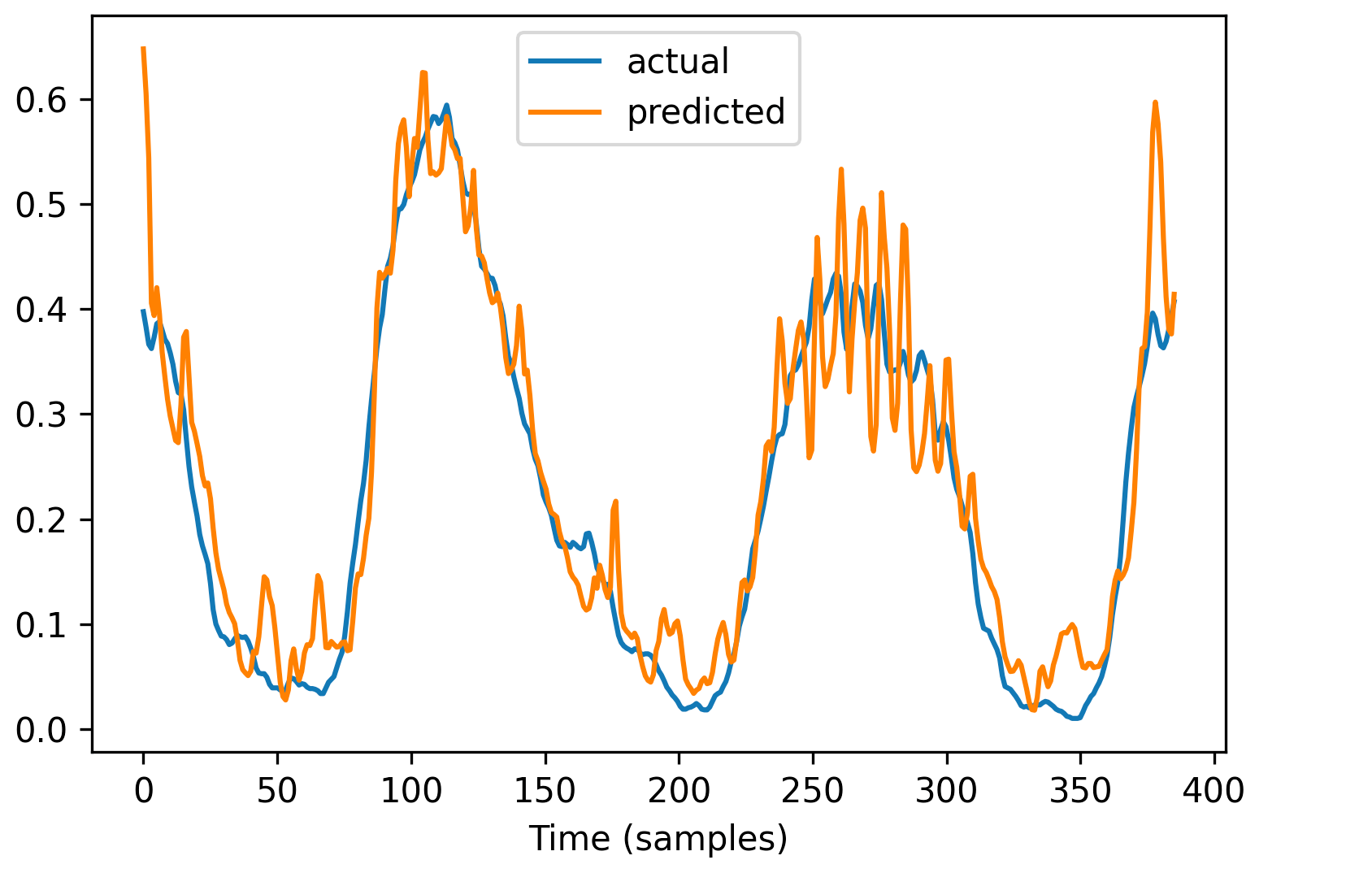}
 }
\caption{Sunspot actual vs predicted values for Encoder-Decoder LSTM Model}
\label{fig:sunspotsingle}
\end{figure*}

\begin{figure*}[htb]
\centering
\subfigure[Step 1]{
\includegraphics[scale =0.13]{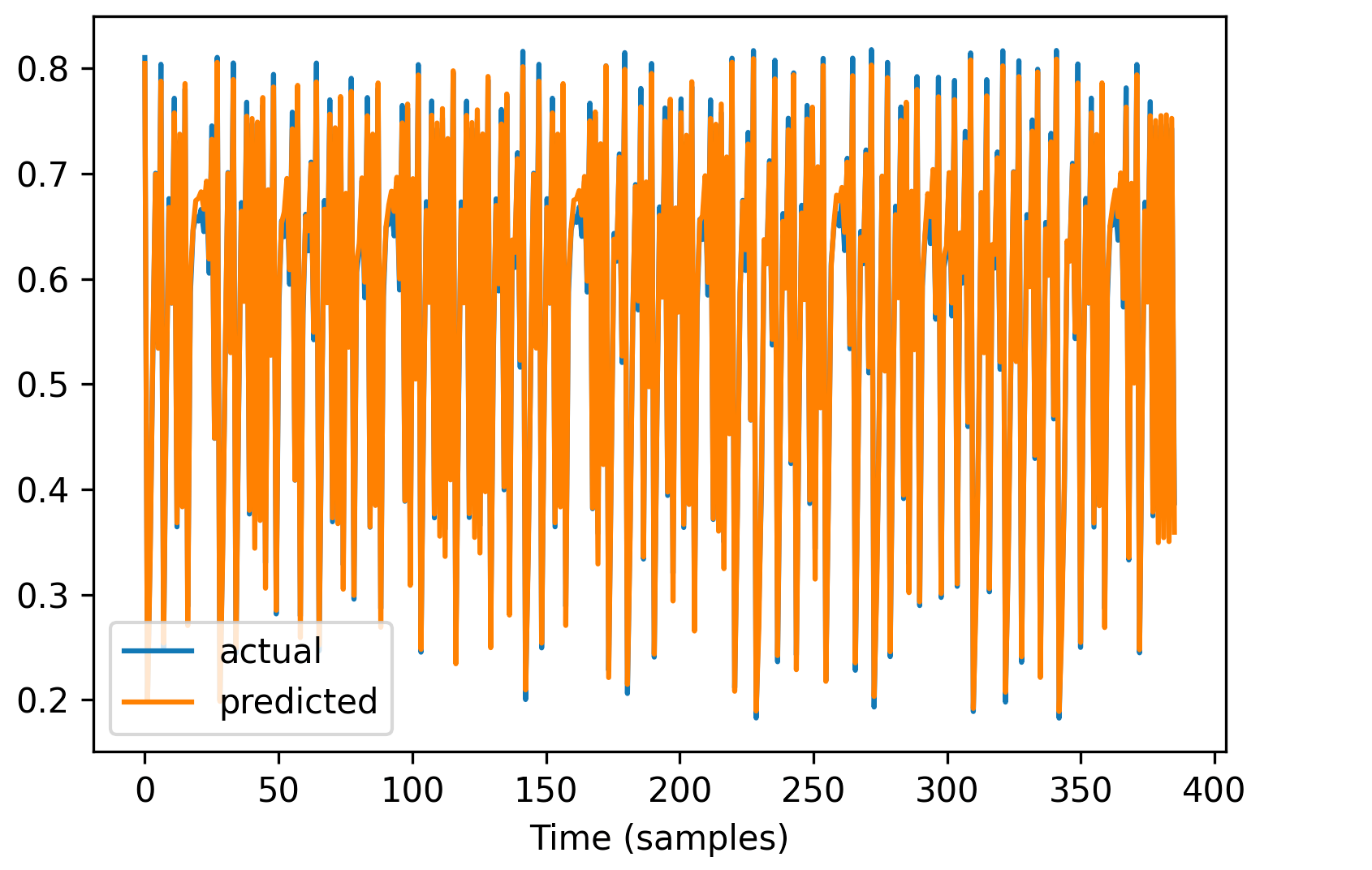}
 }
 \subfigure[Step 3]{
\includegraphics[scale =0.13]{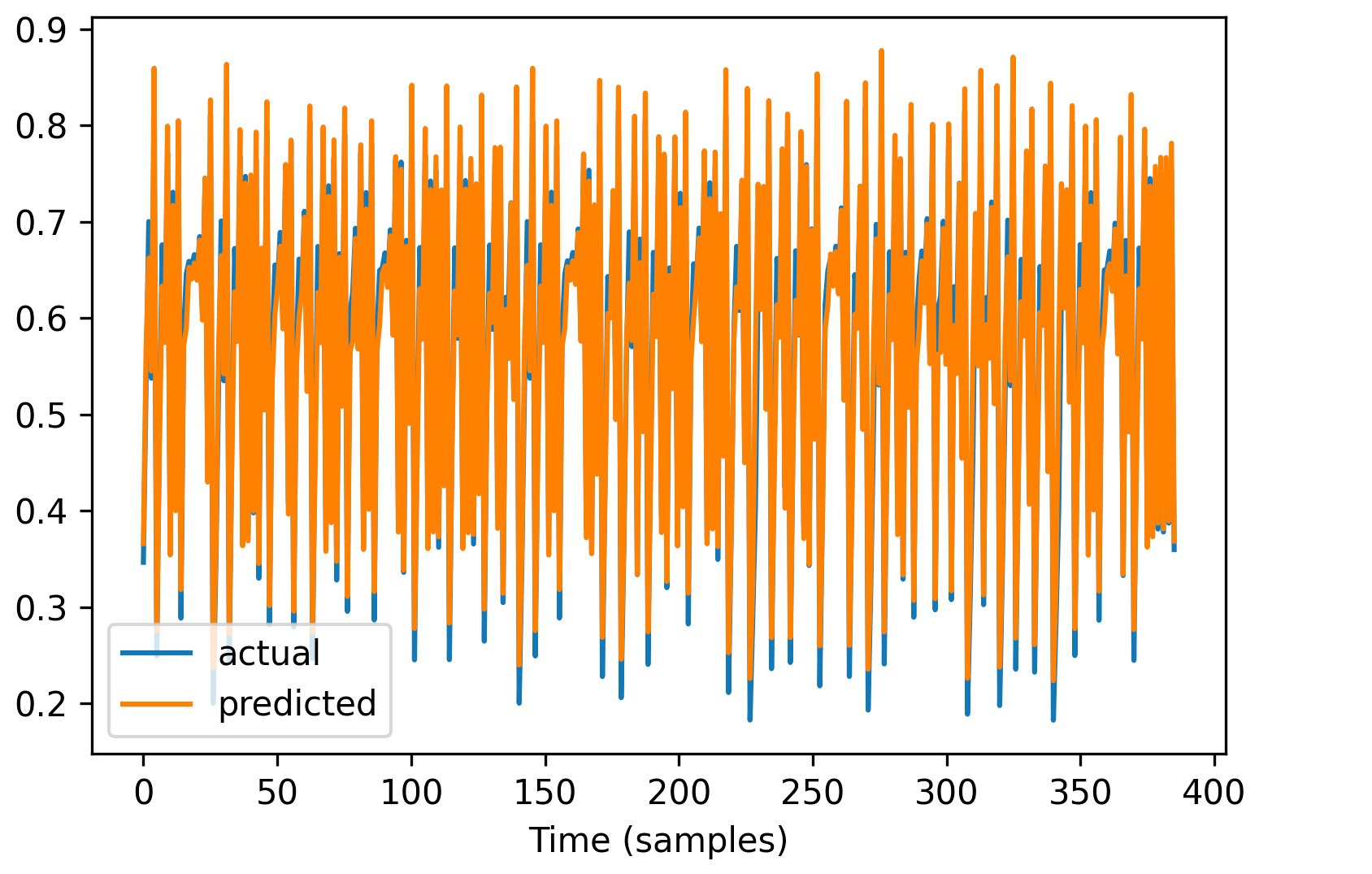}
 }
 \subfigure[Step 5]{
\includegraphics[scale =0.13]{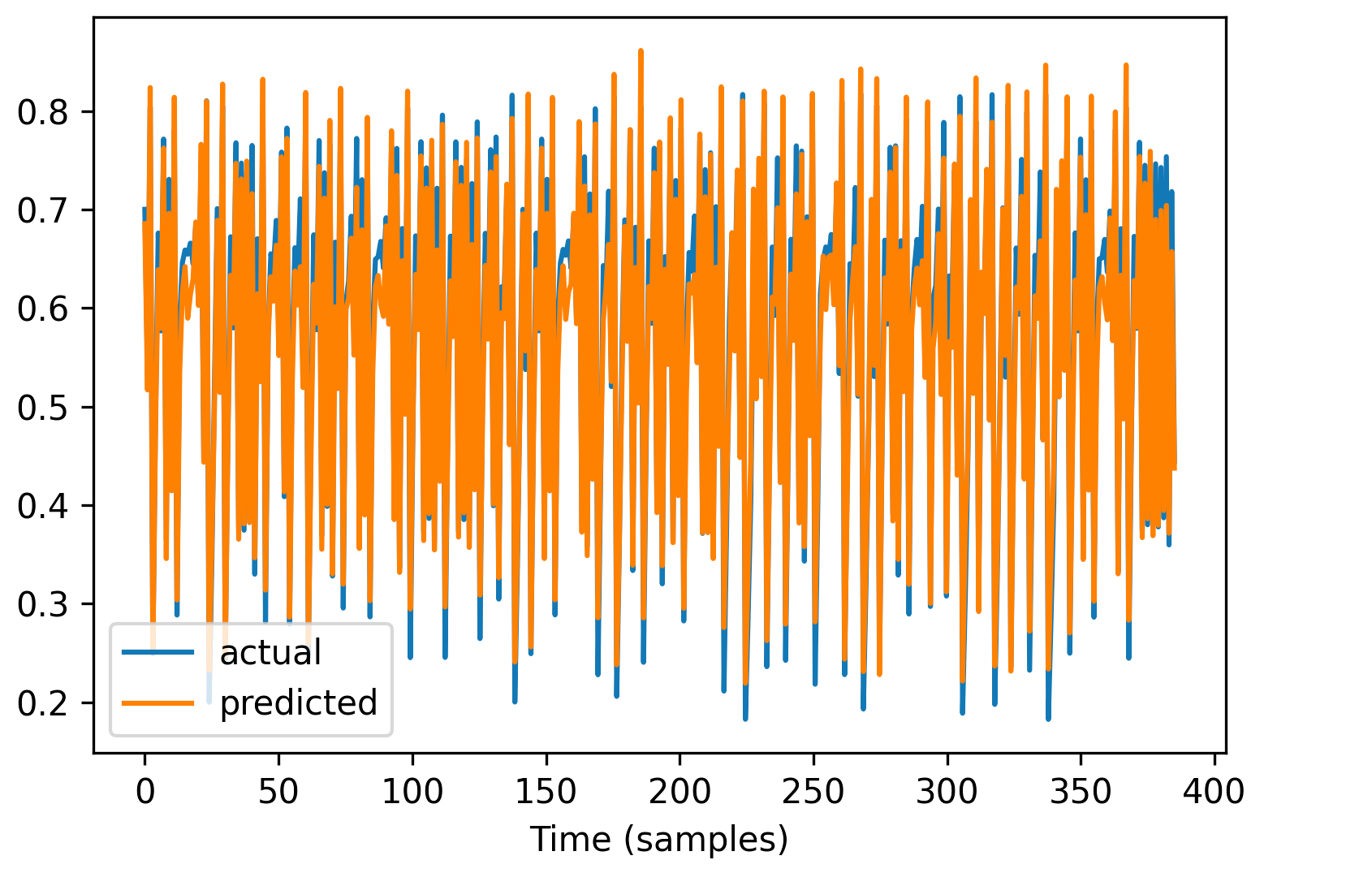}
 }
 \subfigure[Step 10]{
\includegraphics[scale =0.13]{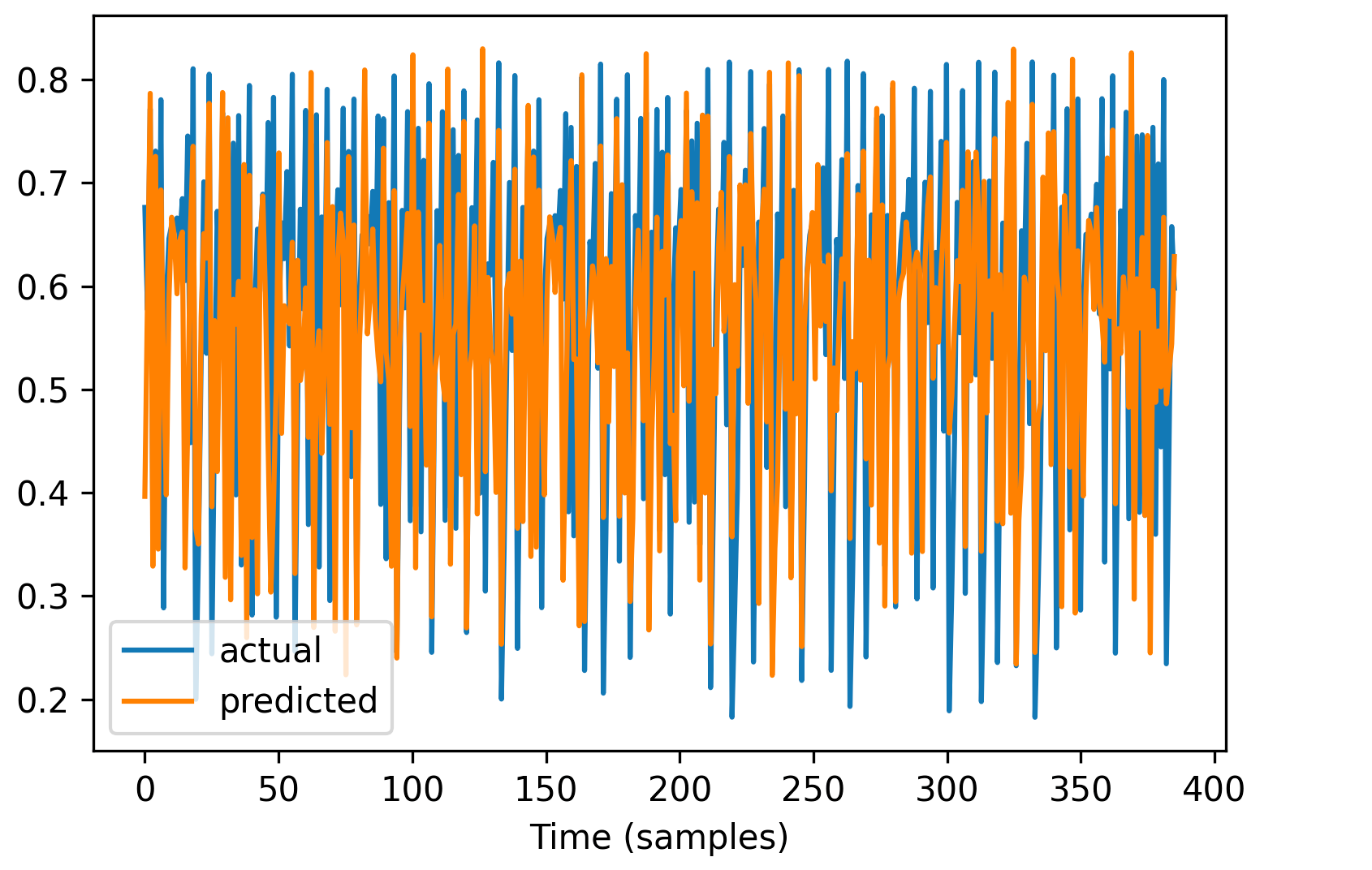}
 }
 
\caption{Henon actual vs predicted values for Encoder-Decoder LSTM Model}

\label{fig:henonsingle}
\end{figure*}

\begin{figure*}[htb]
\centering
\subfigure[Step 1]{
\includegraphics[scale =0.13]{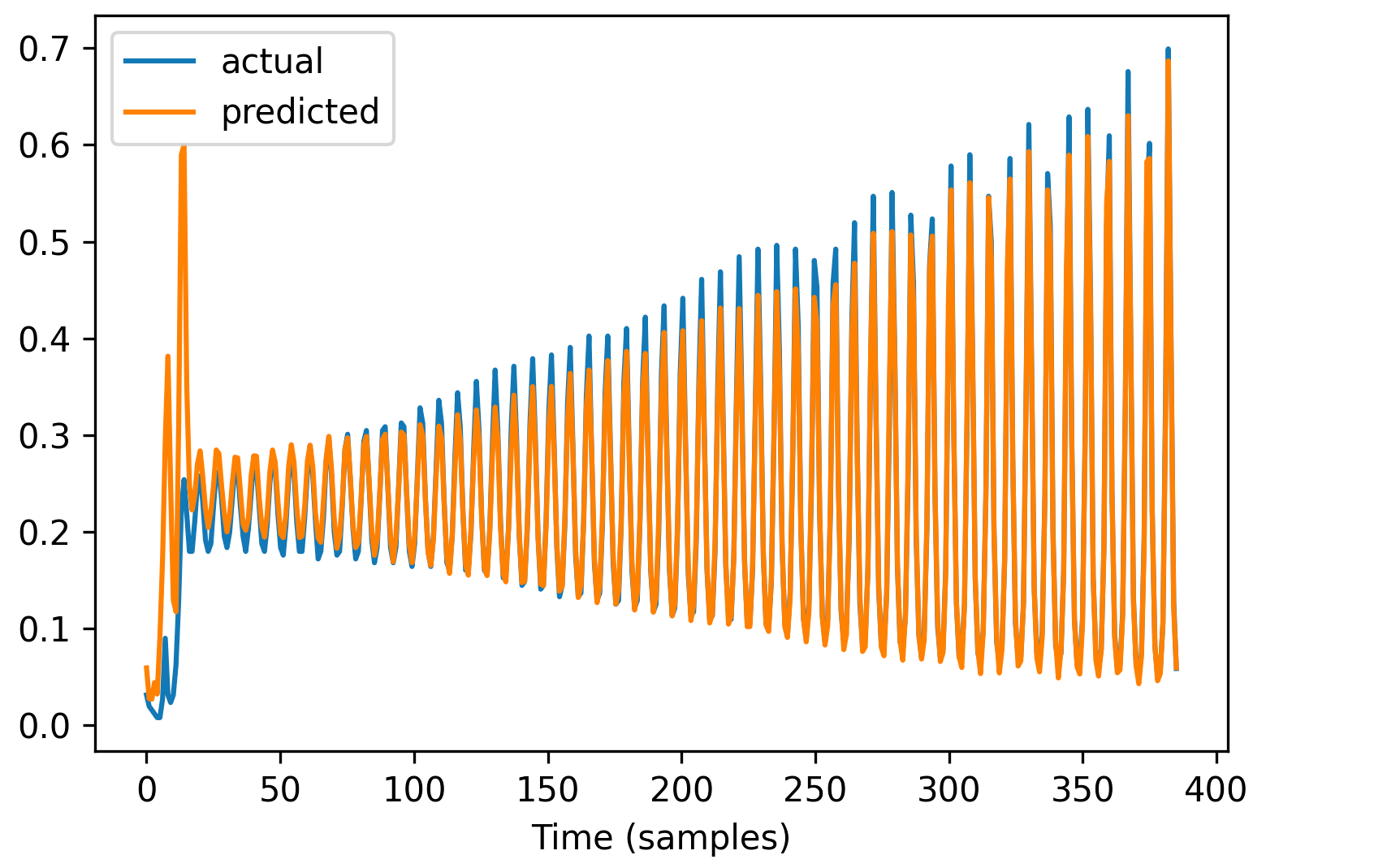}
 }
 \subfigure[Step 3]{
\includegraphics[scale =0.13]{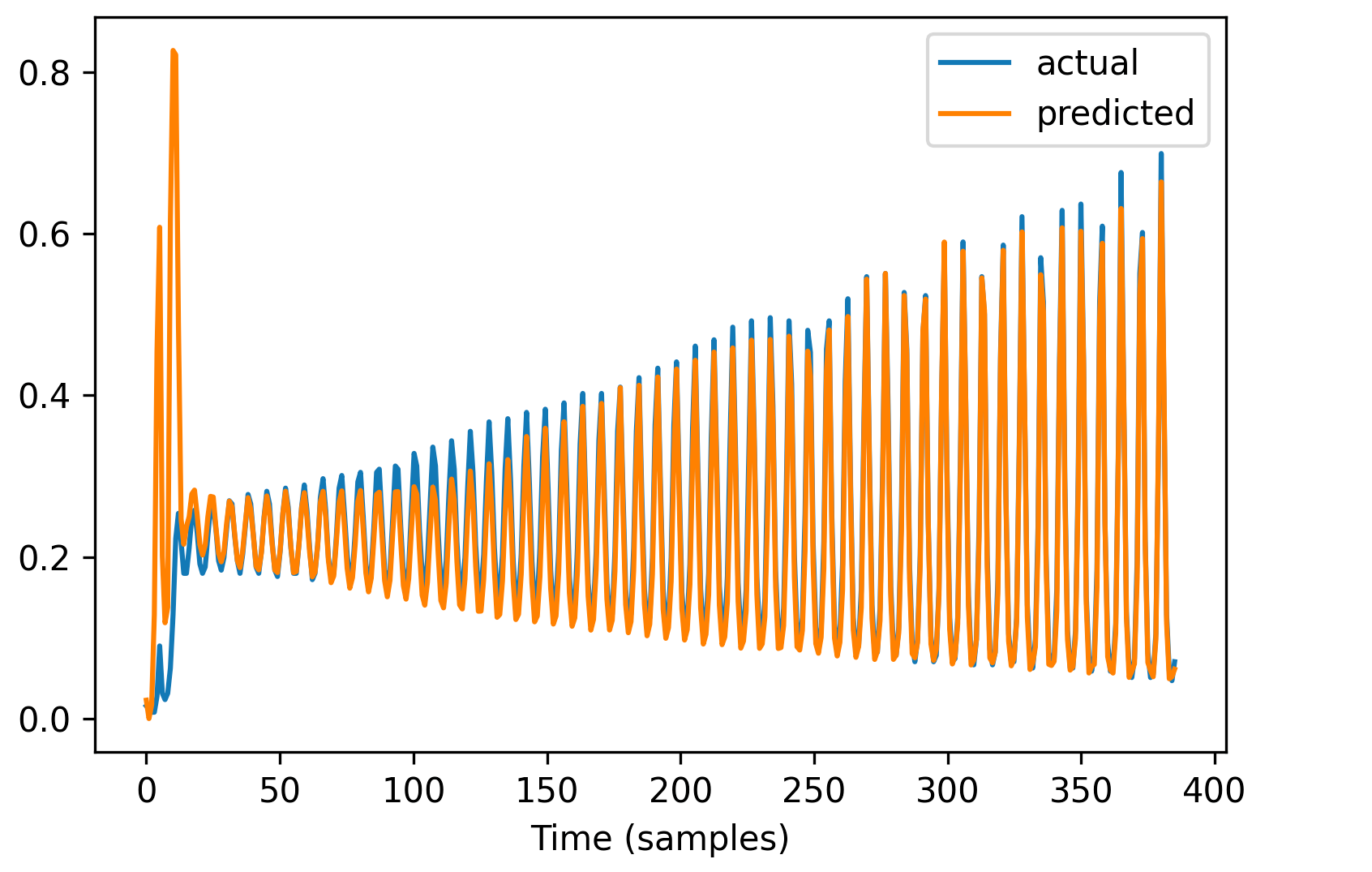}
 }
 \subfigure[Step 5]{
\includegraphics[scale =0.13]{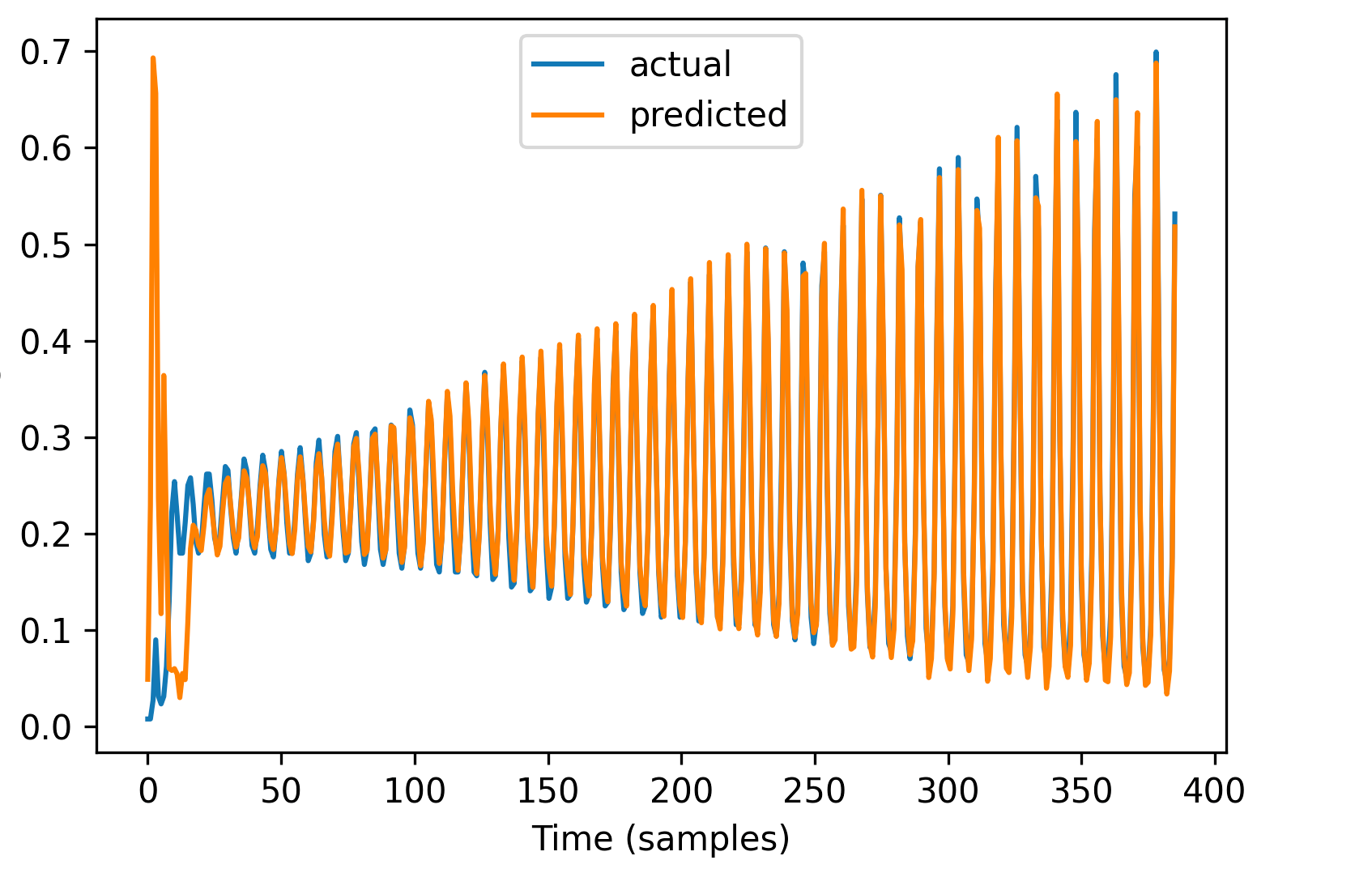}
 }
 \subfigure[Step 10]{
\includegraphics[scale =0.13]{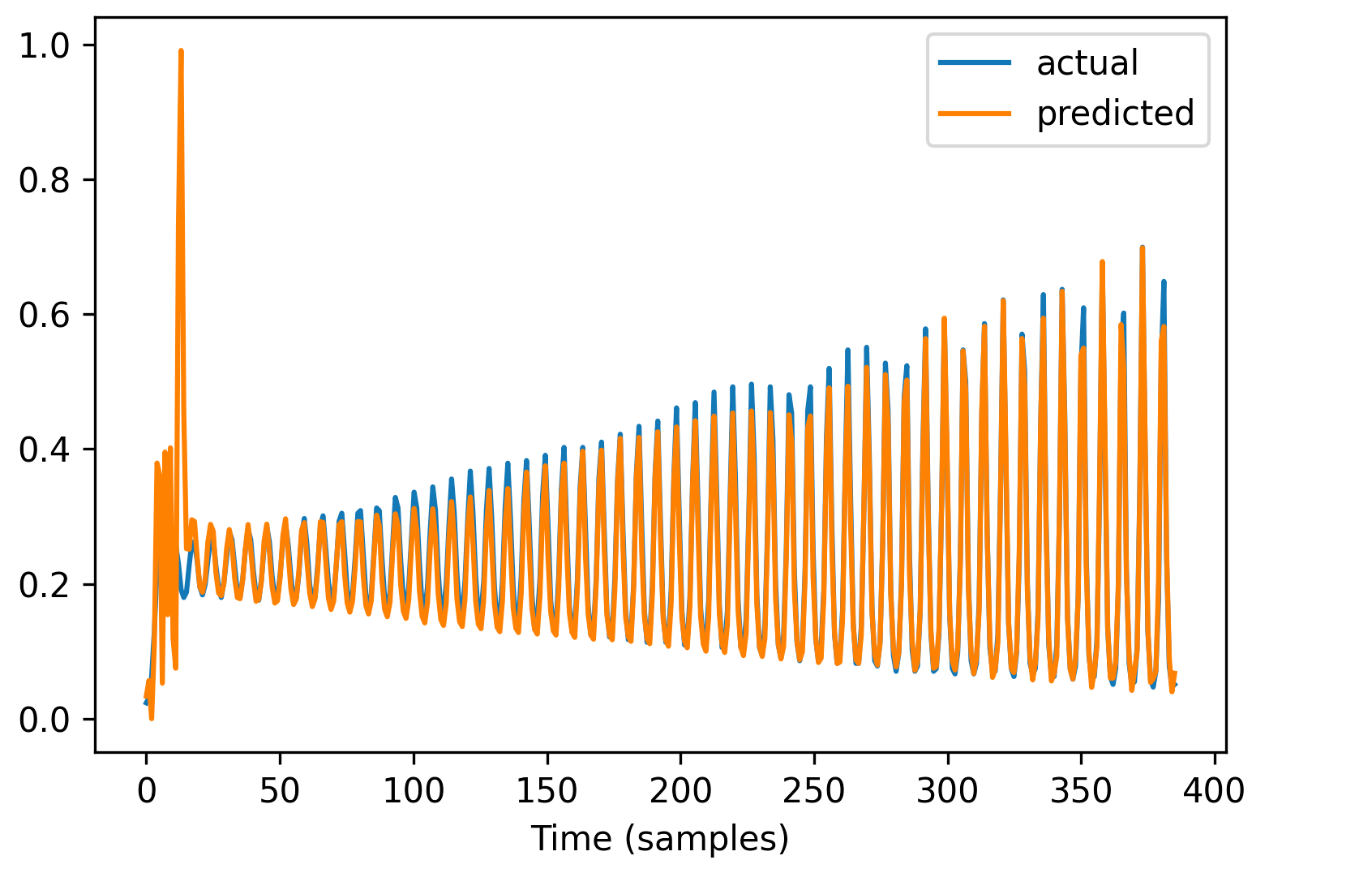}
 }
 
\caption{Lazer actual vs predicted values for Encoder-Decoder LSTM Model}

\label{fig:lazersingle}
\end{figure*}

\begin{figure*}[htb]
\centering
\subfigure[Step 1]{
\includegraphics[scale =0.13]{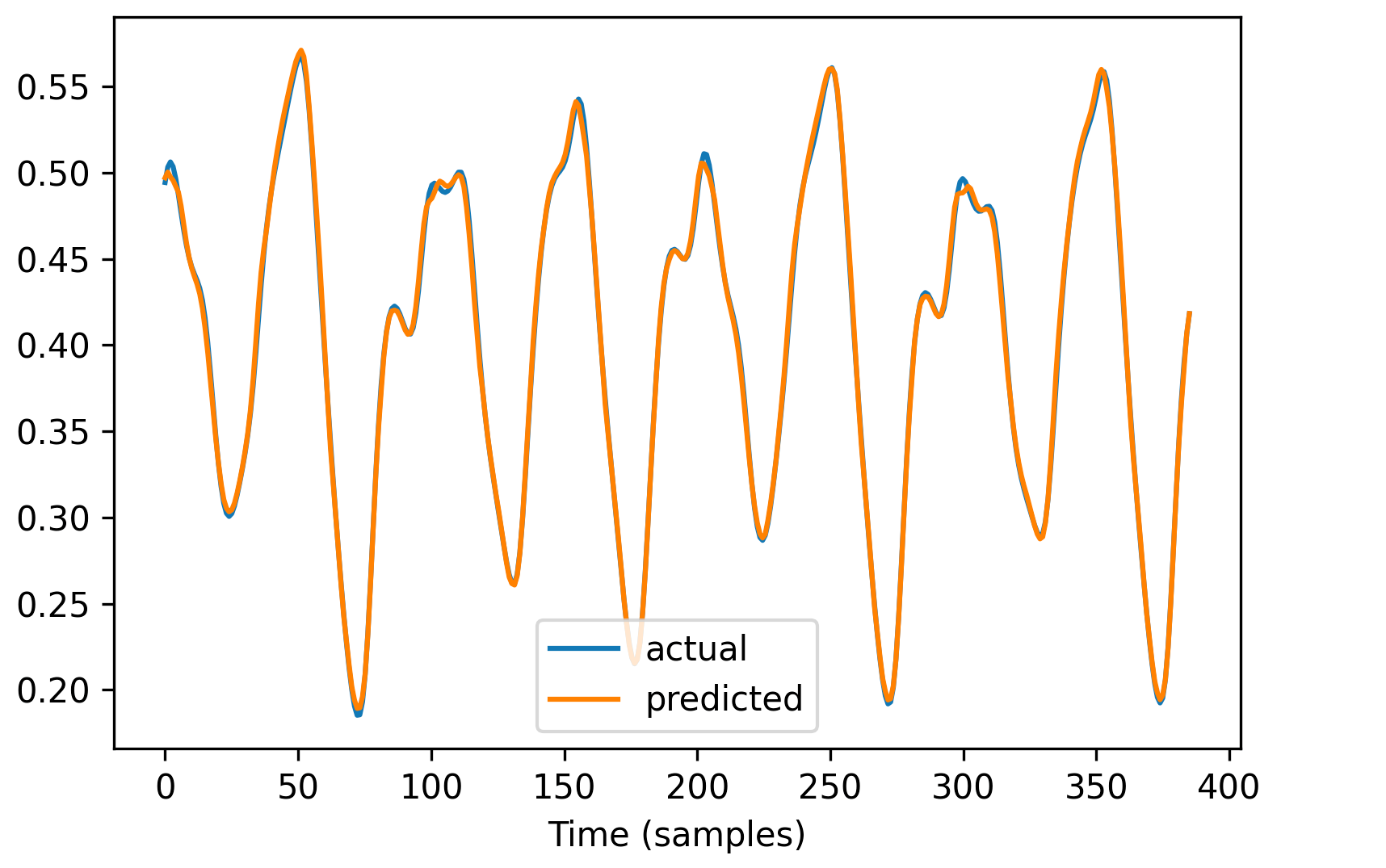}
 }
 \subfigure[Step 3]{
\includegraphics[scale =0.13]{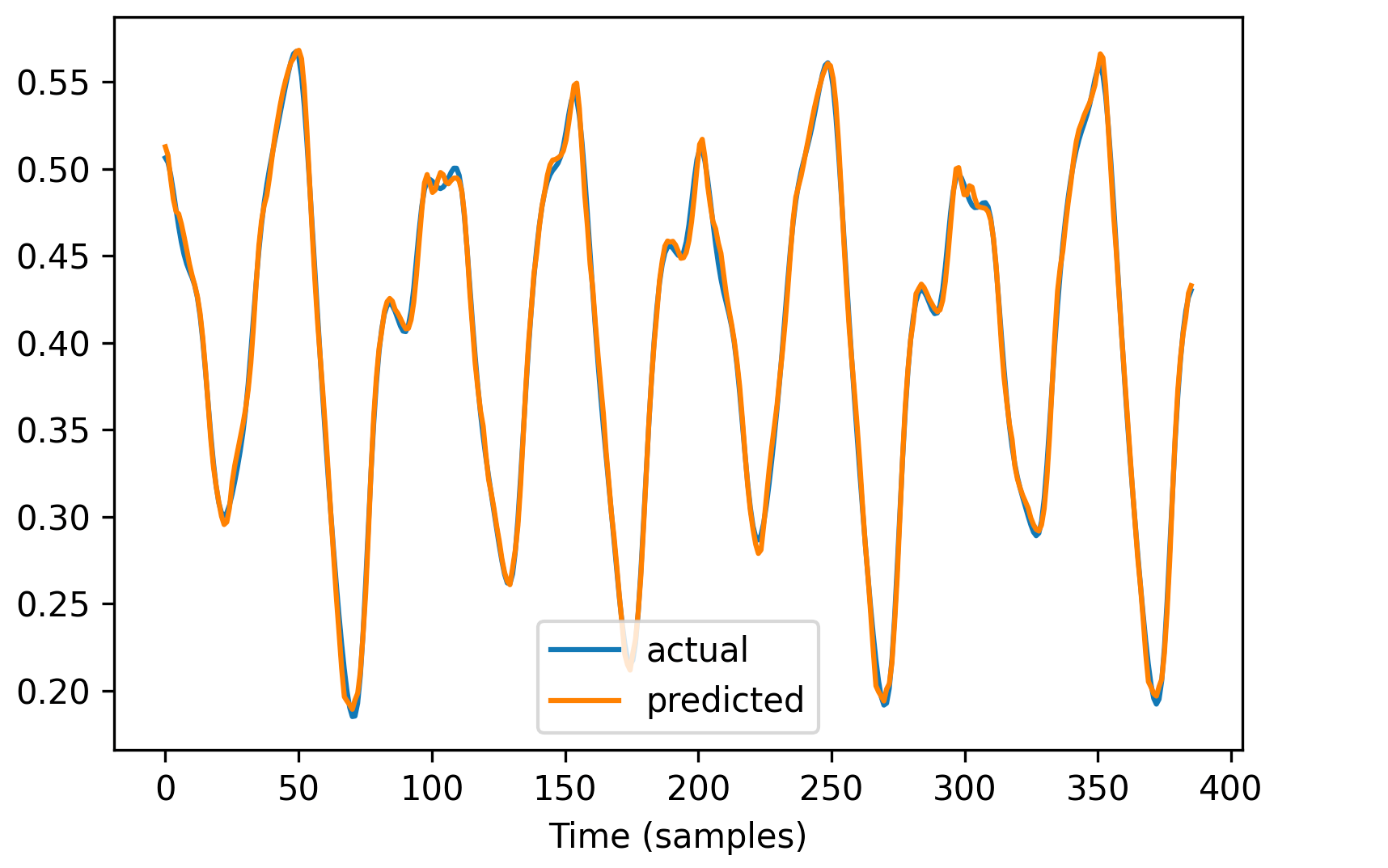}
 }
 \subfigure[Step 5]{
\includegraphics[scale =0.13]{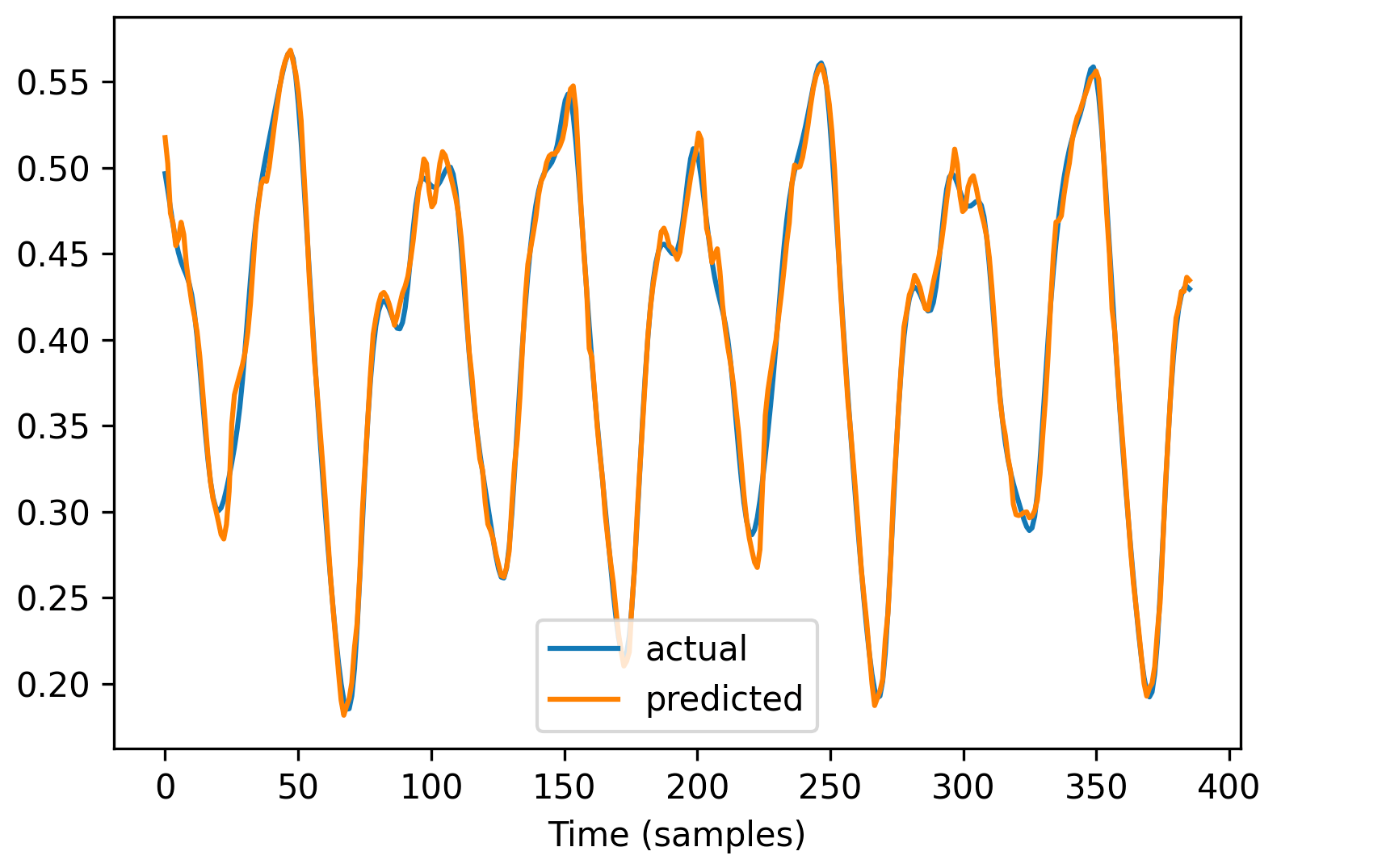}
 }
 \subfigure[Step 10]{
\includegraphics[scale =0.13]{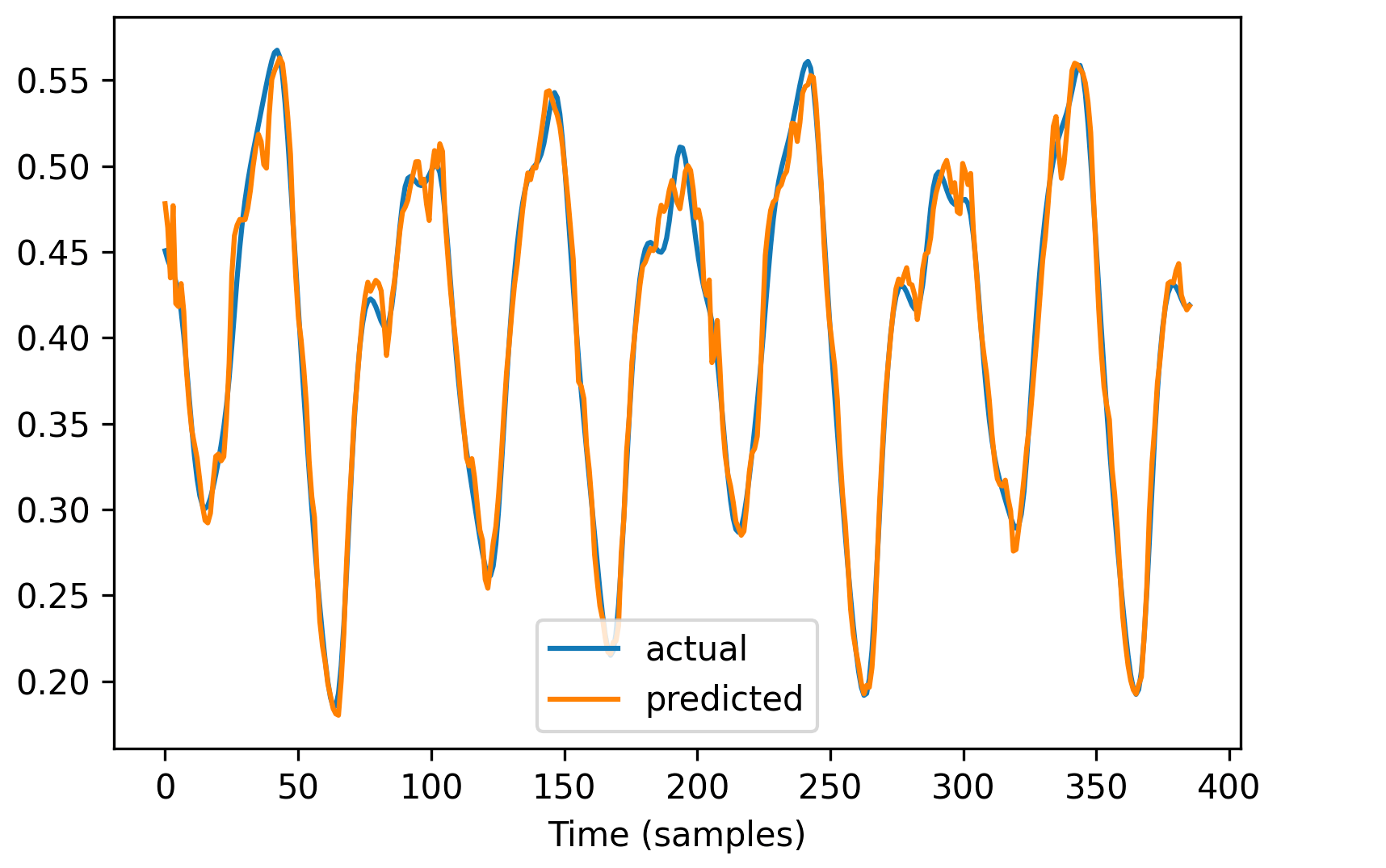}
 }
\caption{Mackey-Glass actual vs predicted values for Encoder-Decoder LSTM Model}
\label{fig:mackeysingle}
\end{figure*}

\begin{table*}[htbp!]
 \small 
 \centering
 \caption{Performance (rank) of different models for respective  time-series problems. Note lower rank denotes better performance. }
\label{tab:resultranks}
\begin{tabular}{llllllll}
\hline
 &  FNN-Adam& FNN-SGD & LSTM & BD-LSTM  & ED-LSTM & RNN & CNN\\
\hline
\hline
ACI-finance  &      
2&7&1&3&4&5&6\\
Sunspot &      
6&7&2&1&3&4&5\\
Lazer &      
6&7&1&2&3&5&4\\
Henon &      
6&7&3&2&1&5&4\\
Lorenz &      
6&7&2&3&1&5&4\\
Mackey-Glass &      
6&7&4&2&1&5&1\\
Rossler &      
6&7&4&1&2&5&3\\
\hline
Mean-Rank & 5.42 & 7.00  & 2.42  & 2.00  & 2.14  & 4.85  & 3.85 \\

\hline &
\end{tabular}

\end{table*}

 \subsection{Comparison with the literature}

\begin{table*}[htbp!]

\centering
 \small
 \caption{Comparison with Literature for Simulated time series. }
  
\label{tab:Simulated}
\begin{tabular}{llllll}

\hline
Problem & Method & 2-step & 5-step & 8-step & 10 steps\\
 \hline
 \hline
Mackey-Glass & &  	   &	& &\\

 &2SA-RTRL* \cite{chang2012reinforced}&  0.0035  	  &	 & & \\


 &ESN*\cite{chang2012reinforced}&    0.0052	&   	&   &\\
 &EKF\cite{Wu2013AMC}&     	&   0.2796	 &   &\\
 &G-EKF \cite{Wu2013AMC}&     	&   0.2202	 &   &\\
  &UKF \cite{Wu2013AMC}&     	&   	 0.1374&   &\\
  &G-UKF \cite{Wu2013AMC}&     	&   	0.0509&  & \\
 &GPF\cite{Wu2013AMC} &     	&   	0.0063&   &\\

 &G-GPF\cite{Wu2013AMC} &     	&   0.0022	&   &\\

&Multi-KELM\cite{YE2019227} &		0.0027&0.0031&0.0028&0.0029\\
&MultiTL-KELM\cite{YE2019227} &	\textbf{0.0025}&	\textbf{0.0029}&\textbf{0.0026}&\textbf{0.0028}\\
&CMTL \cite{chandra2017CMTLMulti}&  0.0550 & 	0.0750 & 0.0105 & 0.1200	\\



&ANFIS(SL) \cite{ZHOU2019343} &0.0051 &0.0213 & 0.0547 & \\
&R-ANFIS(SL) \cite{ZHOU2019343}&0.0045 &0.0195  &	0.0408 & \\
&R-ANFIS(GL) \cite{ZHOU2019343}&0.0042 & 	0.0127 & 0.0324 & \\

&FNN-Adam &   0.0256$\pm$	0.0038   	&   0.0520$\pm$	0.0044     &0.0727$\pm$	0.0050	       &0.0777$\pm$	0.0043\\

&FNN-SGD  &     0.0621$\pm$	0.0051 	&  0.0785$\pm$	0.0025       &0.0937$\pm$	0.0022	      &0.0990$\pm$	0.0026\\

&LSTM   &   0.0080$\pm$	0.0014   	&    0.0238$\pm$	0.0024    &0.0381$\pm$	0.0029	       &0.0418$\pm$	0.0033\\

&BD-LSTM   &   0.0083$\pm$	0.0015   	&   0.0202$\pm$	0.0026     &0.0318$\pm$	0.0027	       &0.0359$\pm$	0.0026\\

&ED-LSTM   &   0.0076$\pm$	0.0014   	&    0.0168$\pm$	0.0027    &	0.0248$\pm$	0.0036       &0.0271$\pm$	0.0040\\

&RNN  &    0.0142$\pm$	0.0001  	&   0.0365$\pm$	0.0001     &0.0547$\pm$	0.0001	       &0.0615$\pm$	0.0001\\

&CNN &     0.0120$\pm$		0.0010 	&   0.0262$\pm$		0.0016     &0.0354$\pm$		0.0018	       &0.0364$\pm$		0.0017\\

\hline

Lorenz &    	   &	& \\

 &2SA-RTRL*\cite{chang2012reinforced}  &  0.0382	   &	 &  \\


 &ESN*\cite{chang2012reinforced}&    0.0476 	   &	&   \\
 
 &CMTL \cite{chandra2017CMTLMulti} & 0.0490	 &  0.0550	&0.0710 & 0.0820  \\
&FNN-Adam &    0.0206$\pm$	0.0046  	&  0.0481$\pm$	0.0072      & 0.0678$\pm$	0.0058	       &0.0859$\pm$	0.0065\\

&FNN-SGD  & 0.0432$\pm$	0.0030     	&   0.0787$\pm$	0.0030     &0.1027$\pm$	0.0025	       &0.1178$\pm$	0.0026\\

&LSTM   & \textbf{0.0033$\pm$	0.0010}     	&  0.0064$\pm$	0.0026      &	0.0101$\pm$	0.0038       &0.0129$\pm$	0.0042\\

&BD-LSTM   & 0.0054$\pm$	0.0026     	&  0.0079$\pm$	0.0036      &0.0125$\pm$	0.0057	       &0.0146$\pm$	0.0059\\

&ED-LSTM   &0.0044$\pm$	0.0012      	&  \textbf{0.0059$\pm$	0.0009}      &\textbf{0.0090$\pm$	0.0009}	       &\textbf{0.0110$\pm$	0.0012}\\

&RNN  &   0.0129$\pm$	0.0012   	& 0.0155$\pm$	0.0024       &0.0186$\pm$	0.0042	       &0.0226$\pm$	0.0058\\

&CNN &  	0.0067	$\pm$0.0007    	&  0.0098$\pm$	0.0009      &0.0132$\pm$	0.0011	       &0.0157$\pm$	0.0015\\
\hline

Rossler &    	  &	& \\

&CMTL \cite{chandra2017CMTLMulti} & 0.0421   & 0.0510  &0.0651	& 0.0742 	   \\

&FNN-Adam & 0.0202$\pm$	0.0024     	&  0.0400$\pm$	0.0039      &0.0603$\pm$	0.0050	       &0.0673$\pm$	0.0056\\

&FNN-SGD  &  0.0666$\pm$	0.0058    	& 0.1257$\pm$	0.0082       &0.1664$\pm$	0.0075	       &0.1881$\pm$	0.0078\\

&LSTM   &  0.0086$\pm$	0.0011    	&  0.0135$\pm$	0.0015      &0.0185$\pm$	0.0022	       &0.0225$\pm$	0.0026\\

&BD-LSTM   &  \textbf{0.0047$\pm$	0.0014}    	& \textbf{ 0.0084$\pm$	0.0021 }     &	\textbf{0.0142$\pm$	0.0027 }      &\textbf{0.0178$\pm$	0.0032}\\

&ED-LSTM   &  0.0082$\pm$	0.0019    	&   0.0128$\pm$	0.0021     &0.0159$\pm$	0.0024	       &0.0180$\pm$	0.0030\\

&RNN  &    0.0218$\pm$	0.0005  	&   0.0314$\pm$	0.0004     &0.0382$\pm$	0.0004	       &0.0424$\pm$	0.0004\\

&CNN & 0.0105$\pm$		0.0011     	&  0.0122$\pm$		0.0016      &	0.0157$\pm$		0.0020	       &	0.0220$\pm$		0.0022\\
\hline

Henon &    	  & 	& \\
&Multi-KELM\cite{YE2019227}&0.0041&	0.2320&	0.2971&	0.2968\\
&MultiTL-KELM\cite{YE2019227}&	 \textbf{0.0031}&	0.1763&0.2452&	0.2516\\
&CMTL \cite{chandra2017CMTLMulti} & 0.2103   & 0.2354  &0.2404	   & 0.2415 \\

&FNN-Adam &  0.1606 $\pm$  0.0024     	&0.1731 $\pm$  0.0005         &	0.1781 $\pm$  0.0005        & 0.1762 $\pm$  0.0009\\

&FNN-SGD  & 0.1711 $\pm$   0.0018     	&  0.1769 $\pm$   0.0007      &0.1805 $\pm$   0.0012	       &0.1773 $\pm$   0.0011\\

&LSTM   & 0.0682 $\pm$   0.0058     	& 0.1584 $\pm$   0.0010       &	 0.1707 $\pm$   0.0008       & 0.1756 $\pm$   0.0005 \\

&BD-LSTM   & 0.0448 $\pm$  0.0026     	&  0.1287 $\pm$  0.0046        &0.1697 $\pm$  0.0008	       &0.1733 $\pm$  0.0003\\

&ED-LSTM   & 0.0454 $\pm$  0.0069     	& \textbf{0.0694 $\pm$  0.0161}        &	\textbf{0.1371 $\pm$  0.0107}        & \textbf{0.1689 $\pm$  0.0046}\\

&RNN  &    0.1515$\pm$ 0.0016    	& 0.1718 $\pm$  0.0001       &	 0.1768 $\pm$  0.0001       &0.1751 $\pm$  0.0002\\

&CNN & 0.0859	$\pm$ 0.0038     	& 	0.1601$\pm$ 	0.0007       &	0.1718$\pm$ 	0.0003       &0.1737$\pm$ 	0.0002\\
\hline

\end{tabular}
\end{table*}

\begin{table*}[htbp!]

\centering
 \small
 \caption{Comparison with Literature for Real World time series. }
  

\label{tab:Real}
\begin{tabular}{llllll}

 \hline
Problem& Method& 2-step & 5-step &8-step& 10-step\\
 \hline
 \hline

Lazer &    	&   	& \\
 
 
 

&CMTL \cite{chandra2017CMTLMulti} &  0.0762  &  0.1333  	& 0.1652& 0.1885  \\

&FNN-Adam &  0.1043 $\pm$  0.0018    	& 0.0761 $\pm$  0.0019       &	0.0642	$\pm$ 0.0020       &0.0924$\pm$	0.0018\\

&FNN-SGD  & 0.0983 $\pm$   0.0046     	&   0.0874 $\pm$  0.0072     &	0.0864$\pm$ 	0.0053       &0.0968$\pm$	0.0052\\

&LSTM   &  \textbf{0.0725 $\pm$ 0.0027 }   	&  \textbf{0.0512 $\pm$   0.0015}      &	0.0464$\pm$ 	0.0015       &\textbf{0.0561$\pm$	0.0044}\\

&BD-LSTM   &  0.0892 $\pm$  0.0022     	&   0.0596 $\pm$  0.0036      &\textbf{0.0460$\pm$ 	0.0015}	       &0.0631$\pm$	0.0037\\

&ED-LSTM   &  0.0894 $\pm$  0.0013     	& 0.0694 $\pm$  0.0073       &	0.0510$\pm$ 	0.0027       &0.0615$\pm$	0.0030\\

&RNN  &   0.1176 $\pm$  0.0019    	&  0.0755 $\pm$ 0.0011      &0.0611$\pm$ 	0.0015	       &0.0947$\pm$	0.0027\\

&CNN &  0.0729$\pm$	0.0014    	&  0.0701$\pm$	0.0020      & 0.0593$\pm$	0.0029	       & 0.0577$\pm$	0.0018\\

 \hline

Sunspot &    	  &	& \\

 &M-SVR \cite{zhang2013iterated} &   	&   &	 & 0.2355  $\pm $ 0.0583 \\
 
 &  SVR-I \cite{zhang2013iterated}&   	&   &	 &  0.2729 $\pm $0.1414  \\
 
 & SVR-D \cite{zhang2013iterated} &   	&   &	 & 0.2151 $\pm $ 0.0538 \\

&CMTL \cite{chandra2017CMTLMulti} &   0.0473 &  0.0623  &0.0771	 & 0.0974  \\ 

&FNN-Adam & 0.0236$\pm$	0.0015     	&0.0407$\pm$	0.0012        &	0.0582$\pm$	0.0019       &0.0745$\pm$	0.0020\\

&FNN-SGD  &  0.0352$\pm$	0.0022    	& 0.0610$\pm$	0.0024       &	0.0856$\pm$	0.0023       &0.1012$\pm$	0.0019\\

&LSTM   &   \textbf{0.0148$\pm$	0.0007 }  	& 0.0321$\pm$	0.0006       &	0.0449$\pm$	0.0007       &0.0587$\pm$	0.0010\\

&BD-LSTM   &  0.0155$\pm$	0.0007    	&   \textbf{0.0318$\pm$	0.0007}     &\textbf{0.0440$\pm$	0.0005}	       &\textbf{0.0576$\pm$	0.0010}\\

&ED-LSTM   &   0.0170$\pm$	0.0004   	& 0.0348$\pm$	0.0004       &0.0519$\pm$	0.0016	       &0.0673$\pm$	0.0022\\

&RNN  &  0.0212$\pm$	0.0003    	& 0.0395$\pm$	0.0002       &0.0503$\pm$	0.0002	       &0.0641$\pm$	0.0003\\

&CNN &   0.0257$\pm$	0.0002   	& 	0.0419	$\pm$0.0004       &	0.0555$\pm$	0.0006       &0.0723$\pm$	0.0008\\
 \hline

ACI-Finance &    	  &	 &\\

&CMTL \cite{chandra2017CMTLMulti} &  0.0486  &  0.0755  &0.08783&0.1017  \\ 

&FNN-Adam & 0.0203 $\pm$  0.0012     	&  0.0272 $\pm$  0.0008       &\textbf{0.0323 $\pm$  0.0004}	       &\textbf{0.0357 $\pm$  0.0008}\\

&FNN-SGD  &   0.0242 $\pm$   0.0020 	&  0.0299 $\pm$   0.0015       &0.0350 $\pm$  0.0021	       &0.0380 $\pm$   0.0018\\

&LSTM   & 0.0168 $\pm$   0.0003    	& \textbf{0.0248 $\pm$   0.0006}      &0.0333 $\pm$  0.0010	       &0.0367$\pm$   0.0015 \\

&BD-LSTM   &   \textbf{0.0165 $\pm$  0.0002}  	&  0.0253 $\pm$  0.0004     &0.0356 $\pm$ 0.0010& 0.0409 $\pm$  0.0015\\

&ED-LSTM   &   0.0171 $\pm$  0.0003     	&  0.0271 $\pm$ 0.0010       & 0.0359 $\pm$  0.0014 	       &0.0395 $\pm$  0.0014\\

&RNN  &   0.0202 $\pm$  0.0003   	&  0.0284 $\pm$ 0.0004     &0.0348 $\pm$ 0.0004 	       & 0.0384 $\pm$  0.0003\\

&CNN &  0.0217	$\pm$0.0004   	&   0.0290	$\pm$0.0002    &0.0363	$\pm$0.0006	       & 0.0401$\pm$	0.0005\\
\hline
\end{tabular}
\end{table*}

 Tables \ref{tab:Simulated} and \ref{tab:Real} show  a comparison with related methods from the 
literature for simulated and real-world time series, respectively. We 
note that the comparison is not  fair as other 
methods may have employed different models with different 
  data processing and also in reporting of results with 
different measures of error. Moreover, some papers report best experimental run and do not show mean and standard deviation of the results. We highlight in bold the best performance for respective prediction horizon.  In Table \ref{tab:Simulated}, we compare the Mackey-Glass and Lorenz time series 
performance for  two-step-ahead prediction by real-time recurrent 
learning (RTRL) and  echo state networks (ESN) \cite{chang2012reinforced}. Note 
that * in  the results  implies  that the comparison 
is  not fair due to  different embedding 
dimension in state-space reconstruction  and it is not clear if the mean or the best run has been 
reported. We show further comparison for Mackey-Glass for 5th prediction horizon 
using Extended Kalman Filtering (EKF), the Unscented Kalman Filtering (UKF) and 
the
 Gaussian Particle Filtering (GPF),  along with their    generalized   
versions G-EKF, G-UKF
 and G-GPF, respectively \cite{Wu2013AMC}. In the case of MultiTL-KELM \cite{YE2019227}, we find that it   beats all our proposed methods for Mackey-Glass, except for the Henon time series.  In general, we find that our proposed deep learning methods (LSTM, BD-LSTM, ED-LSTM) outperform most of the methods from the literature for the simulated time series, except for the Mackey-Glass time series. 
 
In Table  \ref{tab:Real}, we compare the performance of Sunspot  time series  with support vector 
regression (SVR),  iterated (SVR-I), direct (SVR-D), and  multiple models
(M-SVR) methods \cite{zhang2013iterated}. In the respective problems, we also compare with   coevolutionary multi-task learning (CMTL) \cite{chandra2017CMTLMulti}.   We observe that  our proposed deep learning methods have given the best performance for the respective problems for most of the prediction horizons. Moreover, we find  the FNN-Adam overtakes  CMTL in all time-series problems except in 8-step ahead prediction in Mackey-Glass and 2-step ahead prediction in Lazer time series. It should also be noted that except for the Mackey-Glass and ACI-Finance time series, the deep learning methods are the best  which motivates   further applications   for challenging forecasting problems.

 \section{Discussion}

We provide a ranking of the methods in terms of performance accuracy over the test dataset across the prediction horizons in Table \ref{tab:resultranks}. We observe that FNN-SGD gives the worst performance for all time-series problems followed by FNN-Adam in most cases.   We observe that the BD-LSTM and ED-LSTM models provide one of the best performance across different problems with different properties. We also note that   across all the problems, the confidence interval of RNN is the lowest, followed by CNN which indicates that they provide more robust performance accuracy given different model initialisations in weight space.
 
  We note that it is natural for the performance accuracy  to deteriorate as the prediction horizons increases in multi-step ahead problems. The prediction is based on current values and the information gap   increases with  the prediction horizon due to our problem formulated as direct strategy of multi-step ahead prediction, as opposed to iterated prediction strategy. ACI-finance problem is  unique in  a way where there is not a major difference with simple neural networks and deep learning models (Figure 7 b) when considering the higher prediction horizons (7 - 10).

 Long term dependency problems  arise in the analysis of   time series where  the rate of decay of statistical dependence of two points increase with  time interval. Simple RNNs had difficulty in training   long-term dependency problems \cite{hochreiter1998vanishing}; hence, LSTM networks were developed \cite{hochreiter1997long}. The time series  problems in our experiments  are not long-term dependency problems; however, LSTM networks provide better performance  when compared to  simple RNNs. It seems that  the memory gates in LSTM networks help better capture information in temporal sequences, even though they do not have  long-term dependencies. We note that the memory gates in LSTM networks were originally designed to cater for the vanishing gradient problem. It seems the memory gates of LSTM networks are helpful in capturing salient features in temporal series that help in predicting future tends much better than simple RNNs. We note that simple RNNs provide better results than simple neural networks (FNN-SGD and FNN-Adam) since they are more suited for temporal series. Moreover, we find striking results given that CNNs which are suited for image processing  perform better than simple RNNs in general. This could be due to the convolutional layers in CNNs that help in better capturing hidden features for  temporal sequences. 
  
  Moving on, it is important to understand why the novel LSTM network models (ED-LSTM and BD-LSTM)  have given much better results. The ED-LSTM model was designed for language modeling tasks, primarily sequence to sequence  modelling for language translation where  encoder LSTM maps a source sequence to a fixed-length vector, and the decoder LSTM maps the vector representation back to a variable-length target sequence \cite{sutskever2014sequence}. In our case, the encoder maps an input time series to a fixed length vector and then the decoder LSTM maps the vector representation to the different prediction horizons. Although the application is different, the underlying task of mapping inputs to outputs remains the same; hence, ED-LSTM models have been very effective for multi-step ahead prediction. 
  
  Simple RNNs make use of only the previous context states for determining future states. On the other hand, BD-LSTMs process information using two  LSTM models to feature forward and backward information about the sequence at every time step \cite{graves2005framewise}. Although these have been useful for language modelling tasks, our results show that they are   applicable for mapping current and  future states for time series modelling. The  information from past and future states are somewhat preserved which seems to be the key feature in achieving better performance for multi-step prediction problems when compared to conventional LSTM models.

 \section{Conclusion and Future Work}
 
 In this paper, we  provide a comprehensive evaluation of emerging deep learning models for multi-step-ahead time series problems. Our results indicate that encoder-decoder and bi-directional LSTM networks provide   best performance for both simulated and real-world time series problems. The results have significantly improved over related time series prediction methods given in the literature. 
 
 In future work, it would be worthwhile to provide similar evaluation for multivariate time series prediction problems. Moreover, it is worthwhile to investigate the performance of given deep learning models for spatiotemporal problems, such as the prediction of certain characteristics of storms and cyclones. Further applications to other real-world problems would also be feasible, such as air pollution and energy forecasting.

 \section*{Software and Data}
We provide open source implementation in Python along with data for the respective methods  for further research \footnote{\url{https://github.com/sydney-machine-learning/deeplearning_timeseries}}.

\section*{Appendix}

\begin{table*}[htbp]
 \smaller 
 \caption{ACI-finance reporting RMSE mean and 95 \% confidence interval   ($\pm$).}
\label{tab:finance}
\begin{tabular}{llllllll}
\hline
 &  FNN-Adam& FNN-SGD & LSTM & BD-LSTM  & ED-LSTM & RNN & CNN\\
\hline
\hline
Train &  0.1628 $\pm$  0.0012 & 0.1703 $\pm$   0.0018  & 0.1471 $\pm$   0.0014  &  0.1454 $\pm$  0.0021 &  0.1437 $\pm$  0.0019  &  0.1930 $\pm$  0.0018   & 0.1655	$\pm$0.0013\\
 Test &   0.0885 $\pm$  0.0011 & 0.0988 $\pm$   0.0040  & 0.0860 $\pm$   0.0025  &  0.0915 $\pm$  0.0023 & 0.0923 $\pm$  0.0032  & 0.0936 $\pm$  0.0009 & 0.0978$\pm$	0.0011\\
Step-1 &  0.0165 $\pm$  0.0011 & 0.0209 $\pm$  0.0022  & 0.0127 $\pm$  0.0003  &  0.0127 $\pm$ 0.0002 &  0.0130 $\pm$  0.0005  &  0.0173 $\pm$  0.0004  & 0.0193$\pm$	0.0006\\
Step-2 &  0.0203 $\pm$  0.0012 & 0.0242 $\pm$   0.0020  & 0.0168 $\pm$   0.0003  &  0.0165 $\pm$  0.0002 &  0.0171 $\pm$  0.0003  &  0.0202 $\pm$  0.0003   &0.0217	$\pm$0.0004\\
Step-3 &  0.0217 $\pm$  0.0008 & 0.0266 $\pm$   0.0032  & 0.0190 $\pm$   0.0004  &  0.0194 $\pm$  0.0002 &  0.0204 $\pm$  0.0006  &  0.0228 $\pm$  0.0003 & 0.0247$\pm$	0.0003 \\
Step-4 &  0.0249 $\pm$ 0.0009 & 0.0277 $\pm$   0.0023  & 0.0220 $\pm$  0.0004  &  0.0229 $\pm$  0.0003 & 0.0239 $\pm$  0.0008  &  0.0258 $\pm$  0.0004  &0.0266$\pm$	0.0002 \\

Step-5 &  0.0272 $\pm$  0.0008 & 0.0299 $\pm$   0.0015  & 0.0248 $\pm$   0.0006  &  0.0253 $\pm$  0.0004 &  0.0271 $\pm$ 0.0010  &  0.0284 $\pm$ 0.0004 & 0.0290	$\pm$0.0002 \\

Step-6 &  0.0289 $\pm$ 0.0006 & 0.0325 $\pm$   0.0016  & 0.0281 $\pm$  0.0008 & 0.0292 $\pm$  0.0007 &  0.0302 $\pm$  0.0012  &  0.0304 $\pm$  0.0004 & 0.0315	$\pm$0.0004 \\

Step-7 & 0.0311 $\pm$  0.0005 & 0.0342 $\pm$   0.0020  & 0.0302 $\pm$   0.0008  &  0.0331 $\pm$  0.0010 &  0.0334 $\pm$  0.0014  &  0.0327 $\pm$  0.0004 & 0.0340$\pm$	0.0003 \\

Step 8 &  0.0323 $\pm$  0.0004 & 0.0350 $\pm$  0.0021  & 0.0333 $\pm$  0.0010  &  0.0356 $\pm$ 0.0010 &  0.0359 $\pm$  0.0014  &  0.0348 $\pm$ 0.0004  & 0.0363	$\pm$0.0006 \\

Step 9 &  0.0339 $\pm$  0.0005 & 0.0357 $\pm$   0.0012  & 0.0364 $\pm$   0.0013  &  0.0388 $\pm$ 0.0011 &  0.0380 $\pm$  0.0014  &  0.0371 $\pm$  0.0003 & 0.0386$\pm$	0.0006 \\

Step 10 &  0.0357 $\pm$  0.0008 & 0.0380 $\pm$   0.0018  & 0.0367$\pm$   0.0015  &  0.0409 $\pm$  0.0015 &  0.0395 $\pm$  0.0014  &  0.0384 $\pm$  0.0003 & 0.0401$\pm$	0.0005 \\
\hline
\end{tabular}

\end{table*}

\begin{table*}[htbp]
 \smaller 
 \caption{Sunspot reporting RMSE mean and 95 \% confidence interval   ($\pm$).}
\label{tab:sunspot}
\begin{tabular}{llllllll}
\hline
 &  FNN-Adam& FNN-SGD & LSTM & BD-LSTM  & ED-LSTM & RNN & CNN\\
\hline
\hline
							
Train &      
0.2043$\pm$	0.0047&
0.3230$\pm$	0.0064&
0.1418$\pm$	0.0034&
0.1369$\pm$	0.0033&
0.1210$\pm$	0.0057&
0.1875$\pm$	0.0011&0.1695$\pm$	0.0019\\
Test &      
0.1510$\pm$	0.0024&
0.2179$\pm$	0.0041&
0.1160$\pm$	0.0021&
0.1147$\pm$	0.0020&
0.1322$\pm$	0.0032&
0.1342$\pm$	0.0004&0.1487$\pm$	0.0015\\
Step-1 &      
0.0163$\pm$	0.0016&
0.0281$\pm$	0.0032&
0.0072$\pm$	0.0005&
0.0086$\pm$	0.0006&
0.0109$\pm$	0.0006&
0.0132$\pm$	0.0004&0.0186$\pm$	0.0002\\
Step-2 &      
0.0236$\pm$	0.0015&
0.0352$\pm$	0.0022&
0.0148$\pm$	0.0007&
0.0155$\pm$	0.0007&
0.0170$\pm$	0.0004&
0.0212$\pm$	0.0003&0.0257$\pm$	0.0002\\
Step-3 &      
0.0311$\pm$	0.0013&
0.0441$\pm$	0.0025&
0.0220$\pm$	0.0005&
0.0222$\pm$	0.0006&
0.0237$\pm$	0.0003&
0.0285$\pm$	0.0002&0.0321$\pm$	0.0003\\
Step-4 &      
0.0350$\pm$	0.0006&
0.0518$\pm$	0.0022&
0.0275$\pm$	0.0005&
0.0276$\pm$	0.0006&
0.0292$\pm$	0.0003&
0.0346$\pm$	0.0002&0.0376$\pm$	0.0003\\
Step-5 &      
0.0407$\pm$	0.0012&
0.0610$\pm$	0.0024&
0.0321$\pm$	0.0006&
0.0318$\pm$	0.0007&
0.0348$\pm$	0.0004&
0.0395$\pm$	0.0002&	0.0419	$\pm$0.0004\\
Step-6 &      
0.0464$\pm$	0.0016&
0.0677$\pm$	0.0027&
0.0360$\pm$	0.0006&
0.0358$\pm$	0.0006&
0.0402$\pm$	0.0006&
0.0431$\pm$	0.0002&0.0457$\pm$	0.0004\\
Step-7 &      
0.0514$\pm$	0.0019&
0.0771$\pm$	0.0020&
0.0397$\pm$	0.0006&
0.0395$\pm$	0.0006&
0.0458$\pm$	0.0011&
0.0463$\pm$	0.0002&	0.0498$\pm$	0.0005\\
Step 8 &      
0.0582$\pm$	0.0019&
0.0856$\pm$	0.0023&
0.0449$\pm$	0.0007&
0.0440$\pm$	0.0005&
0.0519$\pm$	0.0016&
0.0503$\pm$	0.0002&0.0555$\pm$	0.0006\\
Step 9 &      
0.0653$\pm$	0.0016&
0.0931$\pm$	0.0023&
0.0509$\pm$	0.0009&
0.0498$\pm$	0.0007&
0.0590$\pm$	0.0020&
0.0564$\pm$	0.0002&	0.0633$\pm$	0.0007\\
Step 10 &      
0.0745$\pm$	0.0020&
0.1012$\pm$	0.0019&
0.0587$\pm$	0.0010&
0.0576$\pm$	0.0010&
0.0673$\pm$	0.0022&
0.0641$\pm$	0.0003&	0.0723$\pm$	0.0008\\
\hline
 
\end{tabular}

\end{table*}

\begin{table*}[htbp]
 \smaller 
 \caption{Lazer reporting RMSE mean and 95 \% confidence interval   ($\pm$).}
\label{tab:lazer}
\begin{tabular}{llllllll}
\hline
 &  FNN-Adam& FNN-SGD & LSTM & BD-LSTM  & ED-LSTM & RNN &CNN\\
\hline
\hline
									
Train &  0.3371 $\pm$  0.0026 & 0.4251 $\pm$   0.0157  & 0.1954 $\pm$   0.0082  & 0.1619 $\pm$  0.0106 &  0.1166 $\pm$  0.0147  &  0.3210 $\pm$  0.0020  & 0.2151$\pm$	0.0029\\

Test &  0.2537 $\pm$  0.0024 & 0.2821 $\pm$   0.0098  & 0.1910 $\pm$   0.0042  &  0.2007 $\pm$  0.0042 &  0.2020 $\pm$  0.0057  &  0.2580 $\pm$  0.0031  & 0.2240$\pm$	0.0025\\

Step-1 &  0.0746 $\pm$  0.0027 & 0.0895 $\pm$   0.0041  & 0.0577 $\pm$   0.0021  &  0.0439 $\pm$  0.0027 & 0.0490 $\pm$  0.0039  &  0.0641 $\pm$  0.0037 & 0.0942$\pm$	0.0025\\

Step-2 &  0.1043 $\pm$  0.0018 & 0.0983 $\pm$   0.0046  & 0.0725 $\pm$ 0.0027  &  0.0892 $\pm$  0.0022 &  0.0894 $\pm$  0.0013  & 0.1176 $\pm$  0.0019  & 0.0729$\pm$	0.0014\\

Step-3 &  0.0820 $\pm$ 0.0026 & 0.0816 $\pm$   0.0028  & 0.0807 $\pm$   0.0007  &  0.0773 $\pm$ 0.0016 & 0.0707 $\pm$  0.0014  & 0.0832 $\pm$ 0.0026 & 0.0684$\pm$	0.0022\\

Step-4 &  0.0764 $\pm$ 0.0017 &0.0852 $\pm$   0.0048  &0.0697 $\pm$   0.0015  &  0.0547 $\pm$ 0.0018 &  0.0601 $\pm$  0.0029  & 0.0762 $\pm$  0.0008  &	0.0671$\pm$	0.0013 \\

Step-5 &  0.0761 $\pm$  0.0019 & 0.0874 $\pm$  0.0072  & 0.0512 $\pm$   0.0015  &  0.0596 $\pm$  0.0036 &  0.0694 $\pm$  0.0073  &  0.0755 $\pm$ 0.0011 & 0.0701$\pm$	0.0020\\

Step-6 &  0.0691 $\pm$  	0.0013 & 0.0787 $\pm$   0.0037  & 0.0540 $\pm$  0.0016 &  0.0655 $\pm$  	0.0014 &  0.0606 $\pm$  0.0041  & 0.0730 $\pm$  0.0015&  0.0677	$\pm$0.0014 \\

Step-7 &  0.0632 $\pm$  0.0013 & 0.0740 $\pm$   	0.0061  & 0.0537 $\pm$   0.0024  & 0.0601 $\pm$  0.0007 &  0.0582 $\pm$  0.0031  &  0.0643 $\pm$ 	0.0009& 	0.0643 $\pm$	0.0041 \\

Step 8 &  
0.0642	$\pm$ 0.0020 & 
0.0864$\pm$ 	0.0053 & 
0.0464$\pm$ 	0.0015& 
0.0460$\pm$ 	0.0015& 
0.0510$\pm$ 	0.0027& 
0.0611$\pm$ 	0.0015& 0.0593$\pm$	0.0029\\
Step 9 &  
0.0891$\pm$	0.0021&
0.1032$\pm$	0.0042&
0.0507$\pm$	0.0021&
0.0599$\pm$	0.0019&
0.0527$\pm$	0.0021&
0.0882$\pm$	0.0023& 0.0773$\pm$	0.0019\\
Step 10 &  
0.0924$\pm$	0.0018&
0.0968$\pm$	0.0052&
0.0561$\pm$	0.0044&
0.0631$\pm$	0.0037&
0.0615$\pm$	0.0030&
0.0947$\pm$	0.0027& 0.0577$\pm$	0.0018\\
\hline
 
\end{tabular}

\end{table*}

\begin{table*}[htbp]
 \smaller 
 \caption{Henon reporting RMSE mean and 95 \% confidence interval   ($\pm$).}
\label{tab:henon}
\begin{tabular}{llllllll}
\hline
 &  FNN-Adam& FNN-SGD & LSTM & BD-LSTM  & ED-LSTM & RNN & CNN\\
\hline
\hline
		
Train &  0.5470 $\pm$  0.0023 & 0.5670 $\pm$   0.0015  & 0.4542 $\pm$   0.0071  &  0.4014 $\pm$  0.0100 &  0.3235 $\pm$  0.0316  &  0.5247 $\pm$  0.0027 & 0.4728$\pm$ 	0.0038\\

Test &  0.5378 $\pm$  0.0022 & 0.5578 $\pm$  0.0016  & 0.4516 $\pm$  0.0052  &  0.4127 $\pm$ 0.0066 &  0.3294 $\pm$  0.0290  &  0.5162 $\pm$  0.0027&  	0.4779$\pm$ 	0.0027 \\

Step-1 &  0.1465 $\pm$  0.0058 & 0.1725 $\pm$   0.0031  & 0.0287 $\pm$   0.0045  & 0.0241 $\pm$  0.0014 &  0.0226 $\pm$  0.0039  & 0.0885 $\pm$  0.0093& 	0.0650$\pm$ 	0.0018  \\

Step-2 &  0.1606 $\pm$  0.0024 & 0.1711 $\pm$   0.0018  & 0.0682 $\pm$   0.0058  &  0.0448 $\pm$  0.0026 &  0.0454 $\pm$  0.0069  &  0.1515$\pm$ 0.0016 & 	0.0859	$\pm$ 0.0038 \\

Step-3 &  0.1610 $\pm$  0.0008 & 0.1707 $\pm$   0.0017  & 0.0920 $\pm$  0.0066  &  0.0610 $\pm$  0.0077 &  0.0517 $\pm$  0.0122  &  0.1577 $\pm$  0.0003&  	0.1411$\pm$ 	0.0021 \\

Step-4 &  0.1714 $\pm$  0.0009 & 0.1760 $\pm$   0.0011  & 0.1386 $\pm$  0.0044  &  0.0925 $\pm$  0.0077&  0.0609 $\pm$ 0.0154  &  0.1643 $\pm$  0.0011 & 	0.1519$\pm$ 	0.0022\\

Step-5 &  0.1731 $\pm$  0.0005 &0.1769 $\pm$   0.0007  &0.1584 $\pm$   0.0010  &  0.1287 $\pm$  0.0046 &  0.0694 $\pm$  0.0161  &  0.1718 $\pm$  0.0001 & 	0.1601$\pm$ 	0.0007 \\

Step-6 &  0.1758 $\pm$  0.0006 & 0.1777 $\pm$   0.0010  & 0.1642 $\pm$   0.0007  &  0.1538 $\pm$  0.0017 &  0.0868 $\pm$  0.0165  &  0.1730 $\pm$  0.0003 & 	0.1674$\pm$ 	0.0004 \\

Step-7 &  0.1786 $\pm$  0.0006 & 0.1814 $\pm$   0.0009  & 0.1684$\pm$  0.0011  &  0.1593 $\pm$  0.0016 &  0.1120 $\pm$  0.0139  &  0.1764 $\pm$  0.0003& 	0.1721$\pm$ 	0.0006  \\

Step 8 &  0.1781 $\pm$  0.0005 & 0.1805 $\pm$   0.0012  & 0.1707 $\pm$   0.0008  &  0.1697 $\pm$  0.0008 &  0.1371 $\pm$  0.0107  &  0.1768 $\pm$  0.0001 & 0.1718$\pm$ 	0.0003\\

Step 9 &  0.1757 $\pm$  0.0004 & 0.1788 $\pm$   0.0011  & 0.1723 $\pm$   0.0005  &  0.1737 $\pm$ 0.0004 &  0.1524 $\pm$  0.0077  &  0.1752 $\pm$  0.0001 &  	0.1752$\pm$ 	0.0003\\

Step 10 &  0.1762 $\pm$  0.0009 & 0.1773 $\pm$   0.0011  & 0.1756 $\pm$   0.0005  &  0.1733 $\pm$  0.0003 &  0.1689 $\pm$  0.0046  &  0.1751 $\pm$  0.0002 &  0.1737$\pm$ 	0.0002\\

\hline
 
\end{tabular}

\end{table*}

\begin{table*}[htbp]
\smaller 
\caption{Lorenz reporting RMSE mean and 95 \% confidence interval   ($\pm$).}
\label{tab:lorenz}
\begin{tabular}{llllllll}
\hline
 &  FNN-Adam& FNN-SGD & LSTM & BD-LSTM  & ED-LSTM & RNN & CNN\\
\hline
\hline
	
Train &  
0.1932$\pm$	0.0124&
0.2761$\pm$	0.0048&
0.0242$\pm$	0.0086&
0.0300$\pm$	0.0127&
0.0225$\pm$	0.0031&
0.0538$\pm$	0.0091&0.0354$\pm$	0.0032\\
Test &  
0.1809$\pm$	0.0119&
0.2649$\pm$	0.0048&
0.0254$\pm$	0.0093&
0.0310$\pm$	0.0137&
0.0234$\pm$	0.0031&
0.0542$\pm$	0.0097&	0.0347$\pm$	0.0029\\
Step-1 &  
0.0183$\pm$	0.0042&
0.0336$\pm$	0.0035&
0.0025$\pm$	0.0006&
0.0043$\pm$	0.0019&
0.0051$\pm$	0.0015&
0.0113$\pm$	0.0013&	0.0055$\pm$	0.0006\\
Step-2 &  
0.0206$\pm$	0.0046&
0.0432$\pm$	0.0030&
0.0033$\pm$	0.0010&
0.0054$\pm$	0.0026&
0.0044$\pm$	0.0012&
0.0129$\pm$	0.0012&	0.0067	$\pm$0.0007\\
Step-3 &    
0.0253$\pm$	0.0043&
0.0547$\pm$	0.0028&
0.0042$\pm$	0.0019&
0.0064$\pm$	0.0031&
0.0046$\pm$	0.0010&
0.0143$\pm$	0.0016&	0.0077$\pm$	0.0009\\
Step-4 &    
0.0334$\pm$	0.0048&
0.0651$\pm$	0.0032&
0.0051$\pm$	0.0020&
0.0074$\pm$	0.0035&
0.0052$\pm$	0.0009&
0.0151$\pm$	0.0018&0.0087$\pm$	0.0009\\
Step-5 &    
0.0481$\pm$	0.0072&
0.0787$\pm$	0.0030&
0.0064$\pm$	0.0026&
0.0079$\pm$	0.0036&
0.0059$\pm$	0.0009&
0.0155$\pm$	0.0024&	0.0098$\pm$	0.0009\\
Step-6 &    
0.0527$\pm$	0.0076&
0.0866$\pm$	0.0033&
0.0073$\pm$	0.0029&
0.0094$\pm$	0.0046&
0.0068$\pm$	0.0008&
0.0164$\pm$	0.0029&	0.0109	$\pm$0.0010\\
Step-7 &    
0.0613$\pm$	0.0064&
0.0944$\pm$	0.0031&
0.0089$\pm$	0.0033&
0.0107$\pm$	0.0047&
0.0079$\pm$	0.0009&
0.0171$\pm$	0.0036&	0.0120	$\pm$0.0010\\
Step 8 &    
0.0678$\pm$	0.0058&
0.1027$\pm$	0.0025&
0.0101$\pm$	0.0038&
0.0125$\pm$	0.0057&
0.0090$\pm$	0.0009&
0.0186$\pm$	0.0042&	0.0132$\pm$	0.0011\\
Step 9 &    
0.0885$\pm$	0.0060&
0.1116$\pm$	0.0027&
0.0117$\pm$	0.0045&
0.0133$\pm$	0.0056&
0.0100$\pm$	0.0010&
0.0199$\pm$	0.0049&	0.0142$\pm$	0.0013\\
Step 10 &    
0.0859$\pm$	0.0065&
0.1178$\pm$	0.0026&
0.0129$\pm$	0.0042&
0.0146$\pm$	0.0059&
0.0110$\pm$	0.0012&
0.0226$\pm$	0.0058&	0.0157$\pm$	0.0015\\
\hline
 
\end{tabular}

\end{table*}

\begin{table*}[htbp]
\smaller 
\caption{Mackey-Glass reporting RMSE mean and 95 \% confidence interval   ($\pm$).}
\label{tab:mackey}
\begin{tabular}{llllllll}
\hline
 &  FNN-Adam& FNN-SGD & LSTM & BD-LSTM  & ED-LSTM & RNN & CNN\\
				
\hline
\hline
Train &      
0.1810$\pm$	0.0097&
0.2576$\pm$	0.0053&
0.0890$\pm$	0.0076&
0.0750$\pm$	0.0067&
0.0587$\pm$	0.0090&
0.1317$\pm$	0.0003& 0.0854$\pm$		0.0052\\
Test &      
0.1822$\pm$	0.0098&
0.2599$\pm$	0.0054&
0.0897$\pm$	0.0075&
0.0765$\pm$	0.0068&
0.0602$\pm$	0.0090&
0.1321$\pm$	0.0003&0.0868$\pm$		0.0048\\
Step-1 &      
0.0162$\pm$	0.0037&
0.0485$\pm$	0.0061&
0.0056$\pm$	0.0011&
0.0052$\pm$	0.0009&
0.0059$\pm$	0.0010&
0.0078$\pm$	0.0001&	0.0075$\pm$		0.0007\\
Step-2 &      
0.0256$\pm$	0.0038&
0.0621$\pm$	0.0051&
0.0080$\pm$	0.0014&
0.0083$\pm$	0.0015&
0.0076$\pm$	0.0014&
0.0142$\pm$	0.0001&0.0120$\pm$		0.0010\\
Step-3 &      
0.0398$\pm$	0.0051&
0.0686$\pm$	0.0046&
0.0120$\pm$	0.0018&
0.0117$\pm$	0.0019&
0.0103$\pm$	0.0020&
0.0214$\pm$	0.0001&	0.0167$\pm$		0.0013\\
Step-4 &      
0.0457$\pm$	0.0045&
0.0745$\pm$	0.0049&
0.0178$\pm$	0.0020&
0.0155$\pm$	0.0023&
0.0133$\pm$	0.0024&
0.0290$\pm$	0.0001&0.0216$\pm$		0.0015\\
Step-5 &      
0.0520$\pm$	0.0044&
0.0785$\pm$	0.0025&
0.0238$\pm$	0.0024&
0.0202$\pm$	0.0026&
0.0168$\pm$	0.0027&
0.0365$\pm$	0.0001&	0.0262$\pm$		0.0016\\
Step-6 &      
0.0581$\pm$	0.0042&
0.0880$\pm$	0.0031&
0.0298$\pm$	0.0025&
0.0244$\pm$	0.0027&
0.0200$\pm$	0.0031&
0.0434$\pm$	0.0001&0.0301$\pm$		0.0017\\
Step-7 &      
0.0680$\pm$	0.0047&
0.0912$\pm$	0.0031&
0.0341$\pm$	0.0028&
0.0288$\pm$	0.0028&
0.0227$\pm$	0.0033&
0.0496$\pm$	0.0001&	0.0332$\pm$		0.0018\\
Step 8 &      
0.0727$\pm$	0.0050&
0.0937$\pm$	0.0022&
0.0381$\pm$	0.0029&
0.0318$\pm$	0.0027&
0.0248$\pm$	0.0036&
0.0547$\pm$	0.0001&	0.0354$\pm$		0.0018\\
Step 9 &      
0.0761$\pm$	0.0046&
0.0963$\pm$	0.0031&
0.0406$\pm$	0.0030&
0.0343$\pm$	0.0027&
0.0261$\pm$	0.0038&
0.0586$\pm$	0.0001&	0.0364$\pm$		0.0017\\
Step 10 &      
0.0777$\pm$	0.0043&
0.0990$\pm$	0.0026&
0.0418$\pm$	0.0033&
0.0359$\pm$	0.0026&
0.0271$\pm$	0.0040&
0.0615$\pm$	0.0001&	0.0364$\pm$		0.0017\\
\hline
 $\pm$&
\end{tabular}

\end{table*}

\begin{table*}[htbp]
\smaller 
\caption{Rossler reporting RMSE mean and 95 \% confidence interval   ($\pm$).}
\label{tab:rossler}
\begin{tabular}{llllllll}
\hline
 &  FNN-Adam& FNN-SGD & LSTM & BD-LSTM  & ED-LSTM & RNN& CNN \\
\hline
\hline
		
Train &      
0.1546$\pm$	0.0087&
0.3757$\pm$	0.0100&
0.0416$\pm$	0.0074&
0.0281$\pm$	0.0098&
0.0374$\pm$	0.0082&
0.1088$\pm$	0.0009&0.0367$\pm$		0.0059\\
Test &      
0.1473$\pm$	0.0098&
0.4314$\pm$	0.0122&
0.0488$\pm$	0.0054&
0.0349$\pm$	0.0070&
0.0427$\pm$	0.0072&
0.1030$\pm$	0.0008&	0.0454$\pm$		0.0052\\
Step-1 &      
0.0148$\pm$	0.0026&
0.0467$\pm$	0.0077&
0.0080$\pm$	0.0009&
0.0038$\pm$	0.0008&
0.0085$\pm$	0.0025&
0.0186$\pm$	0.0009&	0.0086$\pm$		0.0010\\
Step-2 &      
0.0202$\pm$	0.0024&
0.0666$\pm$	0.0058&
0.0086$\pm$	0.0011&
0.0047$\pm$	0.0014&
0.0082$\pm$	0.0019&
0.0218$\pm$	0.0005&0.0105$\pm$		0.0011\\
Step-3 &      
0.0252$\pm$	0.0022&
0.0910$\pm$	0.0083&
0.0099$\pm$	0.0013&
0.0061$\pm$	0.0017&
0.0098$\pm$	0.0018&
0.0250$\pm$	0.0005&	0.0118$\pm$		0.0011\\
Step-4 &      
0.0322$\pm$	0.0024&
0.1060$\pm$	0.0078&
0.0117$\pm$	0.0014&
0.0072$\pm$	0.0020&
0.0112$\pm$	0.0021&
0.0290$\pm$	0.0004&	0.0122$\pm$		0.0014\\
Step-5 &      
0.0400$\pm$	0.0039&
0.1257$\pm$	0.0082&
0.0135$\pm$	0.0015&
0.0084$\pm$	0.0021&
0.0128$\pm$	0.0021&
0.0314$\pm$	0.0004&0.0122$\pm$		0.0016\\
Step-6 &      
0.0490$\pm$	0.0063&
0.1424$\pm$	0.0069&
0.0155$\pm$	0.0019&
0.0100$\pm$	0.0023&
0.0141$\pm$	0.0021&
0.0339$\pm$	0.0004&	0.0128$\pm$		0.0019\\
Step-7 &      
0.0527$\pm$	0.0049&
0.1572$\pm$	0.0063&
0.0170$\pm$	0.0020&
0.0120$\pm$	0.0024&
0.0151$\pm$	0.0022&
0.0360$\pm$	0.0004&	0.0139$\pm$		0.0020\\
Step 8 &      
0.0603$\pm$	0.0050&
0.1664$\pm$	0.0075&
0.0185$\pm$	0.0022&
0.0142$\pm$	0.0027&
0.0159$\pm$	0.0024&
0.0382$\pm$	0.0004&	0.0157$\pm$		0.0020\\
Step 9 &      
0.0621$\pm$	0.0036&
0.1818$\pm$	0.0067&
0.0204$\pm$	0.0024&
0.0157$\pm$	0.0030&
0.0167$\pm$	0.0026&
0.0404$\pm$	0.0005&	0.0185$\pm$		0.0021\\
Step 10 &      
0.0673$\pm$	0.0056&
0.1881$\pm$	0.0078&
0.0225$\pm$	0.0026&
0.0178$\pm$	0.0032&
0.0180$\pm$	0.0030&
0.0424$\pm$	0.0004&	0.0220$\pm$		0.0022\\
\hline
 
\end{tabular}

\end{table*}

\bibliographystyle{IEEEtran}
\bibliography{usyd,Chandra-Rohitash,Bays,2018,rr,sample,sample_,aicrg,2020June}
 
\begin{IEEEbiography}[{\includegraphics[width=1in,height=1.25in,clip,keepaspectratio]{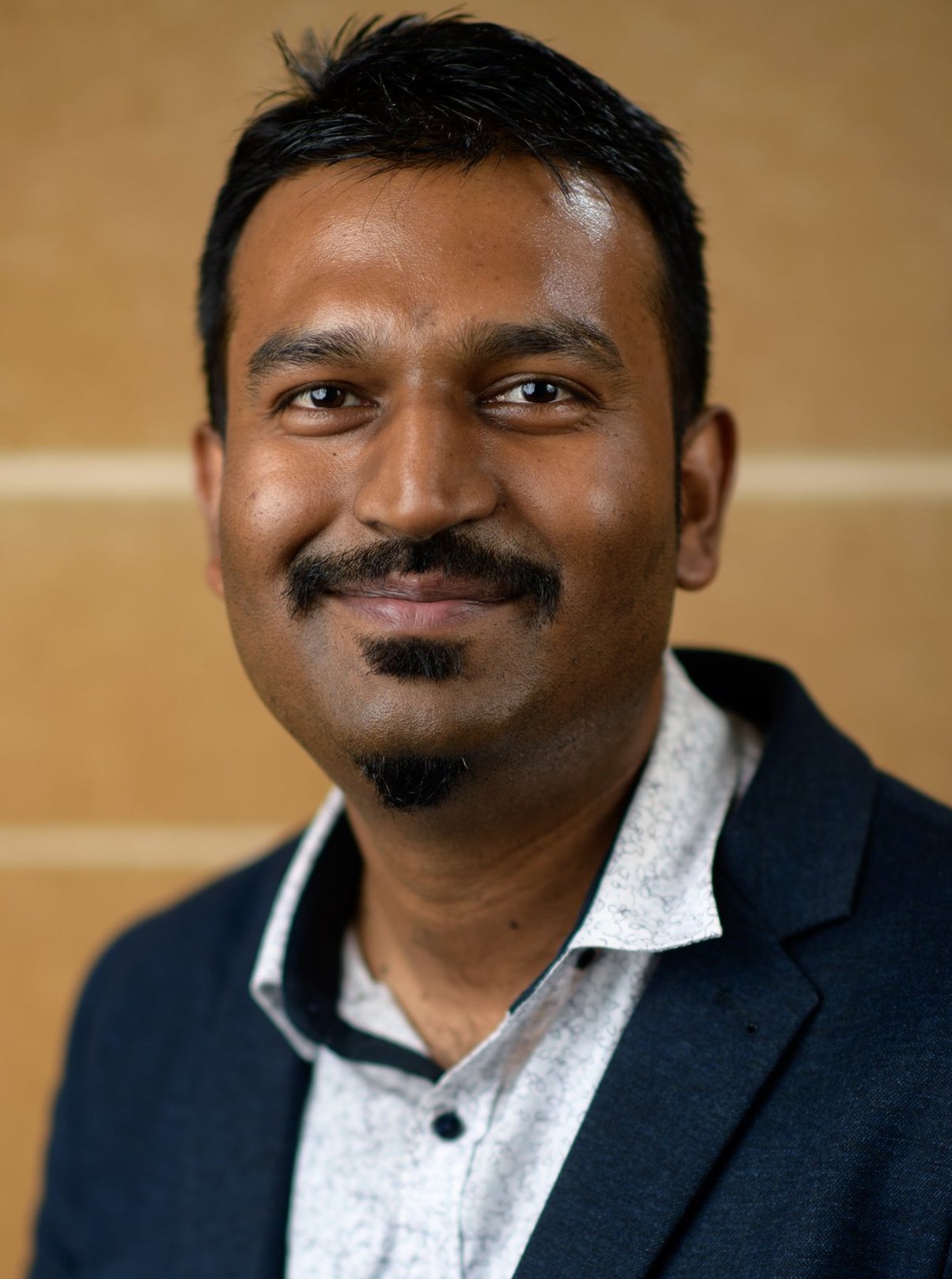}}]{Rohitash Chandra} (IEEE M'2009--SM'2021) is a Senior Lecturer in Data Science at the UNSW School of Mathematics and Statistics. Dr Chandra has built a program of research encircling methodologies and applications of artificial intelligence; particularly in areas of Bayesian deep learning, neuroevolution, Bayesian inference via MCMC, climate extremes, landscape and reef evolution models, and mineral exploration.  Dr Chandra has been developing novel methods for machine learning inspired by neural systems and learning behaviour that include transfer and multi-task learning, with the goal of developing modular deep learning methods. The current focus has been on Bayesian deep learning with a focus on recurrent, convolutional, and graph neural networks, with application to language models involving sentiment analysis and COVID-19. Dr Chandra has attracted multi-million dollar funding with a leading international interdisciplinary team and  part of the Australian Research Council (ARC ITTC) Training Centre for Data Analytics in Minerals and Resources (2020-2025). Dr Chandra is an Associate Editor for Neurocomputing,   IEEE Transactions on Neural Networks and Learning Systems, and  Geoscientific Model Development (Topical Editor).

\end{IEEEbiography}

\begin{IEEEbiography}[
{\includegraphics[width=1in,height=1.25in,clip,keepaspectratio]{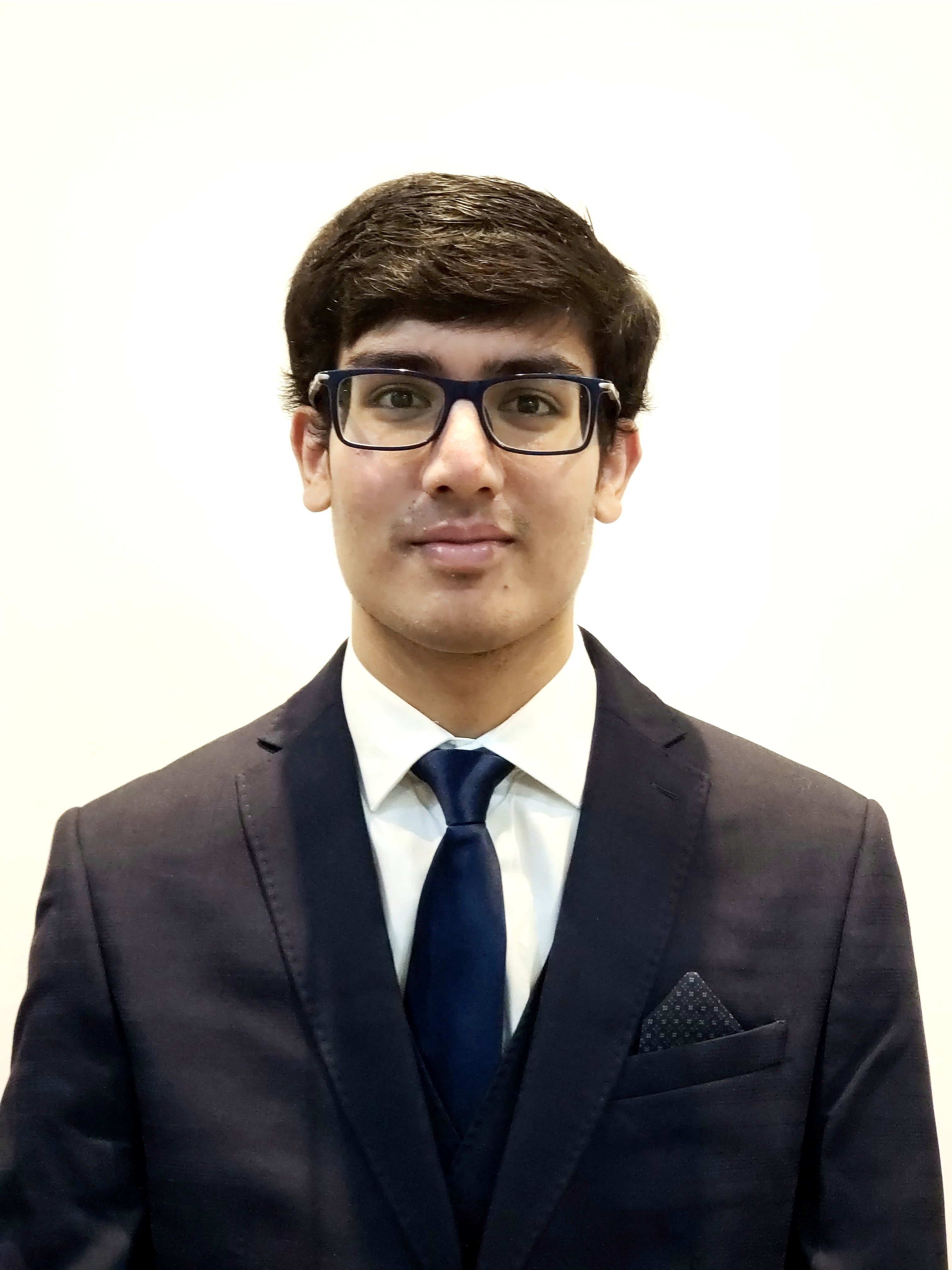}}]{Shaurya Goyal}  (IIT Delhi 2018-present) is  pursuing an integrated B.Tech and M.Tech degree in Mathematics and Computing Department at  Indian Institute of Technology, Delhi. His research interests mainly lie in areas of deep learning, reinforcement learning, and time series prediction.

\end{IEEEbiography}
 
\begin{IEEEbiography}[{\includegraphics[width=1in,height=1.5in,clip,keepaspectratio]{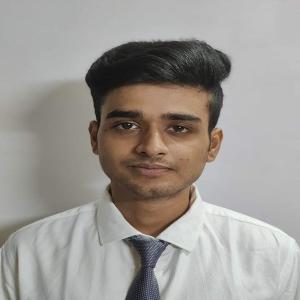}}]{Rishabh Gupta} (IIT Kharagpur 2016-2021) is   pursuing an integrated M.Sc. degree in applied geology and a minor in computer science and engineering at Indian Institute of Technology, Kharagpur. His research interests are in areas of robotics, computer vision, deep learning, and reinforcement learning.

\end{IEEEbiography}

 \EOD

\end{document}